\documentclass[12pt,onecolumn,letter]{IEEEtran}
\usepackage{fancyhdr}
\usepackage{stmaryrd}
\usepackage{amsfonts}

\usepackage{graphicx,times,amsmath} 
\usepackage[utf8]{inputenc}
\usepackage[english]{babel}
\usepackage{algorithm2e}

\makeatletter 
\renewcommand{\@algocf@capt@plain}{above}

\newtheorem{theorem}{Theorem}
\usepackage{mathtools}
\usepackage{hyperref}
\usepackage{multirow}

\usepackage[nomarkers,figuresonly]{endfloat}
\usepackage[justification=centering]{caption}

\IEEEoverridecommandlockouts    

\usepackage{graphicx}
\usepackage{amsmath,amssymb}
\usepackage{times}
\usepackage{color}
\usepackage[pagewise]{lineno}
\usepackage{threeparttable}
\newcommand{\spacedouble}{\renewcommand{\baselinestretch}{1.5}\Huge\normalsize}
\spacedouble

\newcommand{\cL}{{\cal L}}

\newcommand{\cX}{{\cal X}}

\newcommand{\cZ}{{\cal Z}}

\def\f{\mathbf f}
\def\g{\mathbf g}
\def\l{\mathbf l}

\def\p{\mathbf p}

\def\r{\mathbf r}

\def\v{\mathbf v}
\def\w{\mathbf w}
\def\x{\mathbf x}
\def\y{\mathbf y}
\def\z{\mathbf z}

\newcommand{\marginlabel}[1]{}

               {\list{}{\leftmargin=3mm 
                        \labelwidth\z@ \itemindent-\leftmargin
                        }}%
               {\endlist}
\makeatother

\newtheorem{lemma}{Lemma}

\title{\LARGE\bf TITLE OF INVENTION\\
Developmental Network Two,  Its Optimality, and\\
Emergent Turing Machines}

\author
{
	IINVENTORS:\\
	Juyang Weng, U.S. Citizen, Resident of U.S.\\
	Zejia Zheng, Citizen of China, Resident of U.S.\\
	Xiang Wu, Citizen of China, Resident of China\\
	U.S. Provisional Patent Application Serial Number: 62/624,898, filed Feb. 1, 2018\\ 
U.S. Patent Application Number: 16/265,212, filed Feb. 1, 2019\\
Approval pending
}

\begin{document}
	\maketitle

	
	\section*{BACKGROUND OF THE INVENTION}
	
	Defined in Wikipedia, weak AI is AI that is focused on one narrow task.   Still by Wikipedia definition,  the goal of Strong AI or True AI, is to develop artificial intelligence to the point where the machine's intellectual capability is functionally equal to a human's.
	
	\subsection{Need Strong AI}

Our further view is that weak AI is brittle in natural settings.  By natural setting, we mean any setting other than purely human
handcrafted settings, such as a computer game setting (e.g., chess or go) and a computer language setting (e.g., C or C++ computer languages).   

Examples of natural settings are everywhere, from the time you wake up everyday till your brain sleeps at night.   For example, self-driving cars on any roads in the real world must deal with natural settings.   
Because self-driving in any natural setting is a muddy task as defined in Weng 2012 \cite{WengNAI12},
we think that is why self-driving cars have been a heavily invested industrial area for over a decade but there still have no commercially available self-driving cars due to a very long and open-ended list problems, such as those reported in California Autonomous Vehicle Disengagement Reports 2016.   The more 
a company does road tests, the more problems it discovers.    

Why does weak AI have no hope for natural settings, Weng 2012  \cite{WengNAI12} listed five categories
 of reasons --- he called muddiness measures.    A total of 26 muddiness measures, or why weak AI is brittle, 
 has been listed \cite{WengNAI12}.   
 
 For example, environmental controlledness is one.  If city driving
and high-way driving settings are not controlled, you might run into a case where a policeman is 
 making manual gestures to you right at the center of the lane!   Can any laser device that exists today 
 process visual information from the policeman that is far enough to avoid running over him?  None. 
 Such cases are not often, of course, but can any company afford customers to sue due to such weak AI flaws?
 
 There are many other examples.   That is why we need strong AI.
 
 \subsection{Make Strong AI}
 
	A human baby demonstrates impressive abilities in general purpose learning: it receives natural sensory inputs and motor feedback from the real-world, with its behaviors increasingly rule-like and invariant to the noises in the high dimensional inputs. This task non-specific learning procedure seems to be incremental: when encountered with unfamiliar scenarios a human child would try resort to the combination of previously learned skills and gradually adjust to the new situation. Although this phenomenon (also known as concept scaffolding) has been studied by a huge body of literatures in the field of developmental psychology (e.g., \cite{cobb1994}, \cite{piaget1952origins}, \cite{piaget1973child}, \cite{piaget1959language}, \cite{piaget2015grasp}, \cite{piaget2013success}, \cite{vygotsky1980mind}), the computational mechanism for such learning to happen remains elusive.

	In this invention, we teach the essential mechanisms of enabling a robot to learn like a child (cumulative, incremental and transfer old knowledge to new settings). Three conceptual steps guide us toward the targeted framework.

	\textbf{Incremental learning}. Human beings learn new skills without forgetting old knowledge. The most successful machine learning systems, on the other hand, relies on batch data and error back-propagation, which disrupts long-term memory thus is incapable of learning incrementally without reviewing the old dataset. Incremental learning is important in the sense that batch dataset, no matter how large it is, cannot contain all possible cases and variances for real-world application. Thus a generalizable learning mechanism must have the ability to adjust and adapt to novel situations while keeping the old learned skills intact.

	\textbf{Task non-specific learning}. Compared to the learning agents that learn to optimize a task-specific loss function (e.g., cross-entropy loss for classification and L2 loss for regression \cite{schmidhuber2015deep,JOURNALtutorial}), task non-specific learning allows the agent to transfer abstractive concepts, learn incrementally and form hierarchical concepts with emergent behavior with no human intervention (i.e., close-skulled). Task non-specificity is a must for incremental learning, as the designer cannot design task-specific loss functions for the unknown future tasks.

	\textbf{Emergent representation}. Emergent representation is formed from the system's interactions with the external world and the internal world via its sensors and effectors without using handcrafted concepts about the extra-body environments \cite{weng2012symbolic}. Compared to the brittle symbolic representations, in which internal representation contains a number of concept areas where the content of each concept and the boundary between these areas are human handcrafted  \cite{weng2012symbolic}, emergent representation networks is grounded, fault tolerant, and capable to abstract inputs with no handcrafting needed \cite{weng2012symbolic}.  Task non-specificity forbids the learning agent to use symbolic representations as it is impossible to manually define the meanings of internal representations for the real-world with numerous possible variations.

	As Weng pointed out in \cite{weng2015brain}, the Developmental Network (DN-1) was the first general-purpose emergent FA that:
	\begin{enumerate}
		\item uses fully emergent representations,
		\item allows natural sensory firing patterns,
		\item learns incrementally -- taking one-pair of sensory pattern and motor pattern at a time to update the network
		and discarding the pair immediately after, and
		\item learns a Finite Automaton error-free given enough resources.
	\end{enumerate}

	The most important characteristic of DN-1 (WWN-1 through WWN-7, discussed in detail in Sec. \ref{SE:DNs}) lies in its exact learning of entire input-output patterns using neurons with bottom up and top down connections. Training a DN-1 requires the exhaustive teaching of the input and output patterns, which may be time-consuming and labor-intensive.

	This invention teaches DN-2, a novel neural network architecture based on DN-1. DN-2 gives up exact matching of input-output patterns in DN-1. This fundamental idea of DN-2 enabled the feature hierarchy to build on important internal features while disregarding distractors in a huge space of internal features.


	The most important theorem about DN-2 in this invention can be summarized in the following sentence:

	\textit{Under the constraint of skull-closed incremental learning and the pre-defined network hyper-parameters, DN-2 optimizes its internal parameters to generate maximum likelihood firing patterns in its network areas, conditioned on its sensory and motor experience up to the network's last update.}

	An agent with DN-2 is thus a task-nonspecific, general-purpose agent that learns incrementally using natural sensory data while behaving and tuning its internal parameters in under Maximum Likelihood under a unified learning rule.

	The theorem is formally introduced in Sec. \ref{sec: optimal incremental learning}. To prove this theorem, we formulated the learning problem under Maximum Likelihood Estimation, which is attached in the Appendix. Although this theorem is heavily based on statistic concepts, its impact should reach far into the field of machine learning and artificial intelligence.

	From this theorem, we can see that DN-2 shares the following properties of DN-1:
	
	\subsection{Mechanisms of DN-1}

	\subsubsection{No symbolic modules}
	There are no symbolic modules in DN-2. Agents with symbolic functions (e.g., hand-crafted loss function, symbolically defined representations) become task specific or setting specific. Hand-crafted symbolic representation does not have optimality when a human programmer orchestrates a design. Because of its lack in optimality, there are aspects of the environment that have not been modeled in the design. When the environment has such aspects, the design fails to capture the necessary aspects. This is our reason to account for why people have thought that symbolic systems are brittle.

	We use electro-optic cameras instead of LIDAR for two main reasons. The first is the emergence of internal representations directly from camera images and actions.
	These emergent representations are not restricted by handcrafted design, so are impossible to have a major missing aspect. LIDAR devices give an excellent example of human handcrafted representation. Signals from LIDAR devices only capture the range of a single scanned point in the environment. Thus, a mirror at that point causes a failure because there is no reflected laser that the device relies on. The electro-optical information of pixels from a camera is very rich. Such information does not directly give range information. However, the emergent representations from multiple pixels provide not only range information but also other information when the range is too large for LIDAR. For example, the size of a car or pedestrian gives the information about the distance (i.e., distance from scale) provides key information for a larger range. Such long-range information is useful for a human driver to predict the behavior of a car and pedestrian before it is too late when they are near.

	\subsubsection{No curse of dimensionality problem}
	There is no curse of dimensionality problem in DN-2. The curse of dimensionality refers to the phenomena where the performance of the learning agent goes down when the number of hand-crafted features increases. DN-2 learns directly from natural images with no need to hand-craft features. Humans try to improve the performance by adding more features but this process eventually leads to opposite effects.  In DN-2, each neuron detects a feature. Such neuron-detected features are not middle-level features such as SIFT features. Instead, they are local patterns. There are many of them through emergence. The motor zone pools such many features to reach abstraction --- location invariance for type-motor and type invariance for location motor. Thus DN-2 does not have curse of dimensionality because the more neurons, the more patterns, and the better the invariance at the motor zone.

	\subsubsection{No over-fitting problem}
	There is no over-fitting problem in DN-2. The over-fitting problem refers to the phenomena where the learning agent remembers only the training data but generalizes poorly over new data, often due to the large number of parameters and the small number of training samples. As the theorem states, The DN method does not have this over-fitting problem because the number of weights as parameters is always more than the number of (scalar) observations. For example, when a new neuron is generated, the number of parameters as its weights is equal to the number of pixels in the image, not smaller. The image initializes the weight vector. Later, the same weight vector will optimally integrate more pixels where the number of parameters becomes smaller than the number of scalar observations. In other words, over-fitting by initialization of a weight vector as an image patch is optimal for the first vector observation, not suboptimal. The weight initialization ``plants'' the weight cluster right there at the new data vector. There is no need to iteratively move the cluster through a long distance that is typical for a batch processing of big data. Neuron initialization is introduced in detail in Sec. \ref{SE:DN-2-Algorithm}.

	\subsubsection{No local minima problem}
	Finally, there is no local minima problem in DN-2. Typical learning agents (e.g. agents using error back-propagation) aim to minimize a task-specific loss function, and would thus often get stuck at local minima during optimization. Because our system uses incremental learning over ``lifetime'', the initialization as explained above is optimal for the generation of a new neuron for this neighborhood. When the scalar observations exceed the number of weights in a neuron, the neuron, if it wins for this neighborhood, incrementally computes the mean vector of all past vector observations. Such an incremental computation of local means by many vectors does not suffer from the well-known local-minima problem because it converts a highly nonlinear global optimization problem into a highly nonlinear composition of many local linear problems because the incremental computation of local vectors is a local linear problem. The global nonlinearity is manifested by the nonlinear composition by manly local linear computation of neurons. The more neurons the better, and there are not local minima problems. But there is still a local minima problem in the external teaching. For example, if the teacher teaches a complex task first and then teaches a simple task, the learner does have the skill from the simple task to learn the complex task. Thus the learner will not learn the early complex task well. But this ``local minima'' problem is outside the network. The network is itself is still optimal internally.

	Moreover,  DN-2 extends DN-1 by introducing many new mechanisms.
	
	\subsection{New Mechanisms of DN-2}
	The following gives three major new mechanisms of DN-2.  Other new mechanisms of DN-2 will be discussed when we present the details the DN-2 procedure in Sec.~\ref{SE:DN-2-Algorithm}. 

	\subsubsection{Fluid hierarchy}
	\label{sec:fluid}
A fluid hierarchy of internal representation has a dynamic structure of internal hierarchy (e.g., the number of levels, regions, and their inter-connections).  Further, there is no static boundary between regions and nor the statically assigned number of neurons to each region. 
Such properties avoid the limitation of a human-handcrafted hierarchy that prevents computational 
resources to dynamically shared and reassigned between brain regions such as the cross-modal plasticity reported in \cite{VonMelchner00,Voss13}.   Such sharing not only enables an optimality to be established here, but also new representations beyond static regions.  

Such a fluid hierarchy is realized by a new mechanism ---  each internal  neuron has its own dynamic inhibition zone.   Not only excitatory connections are fully plastic, so are inhibitory connections.  The
former determines from which subspace that a neuron takes inputs; the latter determines which neurons competitively work together to form clusters so that different neurons represent different features in the subspace.  
	
	DN-1 does have multiple internal regions and those internal regions do connect with one another \cite{Song15}.  However, in DN-1 the computational resource
	assigned to each region is static and the resources are not automatically and optimally 
	reassigned to different regions. 

	\subsubsection{Multiple types of $Y$ neurons}
	\label{sec:Multiple types}
	With a globally fluid hierarchy, it is difficult for the hierarchy to quickly take the shape of an optimal 
	hierarchy.   The multiple types of $Y$ neurons, those in the internal zone $Y$, enable internal
	neurons, to
	have a good initial guess of their initial connections, like those neurons that connect with the sensory zone $X$ and motor zone $Z$. 
	
		\subsubsection{Each neuron has a 3D location}
	
	In DN-1, each neuron does not have any location.  This situation prevented the hierarchy 
	of internal representation to perform a course-to-fine approximation through lifetime.   
	Because neurons were not locationally assigned, there was no smoothness to define.  In DN-2, 
	the location of each neuron enables new neurons to initially take the neighborhood location and 
	weights of nearby neurons.   Smoothness does not prevent highly precise refinement, but it does enable the huge and highly complex representations to track optimal representation at each step in real time. 
	
	Another advantage for each neuron to have a 3D location is that we can visualize properties of
	a large number of neurons by arranging those neurons according to their locations. 

	Before diving down into the details about the developmental network, we are going to use a small example in navigation to illustrate the important power for DN-2 to learn representation.

	\subsection{Example: Shadows edges and road edges}
	\label{sec: robust edges}
	\begin{figure}
		\centering
		\includegraphics[width=1.0\linewidth]{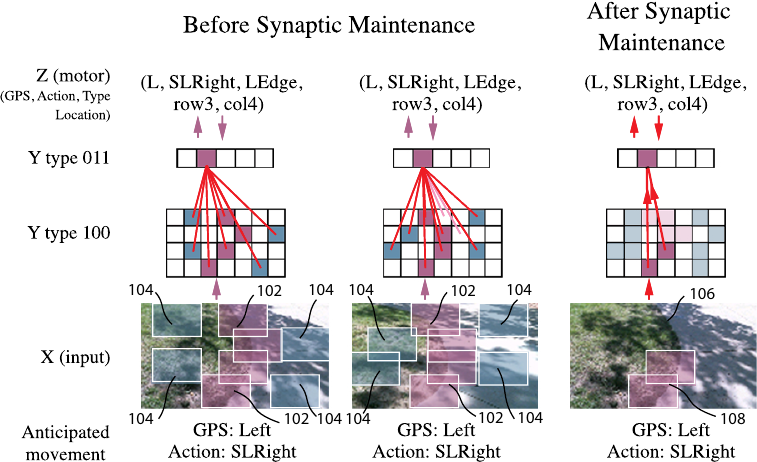}
		 \caption{}
		\label{fig:smallexample}
	\end{figure}
	
	In this example, we are going to show how DN-2 uses the learned where-what representations to form robust representation for navigation. The following discussion corresponds to the steps illustrated in Fig. \ref{fig:smallexample}.  Notation of the figure:  Before synaptic maintenance, the high-level $Y$ neuron of type 011 learns different firing patterns from low-level 100 $Y$ neurons. After synapse maintenance, the stable connections (on constant road edges) are kept while the unstable connections (on varying shadow edges) are cut from the high-level $Y$ neurons, forming a shadow-invariant representation to learn the navigation rule ``correct facing direction when left road edge is in in the middle of the input image''.

	In the context of navigation, the agent needs to pay attention to the left road edge 102 and right road edge. When making a left turn, the agent needs to adjust its facing direction by turning slightly right when the left road edge is recognized in the middle part of the image (unless there is an obstacle at the right-hand side). However, the recognition of road edges is often disrupted by the shadows 104, 106, 108, which are usually monotone and of variant shapes.
	
%

	With type 100 $Y$ neurons (internal neurons accepting local bottom-up input from images) and dynamic inhibition (neurons compete with each other locally), a sparse firing pattern in type 100 $Y$ neurons would be linked to the high-level type 011 $Y$ neuron (neurons focusing on the top-down supervision from motor zone and GPS concepts). When the network is turning left at similar situation (roughly similar location of road edge but different shadows), connections between the constantly firing 100 $Y$ neurons (focusing on road edges) and the high-level 011 $Y$ neuron would be strengthened, while its connections with 100 $Y$ neurons on shadows edges are weakened. In DN-2, synapse maintenance cuts the unstable connections (connections with high variance) while keeping the stable connections (connections with low variance) unchanged. With synapse maintenance, the high-level $Y$ neuron cuts its connection to the unstable connections with the low level $Y$ neurons focusing on shadow edges, while keeping its connection with the low-level road edge $Y$ neurons intact. At this stage, the high-level 011 $Y$ neuron becomes invariant to the changes in the shape of shadows, but learns the navigation rule ``correct facing direction when left road edge is in the middle of the input image''.

	This is just one example under the context of real-time navigation. We want our DN-2 to automatically form these rules that are too many to hand-craft.

	In the following section, we first introduced DN-2 in Sec.~\ref{SE:DNs}.  The detailed algorithm of DN-2 is presented in Sec.~\ref{SE:DN-2-Algorithm}.   Sec.~\ref{SE:optimality} establishes the
	optimality of DN-2.  Sec.~\ref{SE:Vision} presents vision experiments.  {Sec.~\ref{sec: UTM and DN-2} presents the comparison between DN-2 and Universal Turing Machine. Sec.~\ref{SE:Planning}
	describes learning long tasks, such as planning and task chaining.  Sec.~\ref{SE:Audition} report experiments for audition.  Sec.~\ref{SE:conclusions} provide concluding remarks.
	
\section*{BRIEF SUMMARY OF THE INVENTION}
	
Strong AI requires the learning engine to be task non-specific, and furthermore, to automatically construct a dynamic hierarchy of internal features. By hierarchy, we mean, e.g., short road edges and short bush edges amount to intermediate features of landmarks; but intermediate features from tree shadows are distractors that must be disregarded by the high-level landmark concept.  By dynamic, we mean the automatic selection of features while disregarding distractors is not static, but instead based on dynamic statistics (e.g. because of the instability of shadows in the context of landmark).  By internal features, we mean that they are not only sensory, but also motor, so that context from motor (state) integrates with sensory inputs to become a context-based logic machine.  We present why strong AI is necessary for any practical AI systems that work reliably in the real world.  We then present a new generation of Developmental Networks 2 --- DN-2.  With many new novelties beyond DN-1, the most important novelty of DN-2 is that the inhibition area of each internal neuron is neuron-specific and dynamic.  This enables DN-2 to automatically construct an internal hierarchy that is fluid, whose number of areas is not static as in DN-1. To optimally use the limited resource available, we establish that DN-2 is optimal in terms of maximum likelihood, under the condition of limited learning experience and limited resources. We also present how DN-2 can learn an emergent Universal Turing Machine (UTM). Together with the optimality, we present the optimal UTM.  Experiments for real-world vision-based navigation, maze planning, and audition used the same DN-2.  They successfully showed that DN-2 is for general purposes using natural and synthetic inputs.  Their automatically constructed internal representation focuses on important features while being invariant to distractors and other irrelevant context-concepts.

\section{BRIEF DESCRIPTION OF THE DRAWINGS}

The patent or application file contains at least one drawing executed in color. Copies of this patent
or patent application publication with color drawing(s) will be provided by the Office upon request
and payment of the necessary fee.

Fig.~\ref{fig:smallexample}. Demonstration of how low-level where-what representation facilitates learning complex navigation rules.  

Fig.~\ref{fig:dn2theorem}. Summary of the DN-2 framework. 

Fig.~\ref{fig:navigationtrainingroute}. Training and testing routes around the university during different times of day and with different natural lighting conditions. 

Fig.~\ref{fig:circlelinks}. Hidden type 111 neuron embedding and visualization. 

Fig. \ref{fig:lateralweights}. Projected lateral weights for type 111 neurons with no synaptic maintenance. Type 111 neurons with low firing ages forms evenly distributed attention due to the local inhibition  zones of lower level 100 neurons. 

Fig. \ref{fig:lateralweightsmasked-copy}. Projected lateral weights for type 111 neurons with synaptic maintenance. 

Fig.~\ref{fig:planningcomparison}A and Fig.~\ref{fig:planningcomparison}B. Comparison between traditional automatic planning agent Fig. 7A with DN enabled agent Fig.7B. 

Fig.~\ref{fig:sketchconcepthierarchy}. Hierarchy of concepts. 

Fig.~\ref{fig:levelz}. Correspondence between the agent's $Z$ zones with the Where-What Network concept architectures. 

Fig.~\ref{fig:mazegui}A, Fig.~\ref{fig:mazegui}B and Fig.~\ref{fig:mazegui}C: Simulated maze environment and the agent design. 

Fig.~\ref{fig:DN2Network}. DN2 network for simulated maze navigation.

Fig. \ref{fig:skillsroutes}. Skills and means taught to the navigation agent in the simulation experiment. 

Fig.~\ref{fig_diagram}. The DN2 incrementally initializes each $Y$ cluster.  

Fig.~\ref{fig_cutting}. The segmentation setting of phoneme /u:/ is illustrated. 

Fig.~\ref{fig_labelcochlea}. The structure of the modeled cochlea is shown.   

Fig.~\ref{fig_comparsion}. The bottom-up weights of motor neurons from DN-1's concept 1 zone are listed in left side. The corresponding bottom-up weights of motor neurons from DN-2's concept 1 zone are shown in right side. 

Fig.~\ref{fig_location}A, Fig.~\ref{fig_location}B, Fig.~\ref{fig_location}C, and Fig.~\ref{fig_location}D: The $Y$ neurons' locations are shown in the figures. 

\section*{DETAILED DESCRIPTION OF THE INVENTION}

	\section{Developmental Networks}
	\label{SE:DNs}

	As DN-1 is the predecessor of DN-2, let us first briefly review DN-1 so that we can see the novelty of DN-2.

	\subsection{Developmental network 1 (DN-1)}

	\begin{figure}
		\centering
		\includegraphics[width=0.7\linewidth]{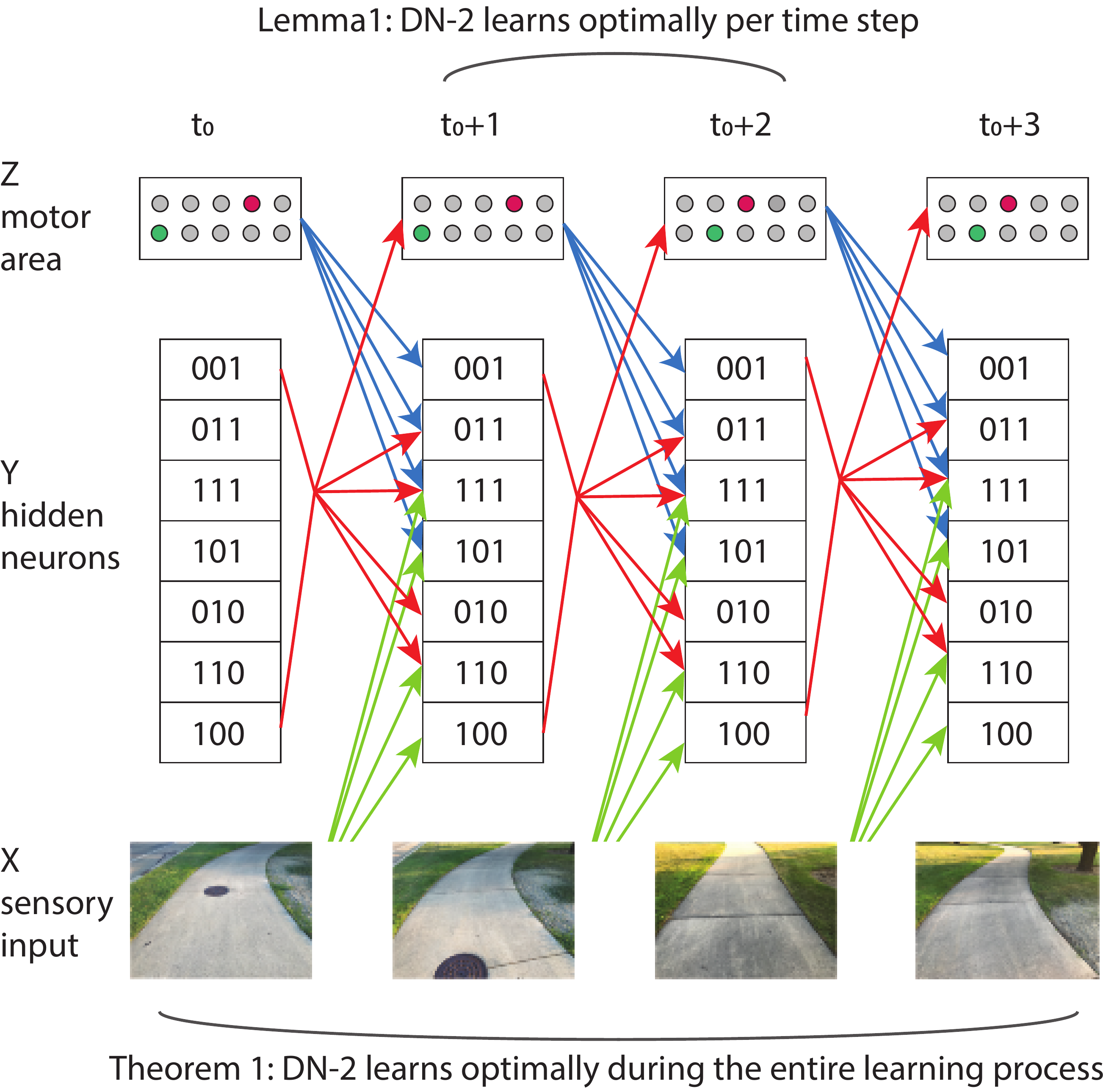}
		\caption{}
		\label{fig:dn2theorem}
	\end{figure}

The DN-1 framework has already had several implementations named as Where-What Networks (WWN), which are used to recognize and localize foreground objects directly from cluttered scenes. WWN-1 and WWN-2 \cite{WWN1} recognizes two types of information for single foregrounds over natural backgrounds: type recognition given location information and location finding given type information. WWN-3 \cite{WWN3} recognizes multiple objects in natural backgrounds. WWN-4 \cite{WWN4} demonstrates advantages of direct inputs from the sensory and motor sources. In WWN-5 \cite{WWN5}, object apparent scales are learned using the added scale motor concept zone. WWN-6 \cite{WWN6} uses synapse maintenance to form dynamic receptive fields in the hidden layers incrementally without handcrafting. WWN-7 \cite{WWN7} learns multiple scales for each foreground object using short time video input.

	\subsection{Developmental Network 2 (DN-2)}
	To extract generalizable rules and go beyond pattern matching, DN-2 uses neurons with different types of connection to form temporal dependencies among internal neurons. $Y$ to $Y$ connections (i.e., lateral connections among internal neurons), in particular, allows hierarchical representation to emerge without any handcrafting. Dynamic competition in DN-2 generates a firing pattern using feature detectors (internal neurons) abstracting the input at different levels.

The DN-2 main process is as follows:

Zones from low to high: $X$: sensory; $Y$ hidden (internal); $Z$: motor.  From low to high bottom-up.  From high to low: top-down.  From one zone to the same zone: lateral.   $X$ does not link with $Z$ zone.\\
Input zones: $X$ and $Z$. \\
Output zones: $X$ and $Z$. \\
The dimension and representation of $X$ and $Z$ zones are based on the sensors and effectors of the species. $Y$ is the skull-closed, not directly accessible by the outside.
\SetAlgoLined
\begin{enumerate}
\item Initialize the $Y$ zone:  Initialize its adaptive part $N_y=(V, L, G)$ and response vector $\y$, where $V$ is an array $n_y$ of weight vectors, $V$, the 3D location of $n_y$ neurons, and $G$ an array of $n_y$ ages.  $V$ have all random weights and
$G$ has all zero firing ages. $L$ consists of an array of $n_y$ location, where each 3D location $\l = (l_h,l_v,l_d)$ of a neuron will be initialized
when the neuron is spawn.  Set the upper bound of $Y$ neuron to be $n_y$, indicating that there are currently $c_y$ active neurons ($c_y \le n_y$).  For top-$k$ competition in a (dynamic) area inside a zone, the area needs at least $k+1$ active neurons initialized.
\item Initialize $Z$ zone:  initializes the adaptive part $N_z$ and the response vector $\z$ in a similar way.
The location of a $Z$ neuron is its muscle location on the agent body.
\item Initialize $X$ zone:  initializes the adaptive part $N_x$ and the response vector $\x$ in a similar way.
The location of each $X$ neuron is its pixel location on the agent body.
\item At time $t=0$, supervise initial state $\z$ as the starting state.  Take the first sensory input $\x$.   $\y$ does not exist.
\item At time $t=1$, compute $Y$ zone's response vector for all $Y$ neurons in parallel using
$(\z, \x)$ without using $\y'$ that does not exist yet:
\begin{equation}
  \y'=f_y(\z, \x)
\end{equation}
where $\f_y$ is the response function of zone $Y$.  Then, replace $\y \leftarrow \y'$, supervise state $\z$ and input $\x$, in parallel.
\item At time $t=2, ... $, repeat the following steps (a), (b) in parallel, before conduction step (c):
\begin{enumerate}
\item Compute $Y$ zone's response vector $\y'$ for all $Y$ neurons in parallel:
\begin{equation}
 \y'=f_y(\p_y)
\end{equation}
where the input vector $\p_y=(\x,\y,\z)$ and $f_y$ is the response function of zone $Y$.  The zone $Y$ performs neuron splitting (mitosis) if the best matched $Y$ neurons dot not match the input vector $\y$ sufficient well.  Compute $N'_y$.
\item Compute $Z$ zone's response vector $\z'$ for all $Z$ neurons in parallel:
\begin{equation}
 \z'=f_z(\p_z)
\end{equation}
where $\p_z=(\y,\z)$ and $f_z$ is the response function of zone $Z$ .  Components in $\z'$ are either supervised by the body (or teacher) if they are new or emerge from $f_z$.  Compute $N'_z$.
\item Replace for asynchronous update: $\y \leftarrow \y'$, $\z \leftarrow \z'$, $N_y\leftarrow N'_y$ and $N_z\leftarrow N'_z$.  Take sensory input $\x$.  These guarantee that all neurons at time $t$ use only old values available at $t-1$, so that all neurons at time $t$ can compute in parallel without waiting for any other neuron to complete computation.
\end{enumerate}
\end{enumerate}

\section{Details of the DN-2 procedure}
\label{SE:DN-2-Algorithm}

\subsection{Initialization}
Weights are initialized randomly.
There are $k+1$ initial neurons in $Y$ zone, and $V = \{ \dot \v_i  \;|\; i=1, 2, ... , k+1\}$ is the current
synaptic vectors in $Y$.  Whenever the network takes an input $\p$, compute the
pre-responses in $Y$.  If the top-$1$ winner in $Y$ has a pre-response lower than almost perfect match $m(t)$, activate a free neuron to fire. The almost perfect match $m(t)$ is defined as follows:
\begin{equation}
m(t) =  \alpha(t) (1 - \theta)
\label{EQ:match}
\end{equation}
where $\theta$ is the machine zero, and $\alpha(t)$ regulates the speed of neuronal growth, implemented by
a handcrafted lookup table.  This lookup table can be further fine-tuned by an evolutionary algorithm but evolution with full lifelong development is very expensive).

As soon as the new $Y$ neuron is added, every
$Z$ neuron will add a dimension in its synaptic vector in the update of the following time $t$. The dimension of its weight vector continuously increases together with the number of active neurons in the $Y$ zone.

\subsection{Mean-contrast normalization with volume}
The objective is to normalize the sensory input vector into each neuron, so as to disregard the effects such as the average brightness (mean of input) and contrast (deviation from the mean).  However, if the input vector is almost constant across all components, we do not want to boost its noise.  We also keep the volume information in the input so that the network can reliably detect silence without wasting many neurons to detect the noise in silence.   In the following representation, the resulting vector $\p'$ is a unit vector in $R^{d+1}$,

\begin{enumerate}
\item Suppose input $\p= (p_1, p_2, ... , p_d)  \in R^d$.  Compute the mean:
\begin{equation}
\bar{m}=\frac{1}{d} \sum_{i=1}^{d}p_i.
\end{equation}
\item Conduct the zero-mean normalization:
\begin{equation}
 p_i \leftarrow p_i - \bar{m}, i=1, 2, ... , d
\end{equation}
\item Compute the  volume (or contrast) as the Euclidean norm:
\begin{equation}
l = \| \p \|
\end{equation}
\item Unit-length normalization: If $l> \epsilon $ (e.g., 10 times of machine zero) normalize the norm to become a unit vector:
\begin{equation}
\p \leftarrow \frac{1}{l}\p.
\end{equation}
Otherwise, keep $\p$ as a nearly all-zero vector.  Note: such a $\p$ will be represented below by
$\p' = (0, 0, 0, ... , 0, 1)$.
\item Add volume dimension: Increase the dimension of $\p$ by one
which is assigned the value of volume $l$ to become
\begin{equation}
\p' \leftarrow ((1-\alpha)\frac{l}{l_M} \p, \alpha\frac{l_M-l}{l_M})
\end{equation}
where $l_M$ is the maximum volume found so far for this neuron and $\alpha$, $0<\alpha<1$, e.g., $\alpha = 0.2$, is the relative contribution of the relative away from peak volume $(l_M-l)/l_M$.
\item Normalize the norm of $\p'$, but now in $R^{d+1}$:
\begin{equation}
l' = \|  \p' \|
\end{equation}
which is never zero, and
\begin{equation}
\p' \leftarrow \frac{1}{l'}\p'
\end{equation}
\end{enumerate}
Namely, if the volume is high, the information is mainly in the first $d$ components.  When
the volume is low, the information is mainly in the last component.   This should help the
detection of silence without waiting a lot of neurons to memorize many patterns of near silences.

\subsection{Post-Synaptic Maintenance}
In parallel computation, the number of post-synaptic neurons should be restricted, because
each post-synaptic neuron competes for access time.   In biology, each post-synaptic neuron competes for
energy from the pre-synaptic neuron.

Suppose a post-synaptic neuron $i$ has its synaptic weight vector $\v_i=(v_{i1},v_{i2},...v_{id})$ and
input vector $\p_i=(p_{i1},p_{i2},...,p_{id})$, respectively, both unit vectors.

Thus,  $v_{ij}$, $j$-th component of $\v_i$, is the weight value between pre-synaptic neuron $j$ and post-synaptic neuron $i$. $p_{ij}$, $j$-th component of $\p_i$, is the input from neuron $j$ to $i$($j=1,2,...,d$). We use amnesic average of $l_1$-norm deviation of match between $v_{ij}$ and $p_{ij}$ to measure expected uncertainty for each synapse. So it must start with a constant value and wait till
the weight value of the synapse has a good estimate. Each synapse will record the neuron's firing age when the synapse connected or reconnected, stores as their spine time $n_{ij}$. Suppose that $\sigma_{ij}(n_i)$ is the deviation at neuron's firing age $n_i$. The expression is as follows:
\begin{equation}
\sigma_{ij}(n_i) =
 \left\{
\begin{array}{ll}
 \delta /\sqrt{12}  &\mbox {if  $\Delta n \le n_0 $} \\
  w_1(\Delta n_{ij})  \sigma_j(n_i-1)\\
   + w_2(\Delta n_{ij})   | v_{ij}  - p_{ij} | &  \mbox {otherwise }
  \end{array}
\right.
\label{EQ:deviation}
\end{equation}
where $\Delta n_{ij} = n_i-n_{ij}$ is the number of firing this synapse advanced, $w_2(\Delta n_{ij})$ is the learning rate depending on the firing age (counts) of this synapse and $w_1(\Delta n_{ij})$ is the retention rate, $w_1(\Delta n_{ij}) + w_2(\Delta n_{ij})\equiv 1$. $n_0$ is the waiting latency (e.g. $n_0 = 20$). The expected synaptic deviation among all the synapses of a neuron is defined by:
\begin{equation}
\label{EQ:barsigma1}
\bar{\sigma}_i (n_i) = \frac{1}{d}\sum_{j=1}^d  \sigma_{ij}(n_i) .
\end{equation}

We define the relative ratio:
\begin{equation}
\label{EQ:rij}
r_{ij}(n_i)= \frac{\sigma_{ij} (n_i)}{\bar{\sigma}_i(n_i)} .
\end{equation}
And we introduce a smooth synaptogenic factor $f(r ) $ defined as:
\begin{equation}
\label{EQ:fr}
f (r) = \left\{
\begin{array}{ll}
  1  &  \mbox {if  $r < \beta_s(t)  $} \\
  \frac{\beta_b(t)  - r}{\beta_b(t)  - \beta_s(t)} \ &  \mbox {if  $\beta_s(t)  \le r   \le  \beta_b(t) $  } \\
  0 &  \mbox {otherwise}
\end{array}
\right.
\end{equation}
where $\beta_s(t)$ and $\beta_b(t)$ are parameters (can update with step through time) to control the number of synapse with active connection ($f(r)>0$). Suppose the number of synapse with active connection of each neuron is $n_s$, $\beta_s(t)$ and $\beta_b(t)$ can control $n_s$ less than the upper boundary $n_u$ (e.g. $n_u=1000$).

We rank $0.2 n_s$ synapses with 0 factor value ($f(r)=0$) but lower relative ratio value, and put them in the buffer zone. These synapses in buffer zone can be considered as non-active connections, they can keep their weight value and update the deviation next time. But they don't attend the pre-response computation.

When post-synaptic neuron $i$ advances every $n_0$ firing, we find the pre-synaptic neuron $e$ with the most stable connection. Then we find the nearest neighbor neuron $j$ of neuron $e$, and put the synapse $v_{ij}$ enters the buffer zone to replace the synapse with the largest ratio in buffer zone.

The distribution of neurons' locations is described in the algorithm of spatial distribution of neurons. To avoid $c^2$ complexity, we can search neuron $e$'s nearest neighbor by assistance of $8$ glial cells around this neuron. Since each glial cell has recorded $k$ nearest neighbor neurons, we can compare distances between neuron $e$ and these neighbor neurons of $8$ glial cells to find the nearest neighbor neuron $j$.

Then trim the weight vector $\v_i=(v_{i1}, v_{i2}, ... , v_{id})$ to be
\begin{equation}
v_{ij} \leftarrow f (r_{ij}) v_{ij}
\label{EQ:factoredInnerProduct}
\end{equation}
$j=1, 2, ... , d$. Similarly, trim the input vector $\p_i=(p_{i1}, p_{i2}, ... , p_{id})$.

The $\epsilon$-mean normalization and the unit-length normalization of {\em both} the weight vector $\v_i$ and the input vector $\p_i$ must be done {\em both} before and after the trimming.  Namely, four times.

\subsection{Pre-Synaptic Maintenance}
Pre-synaptic maintenance is for the pre-synaptic neurons, suppose pre-synaptic neuron $i$ connect to $d$ post-synaptic neurons, and we use the deviation $\sigma_{ij}(n_i)$ which is between $v_{ij}$ ($v_{ij}$, $i$-th component of $\v_j$, is the synaptic weight value between neuron $i$ and $j$) and $p_j$ ($p_{ij}$ is the input from $i$ to $j$) at age $n_i$ to calculate its expected synaptic deviation. The equation is same as Eq.~\eqref{EQ:barsigma1}.

Then we can calculate the relative ratio using Eq.~\eqref{EQ:rij}, calculate synaptogenic factor using Eq.~\eqref{EQ:fr} and trim synapses.

To keep consistency of the connection of post-synaptic neuron $i$ and pre-synaptic neuron $j$, only both post-synapse and pre-synapse are with active connection we consider the synapse is with active connection. If one of these two synapses is in buffer zone, we put another one into buffer zone.
\subsection{Prescreening}
Many neural studies have found that cortical areas have laminar architectures and unambiguous connection rules. Prescreening seems a consequence of these rules.

Prescreening is necessary because three component matches, the bottom-up, top-down and lateral pre-responses are not clean logic values. The prescreening can avoid hallucination that is caused 
a bad match in one component (e.g., in $X$) is excessively compensated by good matches in other components.  In other words, a bad match in one component is screened out before the three-component integration. 

We do prescreening for the bottom-up, top-down and lateral pre-responses in parallel. For $Y$ neuron $i$, both its weight vector and input are consist of bottom-up, top-down, and lateral parts ($\p_i=(\x_i,\y_i,\z_i)$, $\v_i=(\v_{b,i},\v_{t,i},\v_{l,i})$). We denote $\dot {\v}$ as the vector of $\v$ with a unit Euclidean norm: $\dot {\v} = \v /\| \v \|$. So we can compute bottom-up pre-response $r'_{b,i}$ as:
\begin{equation}
r'_{b,i} = \dot{\v}_{b,i} \cdot \dot{\x}_i.
\label{EQ:bottomup}
\end{equation}
Top-down pre-response $r'_{t,i}$ and lateral pre-responses $r'_{l,i}$ are computed in similar way.

Then we rank top $k_1$ bottom-up pre-responses and find the $k_1$-th pre-response to define it as $r'_B$ ($k_1$ is proportional to the number of $Y$ neurons). Similarly we rank top $k_2$ top-down pre-responses and define the $r'_T$, also rank top $k_3$ lateral pre-responses and define the $r'_L$. After that, we build the bottom-up prescreening set for the neurons with top pre-responses. We define the bottom-up prescreening set $S_b$ to include the neurons with bottom-up pre-response among top $k_1$ ones as follows:
\begin{equation}
S_b = \{ i \mid r'_{b,i} \geqslant r'_B \}.
\label{EQ:bottomup set}
\end{equation}
Top-down prescreening set $S_t = \{ i \mid r'_{t,i} \geqslant r'_T \}$  and lateral prescreening set $S_l = \{ i \mid r'_{l,i} \geqslant r'_L \}$ are built in the same way.

Finally we can define the prescreening set $S_p$ including the neurons passed all prescreening process.
\begin{equation}
S_p = S_b \cap S_t \cap S_l
\label{EQ:final set}
\end{equation}

If the number of neurons in $S_p$ is less than $k$ ($k$ is the number of winners), we need to lower the standard--dropping the prescreening of top-down pre-responses. We drop prescreening of top-down pre-responses firstly because top-down pre-response is most clean among the three. Then the set $S_p$ is changed to $S_p = S_b \cap  S_l$. If the number of neurons in $S_p$ is still less than $k$, we have to drop the prescreening of lateral pre-responses, and the set $S_p$ is changed to $S_p = S_b $.
\subsection{Response Computation and Competition}
The zone function $f_y$ and $f_z$ are based on the theory of Lobe Component Analysis (LCA), a model for self-organization by a neural area.

Each $Y$ neuron in $S_p$ can calculate its pre-response. For neuron $i$, its pre-response $\r'_i$ is calculated as:
\begin{equation}
r'_i= r'_{b,i} + r'_{t,i}+ r'_{l,i}.
\end{equation}

To simulate inhibitions within $Y$, we define dynamic competition set for each neuron. Only this neuron is among top-k winners in its dynamic competition set, it can fire.

We find the rank of neuron $i$'s pre-response in its dynamic competition set $Y_i$ (The definition is in next section).
\begin{equation}
\rho(i)=\mbox{rank}\max_{j \in Y_i}(r'_j)
\end{equation}
And only rank the top-k pre-responses: $r'_{1,i} \geq r'_{2,i}...\geq r'_{k+1,i}$.

If neuron $i$ is among top-k winners, we scale its response value $y_i$ in $(0, 1]$:
\begin{equation}
y_i=\frac{r'_{\rho(i),i}-r'_{k+1,i}}{r'_{1,i}-r'_{k+1,i}}.
\end{equation}
Otherwise, $y_i = 0$. After all $Y$ neurons compute their responses in parallel, we obtain $\y'$.

The $Z$ zone computes its response $\z'$ using above method similarly.

Finally we do the replacement: $\y \leftarrow \y'$, $\z \leftarrow \z'$.
\subsection{Hebbian Learning of excitation}
The connections in DN-2 are learned incrementally
based on Hebbian learning --- cofiring of the pre-synaptic activity and the post-synaptic activity of the firing neuron. When neuron $i$ fires, its firing age is incremented $n_i \leftarrow n_i+1$ and each component $v_{ij}$ of
its weight vector $\v_i$ is updated by a Hebbian-like mechanism. If $j$ didn't connect with $i$ before, $v_{ij}$ is updated as:
\begin{equation}
  v_{ij} \leftarrow y_i \dot{p}_{ij} .
\label{EQ:learn1.1}
\end{equation}
Otherwise, $v_{ij}$ is updated as:
\begin{equation}
  v_{ij} \leftarrow w_1(n_i) v_{ij} + w_2(n_i) y_i \dot{p}_{ij}.
\label{EQ:learn1.2}
\end{equation}
where $\dot{p}_{ij}$ is $j$-th component of input vector $\dot{\p}_i$.

The firing neurons also update their deviation between weights and inputs using Eq.~\eqref{EQ:deviation} for synaptic maintenance.
\subsection{Hebbian Learning of inhibition}
For the inhibition connections, if and only if neuron $i$ doesn't fire, its negative neuron $-i$ fires. The negative neuron's firing age is incremented $n_{-i} \leftarrow n_{-i}+1$, and it updates the weight vector. If $j$ didn't connect with $-i$ before, $v_{-ij}$ is updated as:
\begin{equation}
v_{-ij} \leftarrow \dot{p}_{ij}.
\label{EQ:learn2.1}
\end{equation}
Otherwise, $v_{-ij}$ is updated as:
\begin{equation}
v_{-ij} \leftarrow w'_1(n_{-i}) v_{-ij} + w'_2(n_{-i})\dot{p}_{ij}.
\label{EQ:learn2.2}
\end{equation}
where $w'_2(n_{-i})$ is the learning rate depending on the un-firing age (counts) $n_{-i}$ of this neuron and $w'_1(n_{-i})$ is the retention rate, $w'_1(n_{-i}) + w'_2(n_{-i})\equiv 1$.

We use the current response for $\dot{p}_{ij}$ above because it is available after the current top-$k$ competition.

Then we calculate the average value $\bar{v}_{-i}$ of all components in $\v_{-i}$.
\begin{equation}
\bar{v}_{-i}=\frac{1}{c}\sum_{j=1}^c  v_{-ij}
\end{equation}
where $c$ is the number of neurons in the input field of neuron $i$.

We define the dynamic competition set of neuron $i$ as follows:
\begin{equation}
Y_i=\{ j \mid v_{-ij}- \bar{v}_{-i} >0 \}
\end{equation}
where $j$ is in the input field of neuron $-i$.
\subsection{Splitting}
When the number of active neurons in $Y$ zone is less than $n_y$ and the top-1 winner neuron $i$ has a pre-response lower than almost perfect match $m(t)$, we do splitting for the neuron $i$. The splitting takes the following steps:
\begin{enumerate}
    \item Increase the number of neurons in $Y$, $c_y \leftarrow c_y+1$.
	\item Create a neuron $j$. The new neuronal vectors is copied from neuron $i$.

    $\v_j \leftarrow \v_i$, $\v_{-j} \leftarrow \v_{-i}$

    \item Set new neurons' firing age: $n_j \leftarrow 0$.

    Also set its negative neuron's firing ages: $n_{-j} \leftarrow 0$.

    \item Supervise neuron $j$ to fire at $1$ and use Eq.~\eqref{EQ:learn1.1} to update weight.
    \item Set neuron $j$'s location to be near around the neuron $i$, only with a distance of $5 \theta$ (e.g. set the location to be $(l_{h,i},l_{v,i}+5 \theta,l_{d,i})$, $(l_{h,i},l_{v,i},l_{d,i})$ is the location of neuron $i$).
\end{enumerate}

\subsection{Spatial distribution of neurons}
We calculate the distribution of the neurons after the DN-2 advances every $n_{dn}$ updates (e.g. $n_{dn}=50$).
Suppose there are $n_g$ glial cells distributed evenly in the $3D$ space inside the skull as well as $c$ neurons.

\begin{algorithm}[h]
\caption{Updating the locations of neurons}
\SetAlgoLined
\For{$i=1,2,3,...,n_g$}{
    Glial  cell $i$ with location $\g_i=(h_i,v_i,d_i)^{\top}$ finds its $k$ nearest neurons at location $\l_j$ ($j=1,2,...,k$), and pulls these neurons with a force vector\;
    \For{$j=1,2,3,...,k$}{
        Neuron $j$ with location $\l_j=(l_{h,j},l_{v,j},l_{d,j})^{\top}$ in the $k$ nearest neurons generates the force vector $\f_{ij}$.
        \begin{equation}
            \f_{ij} = \g_i - \l_j
        \end{equation}
    }
}
\For{$m=1,2,3...,c$}{
    Suppose neuron $m$ has got $n_{m^{(t)}}$ glial cells that pull it, and mark them as glial cells $1,2,...,n_{m^{(t)}}$. Then calculate each neuron's force vector:
    \begin{equation}
        \f_m = \frac{1}{n_{m^{(t)}}}\sum_{i=1}^{n_{m^{(t)}}} \f_{i,m}
    \end{equation}
    Update each neuron's location:
    \begin{equation}
        \l_m \leftarrow (1-\gamma)\l_m + \gamma \f_m
    \end{equation}
    where $\gamma$ is the learning rate, e.g. $\gamma=0.1$.
}
\end{algorithm}

\subsection{Patterning}
Patterning is about the initial input sources of $Y$ neurons in terms of zones $Z$, $Y$ and $X$.  There are 7 types of initial patterning,
named as a binary number $s_xs_ys_z$ where $s_x, s_y$ and $s_z$ are either 0 or 1.
The following table gives 7 types of $Y$ neurons:
\begin{table}[h]
\caption{Seven type of $Y$ neurons}
\begin{center}
\begin{tabular}{|c|ccc|}
\hline
Type & Z & Y & X \\
\hline
001 & No & No & Yes \\
010   & No & Yes & No \\
011   & No & Yes   & Yes \\
100  & Yes & No   & No \\
101  & Yes & No  & Yes \\
110  & Yes & Yes & No \\
111  & Yes & Yes & Yes \\
\hline
\end{tabular}
\end{center}
\label{default}
\end{table}%

Each source corresponds to a separate vector-length normalization.
For example, for type 111 $Y$ neurons, each input zone is normalized
to become a unit vector.   Then, the maximum pre-response is $(1+1+1)/3=1$.
As another example, for type 011 $Y$ neurons, the maximum pre-response is
$(1+1)/2=1$ .  Therefore, the maximum response of a $Y$ neuron
is 1, regardless the type.

The patterning in $X$ and $Z$ means the initial zoning for sensors and effectors.
Each sensor or effect corresponds to a zone.   For example, if the system has
two microphones, two cameras, and one skin, it has 5 zones in $X$.

 If the system
has 100 concept zones in $Z$, each concept zone  has $1+11$ neurons, where the
first neuron means ``none'' and other neurons represent 0, 1, 2, ... , 10, respectively then the $Z$ zone has 100 zones
and a total of 100*12 = 1200 neurons.

	\section{Maximum Likelihood Optimality of DN-2}
	\label{SE:optimality}

	\subsection{Review of the three theorems in DN-1}
	In order the understand the property of DN-2, we need to look back at the three important properties of DN-1 proved in \cite{weng2015brain}:
	\begin{enumerate}
		\item With enough neurons, DN-1 learns any FA incrementally error-free by observing the transitions in the target FA for only once with two updates of the network, with supervision in $X$ and $Z$.
		\item When frozen, DN-1 generates responses in $Z$ with maximum likelihood estimation, conditioned on the last state of the network.
		\item Under limited resources, DN ``thinks" (i.e., learns and generalizes) recursively and optimally in the sense of maximum likelihood.
	\end{enumerate}

	The proof in \cite{weng2015brain} is under two assumptions, which are no longer satisfied in the context of DN-2:
	\begin{enumerate}
		\item Only type 101 $Y$ neurons. DN-1 only uses $Y$ neurons with bottom-up and top-down connections. However, DN-2 now has seven types of neurons, as shown in Fig. \ref{fig:dn2theorem}.  Notation of the figure: Blue lines: top-down connection from $Z$. Red lines: lateral connection from $Y$. Green lines: bottom-up connection from $X$.
		\item Global top-1 competition among $Y$ neurons. Only one $Y$ neuron in DN-1 is firing at any given time, thus learning the exact $X$ pattern with a specific $Z$ pattern associated to this $Y$ neuron. However, DN-2 has several $Y$ neurons firing at any given time.
	\end{enumerate}

	Thus the assumptions for DN-1 optimality are no longer available, requiring a new way to prove the optimality of learning in DN-2.

	\subsection{Definition of DN-2}
	A DN-2 network at time $t$ can defined as:
	\begin{eqnarray}
		N(t) &=& \{\theta(t), D(t), \Gamma\}
	\end{eqnarray}
	Where $\Gamma$ denotes the hand-picked hyper parameters for the network, $D(t)$ denotes the long-term statistics inside the network. Parameters in $D(t)$ are not part of optimization. $\theta(t)$ contains the weights that are part of optimization at time $t$.

	\begin{eqnarray}
		\Gamma    &=& \{G, k, l_{\textnormal{in}}, l_{\textnormal{out}}, n_y, n_z\} \nonumber\\
		D(t)      &=& \{ \{g | g\in Y, Z\}, \{\bar{g}(t) | g \in Y, Z\}, a(t), L(t)\} \nonumber\\
		\theta(t) &=& \{W(t)\} \nonumber
	\end{eqnarray}

In the above expressions, $G$ is the growth rate table. $k$ is the top-$k$ parameter for competition. $l_{\textnormal{in}}$ is the limit on the number of connections to a specific neuron. $l_{\textnormal{out}}$ is the limit on the number of connections from a neuron to other neurons. $n_y$ is the maximum number of neurons in $Y$ zone. $n_z$ is the maximum number of neurons in $Z$ zone. ${g}$ is the set of neurons in $Y$ and $Z$ zone. $\bar{g}(t)$ is the inhibition field of the specific neuron at time $t$. $L(t) = \{L(g, t)| g\in Y, Z\}$ is set of receptive field (line subspace) for each neuron at time $t$. $a(t) = \{a(g, t) | g \in Y, Z\}$ is the set of connection ages for each neuron at time $t$. $W(t) = \{W(g, t) | g \in Y, Z\}$ is set of weights for each neuron at time $t$.

	\subsection{Conditions on DN-2 learning}
	During learning, DN-2 is constrained by the following conditions, denoted as $ \textbf{C} = \{C_i \;|\; i = 1,2,3\}$.
	\begin{enumerate}
		\item[$C_1$:] Incremental learning.  The network never stores training data. It updates
		its network using input from the environment.  As soon as the update is done, it 
		discards the input before taking the next input from the environment which may depend on
		its current output (e.g,, action). 
		\item[$C_2$:] Skull-closed learning. Human teachers can only supervise the motor port 
		and sensory port of the network once the network starts learning.
		\item[$C_3$:] Limited resources. The hand-picked hyper-parameter $\Gamma$ (which limits the number of neurons used by the network) remains unchanged during learning, simulating the
		limited resources of a machine ``species'' but avoiding the high cost of evolutionary algorithms.
	\end{enumerate}

	\subsection{Lemma 1: DN-2 optimizes its weight under maximum likelihood for each update, conditioned on $ \textbf{C}$.}
	\begin{lemma}
		\label{th: learns MLE}
		Define $\p(t + 1) = \{\x(t), \y(t), \z(t)\}$ ($\p$ is thus a set of all responses in the three zones). At time $t + 1$, DN incrementally adapts its parameter $\theta$ as the Maximum Likelihood (ML) estimator for input in $X$, $Y$ and $Z$, based on its learning experience with limited resources $\Gamma$ :
		\begin{equation}
			\theta(t + 1) = \max_{\theta}f\{\p(t + 1) | N(t), \textbf{C}\}, t \geq 0
		\end{equation}
		The probability density $f\{\p(t + 1) | N(t)\}$ is the probability density of the new observation $\p(t+1)$, conditioned on the last status of the network $N(t)$, based on the network's learning experience.
	\end{lemma}

	Proof for Lemma \ref{th: learns MLE} is attached in the appendix. Compared to other optimization theories on machine learning, the proof of this lemma has the following features:
	\begin{enumerate}
		\item We don't estimate variance or covariance in high-dimensional space as those estimations are usually expensive and computationally costly. In the DN framework, NE and ACH (introduced in \cite{fish2013novelty}) are rough approximates of variances as they are used to detect novelty in the new pattern.
		\item Compared to mixed Gaussian models we use tessellation in high dimensional spaces. The clusters in our model are non-parametric, in the sense that the parameters (connection weight and range) are dynamic, instead of a fixed set.
	\end{enumerate}

	Lemma \ref{th: learns MLE} states that each time DN-2 is providing the best estimate. Inside the `skull` neurons' weights are updated optimally based on $\mathbf{C}$. But we do not guarantee that the external environment would provide the optimal teaching schedule.

	By using this lemma recurrently and we can have the following theorem:



	\subsection{Theorem: DN-2 learns optimally under maximum likelihood from its incremental learning experience, conditioned on $ \textbf{C}$.}
	\label{sec: optimal incremental learning}
	\begin{theorem}
		\label{th:optimal_learning}
		Define $ \cX_{t_0}^t = \{\x(t_0), \x(t_0 + 1), ... \x(t-1) \}, \cZ_0^{t_0} = \{\z(t_0), \z(t_0+1), ... \z(t-1)\}$. At time $t+1$ DN-2 adapts its parameter as the ML estimator for the current $\p(t+1) = \{\x(t), \z(t)\}$, conditioned on the sensory experience $\cX_0^t$ and $\cZ_0^t$ with limited resources $\Gamma$:
		\begin{equation}
			\label{eq: MLE entire sequence}
			\theta(t+1) = \max_{\theta}f'\{\p(t+1) | \cX_0^t, \cZ_0^t, \textbf{C}\}, t\geq 0
		\end{equation}
		The probability density $f'\{\p(t+1) | \cX_0^t, \cZ_0^t, \Gamma\}$ is the probability density of the new observation $\p(t+1)$, conditioned on the entire sensory and motor experience and the pre-defined hyper-parameters.
	\end{theorem}

	Proof for theorem \ref{th:optimal_learning} is attached in the Appendix.

	This theorem is important as it shows that a DN-2 equipped agent behaves in a maximal likelihood fashion while learns incrementally and immediately without the need to store batch data or iterate through training data for multiple times.

	Moreover, with ML firing patterns in each zone, DN-2 is generalizing its learned ``piece-meal'' knowledge taught by individual teachers at different times to many other similar settings (e.g. infinitely many possible navigation sequences which contains a traffic light). Any DN-2 can do such transfers automatically because of the brain-inspired architecture of the DN. DN-1 behaves poorly in this aspect as its learning is based on exact matching of sensory and motor inputs. Prior neural networks and any conventional databases cannot do that, regardless how much memory they have.
	
	In the following sections, we present experiments for three very different tasks and modalities, vision, planning, and audition, respectively.

	\section{Vision-Based Real-world Navigation}
	\label{SE:Vision}
	
	We tested DN-2 with two modalities: audition and vision. A manuscript about the audition experiments is currently under review. In this invention, we present our experimental results with DN in real-world navigation using real-time video inputs. Here we show the performance of the network with the visualization of its learned representation.

	Vision-guided autonomous navigation is hard for the following reasons:
	\begin{enumerate}
		\item GPS is often missing and inconsistent. GPS directions may conflict with local information during navigation (e.g., obstacle, detour, and wrong facing direction).
		\item Much of the sensory information is irrelevant. Most features in the sensory input are not directly related to the current move. During navigation, we are interested in traffic signs, obstacles, lane markings, while the input image usually contains large chunks of backgrounds irrelevant to the current navigation movement.
		\item Much of the information in context is irrelevant. Action as context is very rich, some are related to the next move while some are not. If use all then the complexity of motion is high because the complexity may not be observed and the generalization power is weak. Context attention is important (only pick up context that is relevant to the next action).
	\end{enumerate}
	In short, we are facing the need for a general learning framework although the current setting is real-time navigation. This is not a simple task since there are many muddy rules that cannot be handcrafted sufficiently well by a human designer.

\subsection{Experiment set up}

	\begin{figure}[tbh]
		\centering
		\includegraphics[width=1.0\linewidth]{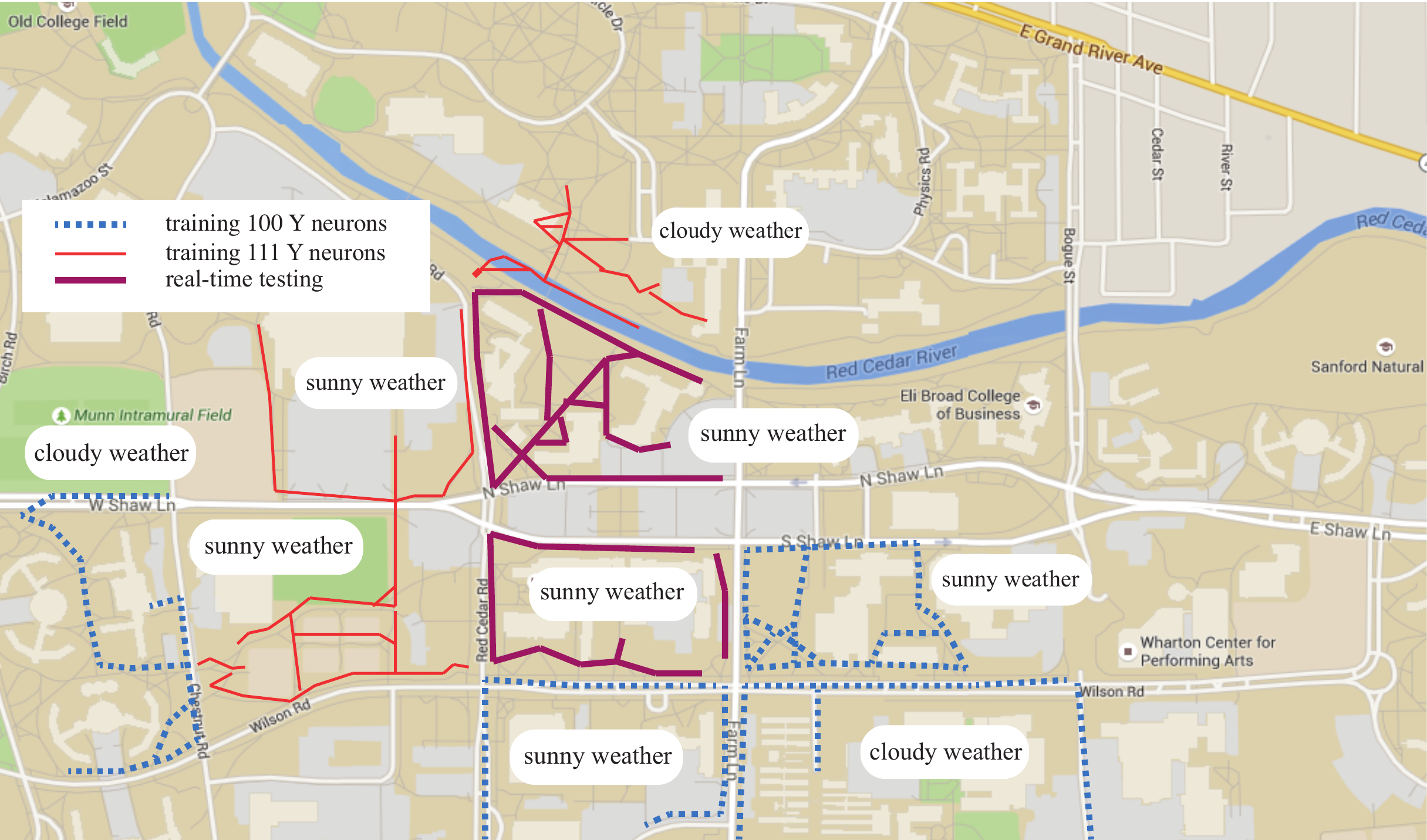}
		\caption{}
		\label{fig:navigationtrainingroute}
	\end{figure}

		The network is trained around the campus of the university to learn the task of autonomous navigation on the walk side. Fig. \ref{fig:navigationtrainingroute} provides an illustration of the extensiveness of training and testing.  The inputs to the DN were from the same mobile phone that performs computation, including the stereo image from a separate camera and the GPS signals from the Google Directions interface. The outputs of the system include heading direction or stop, the location of the attention, and the type of the object at the attended location (which detects a landmark), and the scale of attention.  Disjoint testing sessions were conducted along paths that the machine has not learned.			

	\begin{figure*} [tbh]
		\centering
		\includegraphics[width=0.9\linewidth]{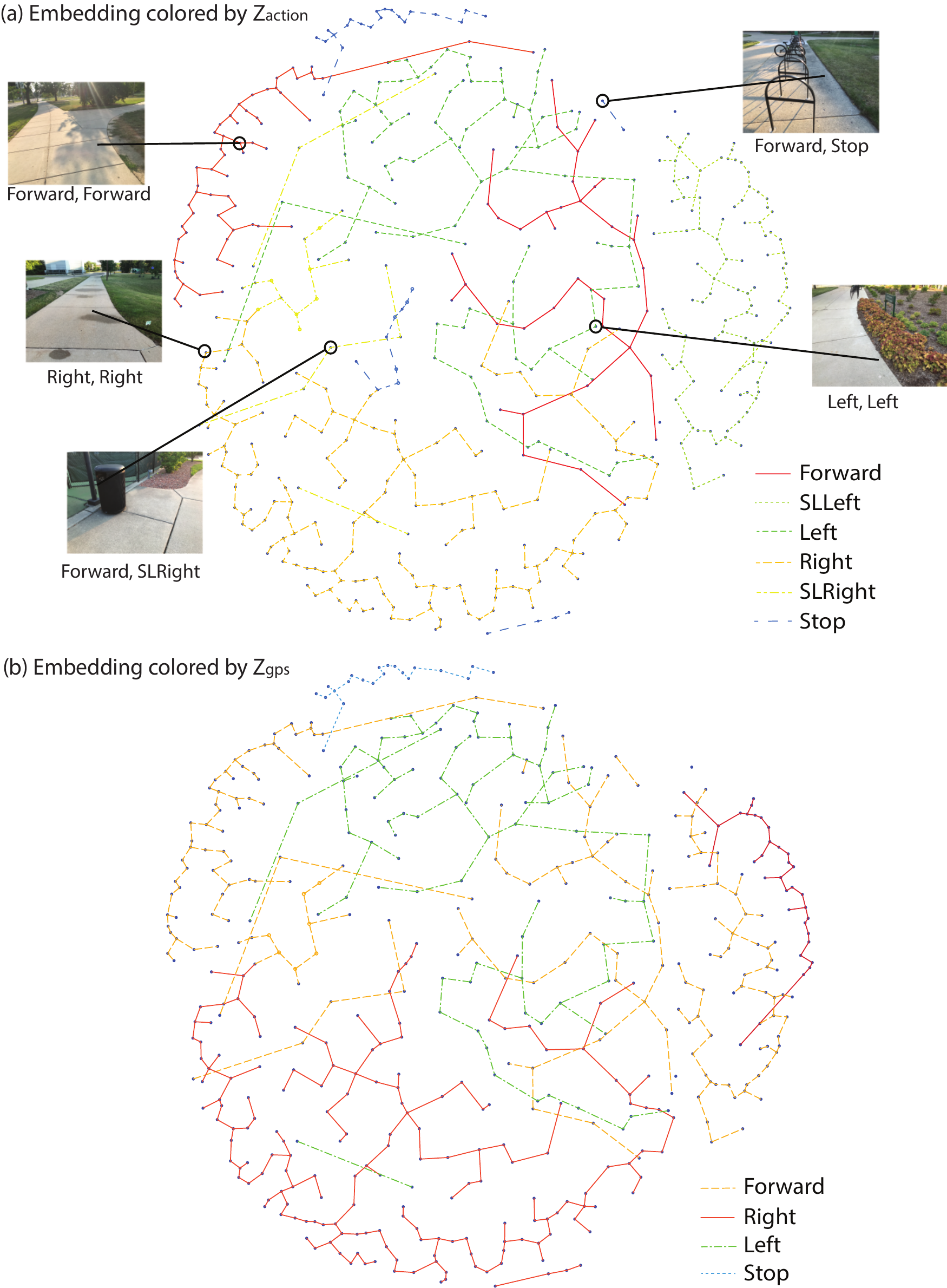}
		\caption{}
		\label{fig:circlelinks}
	\end{figure*}

	The wide variety of real-world visual scenes implied by the extensive routes in Fig. \ref{fig:navigationtrainingroute} presented great and rich challenges to this camera-only system without using any laser device. DN-2 uses multiple types of neurons to form robust representations about the road edges and obstacles as discussed in Sec.\ref{sec: robust edges}.   As illustrated in Fig.~\ref{fig:navigationtrainingroute}, the network is generated by training a DN-2 agent during real-time navigation experiment with supervised $Z_\textnormal{action}, Z\textnormal{GPS} $. In each figure, blue dots represent type 111 Y neurons in the network for outdoor navigation. Connection between Y neurons indicate that these neurons are laterally connected (when a neuron is connected to multiple hidden neurons, we choose one with the strongest connection). Left figure: Type 111 $Y$ neuron embedding colored according to action motor ($Z_\textnormal{action}$). Example of navigation data triggering specific neurons firing is plotted in this figure. SLLeft is short for slightly left and SLRight is short for slightly right. Right figure: Type 111 $Y$ neuron embedding colored according to GPS motor ($Z_\textnormal{GPS}$).

	\subsection{Training and testing}
	To illustrate the power of DN-2 we set the growth hormone of DN-2 to use only two types of neurons: type 100 for low-level image feature extraction and type 111 for high-level robust representations. The growth hormone is set in a way such that type 111 neurons would start learning only after most type 100 neurons have fired extensively (forming stable weights as feature extractors).

	Testing is performed with the network frozen (weights not updating but the neurons are still generating responses). As shown in Fig. \ref{fig:navigationtrainingroute} the testing routes are novel settings to the learned network, but with similar obstacles, roads, and bushes compared to the training settings.

	Details about training can be found in Table \ref{table: real-time training testing detail}.  The performance of the network is summarized in Table \ref{table: testing details}. 
	
	The performance is evaluated using two different metrics: different from user's intention (diff) and absolution errors (error). The difference is that in a `diff' situation the network still can recover from the movement in subsequent frames, while the absolute errors are defined as the situations where the network gets stuck into an unrecoverable action (e.g. stop and not moving or bump into obstacles).

	As shown in Table \ref{table: testing details}, the most errors we got are from untrained obstacles (e.g. bicycles in the middle of the lane, or pedestrians on skateboards). This can be resolved by more extensive training and a larger network.

	\subsection{Visualization of learned weights}
	
	\subsubsection{Lateral weights from type 100 neurons to type 111 neurons}
	
	Fig. \ref{fig:lateralweights} and Fig. \ref{fig:lateralweightsmasked-copy} show the projected lateral weights of high-level 111 neurons in the trained network.  Each sub-figure shows the $Y$ to $Y$ connection of a 111 neuron, with the connected 100 neurons' bottom-up weights projected into the image space. Each subfigure is equivalent to showing only the red receptive fields for type 111 neurons in Fig.\ref{fig:smallexample}.

	Without synaptic maintenance in Fig. \ref{fig:lateralweights}, Type 111 neurons with low firing ages forms evenly distributed attention due to the local competition zones of lower level 100 neurons.  
	
	After synaptic maintenance in Fig. \ref{fig:lateralweightsmasked-copy}, the high-level neurons focuses attention on consistent road edges and become invariant to the changes in the highly variant shadow shapes. The connections from type 100 neurons to type 111 neurons shits towards these consistent features, while the highly unstable connections are cut from the range of connection.   Synaptic maintenance cuts away the unstable connections with high variances. As shown in the visualization, these neurons focus their attention on road edges and become invariant to the changes in monotone shadows. This figure shows that the network performed synaptic maintenance as we expected in Sec. \ref{sec: robust edges} and Fig. \ref{fig:smallexample}. The trimmed lateral connections from type 100 neurons to type 111 neurons are exactly as the kept stable connections shown in Fig. \ref{fig:smallexample}.

	This visualization proves that using multiple types of neurons combined with lateral connection (major novelties in Dn-2, see Sec. \ref{sec:Multiple types} and \ref{sec:fluid}), we can from robust hierarchies of representation with the internal features.

	\subsubsection{Lateral weights among type 111 neurons with contextual embedding}
	The lateral connections among high-level 111 neurons are shown in Fig. \ref{fig:circlelinks}.  The network is generated by training a DN-2 agent during real-time navigation experiment with supervised $Z_\textnormal{action}, Z\textnormal{GPS} $. In each figure, blue dots represent type 111 Y neurons in the network for outdoor navigation. Connection between Y neurons indicate that these neurons are laterally connected (when a neuron is connected to multiple hidden neurons, we choose one with the strongest connection). Left figure: Type 111 $Y$ neuron embedding colored according to action motor ($Z_\textnormal{action}$). Example of navigation data triggering specific neurons firing is plotted in this figure. SLLeft is short for slightly left and SLRight is short for slightly right. Right figure: Type 111 $Y$ neuron embedding colored according to GPS motor ($Z_\textnormal{gps}$).  These connections form temporal relationships among the high-level abstractions.  
	As shown in Fig. \ref{fig:circlelinks}, a formed representation learns the transition among different navigation rules based on the current image and the context of navigation (top-down input from the $Z$ zone). The plotted lateral connections are colored with the navigation context information (top-down inputs from $Z$), grouping sequentially firing neurons closer together.

	These learned temporal rules are the key to DN-2's low error rate in real-world navigation. Compared to DN-1 where the agent was just a pure image classifier (see \cite{zhengj2016mobile}), DN-2 uses these learned temporal rules to perform sequential navigation tasks (e.g. constant winding road navigation in segment 17 and 18, table \ref{table: testing details}, avoiding obstacles and facing direction corrections).

	It seems impractical to hire humans to effectively translate the numerical rules represented by the DN-2 in Fig. 2, because those rules are too muddy and too many. This also shows that DN-2 goes beyond simple image classifiers (compared to feed-forward neural networks like convolutional neural networks), as the hierarchical representation formed in DN-2 is both temporal and spatial.


	\begin{figure}[tbh]
		\centering
		\includegraphics[width=1.0\linewidth]{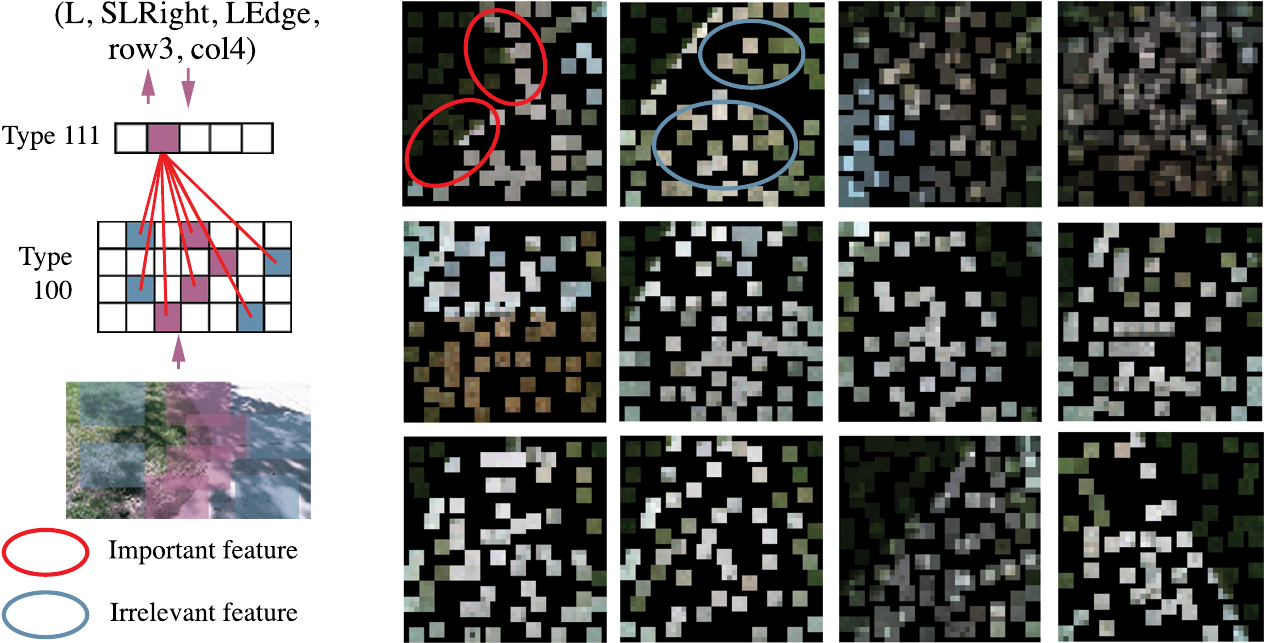}
		\caption{}
		\label{fig:lateralweights}
	\end{figure}

	\begin{figure}[tbh]
		\centering
		\includegraphics[width=1.0\linewidth]{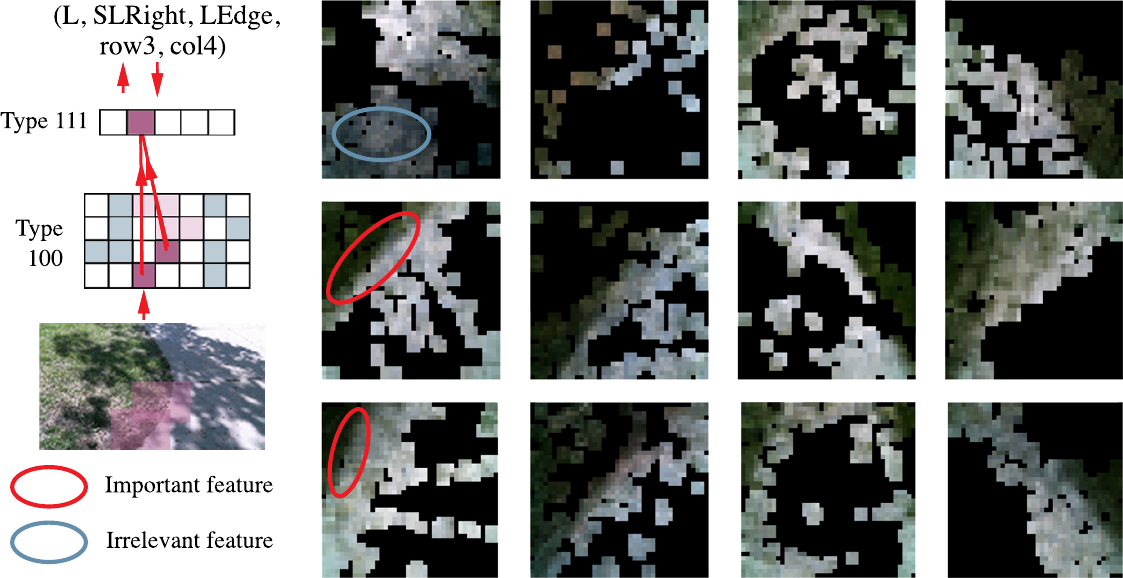}
		\caption{}
		\label{fig:lateralweightsmasked-copy}
	\end{figure}


\section{Universal Turing Machine and Developmental Network-2}
\label{sec: UTM and DN-2}
A lot of the real-world traffic lights can be broken down into state transitions. For example, when in the state of ``moving forward'', an agent should take action ``stop'' if the input is recognized as ``obstacle''. These rules can be formulated as a Finite Automaton: $q(t)\xrightarrow{\sigma(t)}q(t+1)$, where $q(t)$ is the state of the agent at time $t$, and $\sigma(t)$ is the input observed by the agent at time $t$. However, in real-world navigation we not only act according to the input, but we are also changing the environment. E.g., we are actively changing the signal of the traffic light when we are pressing the wait button.

Turing machine \cite{turing1937computable, hopcroft2006automata, martin1991introduction} is a better computation model in this sense as it offers an additional read-write head that allows the agent to alter the input tape. The input tape in our case is the environment. Following the definition of Turing Machine, we can thus formulate a navigation sequence (e.g., navigating from the university library to my apartment) as $T=(Q, \Sigma, \Gamma, q_0, \delta)$, where $Q$ is the set of states (navigation states as in ``moving forward'', ``turning left'', ``arriving''), $\Sigma$ is the input (current input images), $\Gamma$ is the tape alphabet (all possible images), $q_0$ is the initial state (``start navigation''), and $\delta$ is the transition function:
\begin{equation}
\label{equation: turing machine}
\delta:  Q \times \Gamma \rightarrow Q \times \Gamma \times \{R, L, S\}
\end{equation}
where \{R, L, S\} are the head motion right, left and stationary in the context of Turing Machine, but can be redefined and expanded to the actions taken by the agent to alter the environment. 

Our DN-2 aims to incrementally learn these individual Turing Machines of single navigation segments. Thus our DN-2 is a universal Turing Machine (UTM) \cite{hopcroft2006automata, martin1991introduction}. 

\subsubsection{Definition of UTM} A Universal Turing machine (UTM) simulates the behavior of any TM, given the TM and the input is encoded onto the input tape of the UTM. Formally, an UTM receives an input string in the form of $e(T)e(x)$ on the input tape, where $e$ is the encoding function, $T$ is the targeted TM and $x$ is the input. It simulates the computation of $T$ on data $x$, and output $e(z)$, where $z$ is the output of $T$ on data $x$. The UTM does not really know what the meaning of $z$ is \cite{martin1991introduction}. 

\subsubsection{Differences between UTM and DN-2} The encoding function under the UTM framework is hand-crafted. There are many possible ways to build such an encoding function, which is designed to represent the symbolic transition table. This approach is not desirable for our ETM to simulate how the brain works.  DN-2 does not have this encoding function because it does not use symbolic representation. DN-2 uses natural input as patterns in $X$. It also uses emergent state as patterns in $Z$. The actual encoding is the transformation from the $\{X, Z\}$ patterns to the neuronal weights in the network. This transformation is not hand-crafted encoding, but rather the result of competition based on the biologically inspired mechanisms of DN such as Hebbian Learning in Lobe Component Analysis and dynamic inhibition regions.

To summarize in computational terms, UTM performs using encoding function. It searches current input $e(x)$ in $e(T)$, and produces corresponding state $e(q)$ and movement $e(z)$.
DN performs using emergent patterns in $X, Y$ and $Z$. It searches the entire learning experience embedded in all weights of the network. After competition, the corresponding states and movement emerge in $Z$.

\subsubsection{Universal learning in DN-2}
In this section we explain how DN-2 simulates any TM through its learning experience by answering the following three questions:

How does DN-2 get any programs $T$? It learns from lifetime from simple to complex.  Earlier simple skills facilitate learning of later more complex skills. In the case of navigation, learning where and what of landmarks helps to learn to navigate in a new setting with different appearance at different locations. 

How does DN-2 get data $x$? Using learned attention in a grounded way from the real environment through its sensors.  UTM cannot read all data but DN-2 potentially can depending on the maturity of the skill set it has learned. By maturity we mean the power of generalization. E.g., at early age, visual appearance of river will cause the agent to stop. At later age, the same appearance will trigger the agent to find a boat/bridge, or even build a bridge/boat to cross the river. 

How does DN-2 tell the TM part $T$ from data part $x$? There is no fixed or static division between rules $T$ and data $x$. Earlier sensorimotor experiences tend to be considered as data $x$ for the agent to recognize and learn their rules. Later sensorimotor experiences enable the learning agent to get rules very quickly with very few examples because the rules are in abstract forms in such experiences. E.g., when reading a book by an adult the rules stated in a textbook are data that can be translated immediately by the adult reader. Here data and rules are indistinguishable because the adult reader trusts the textbook.

\section{Planning and Task Chaining}
    \label{SE:Planning}
    In this section, we report for long sequential tasks, such as planning and task-chaining. This follows the discussion of Universal Turing Machine as TM is essential for the success of an autonomous navigation system.

\subsection{Autonomous navigation needs Emergent Turing Machine}
\label{sec:DN_UTM}
Here we apply our DN-2 theory to a real-world scenario: autonomous navigation. Autonomous navigation requires general purpose learning like DN-2 instead of hand-craft rules and feature detectors for the following two reasons:
\begin{enumerate}
	\item Dynamic environment. In a real-world setting we cannot anticipate the kind of landmarks, concepts and context needed for unknown driving environment.  Internal representations and navigation contexts to emerge on the fly. 
	\item Complex traffic rules. The sheer number of rules for a navigation FA is too large to enumerate. These rules also interact with each other and the interaction is impossible to hand-craft. 
\end{enumerate}

How does DN-2 deal with these challenges? In DN-2, the interactions among different rules are handled by lateral connections. The representation is emergent from the context. Concepts and contexts are generated `on demand`. I.e., if the agent is familiar with the current navigation setting (internal firing value close to perfect), no new context would be created. If not, new context would be created by the agent or taught by the teacher on the fly to make up the
lacking of the context. 

In a navigation setting, we have the $X$ inputs as the sensor inputs (e.g. GPS input, vision input from cameras, or LIDAR input from the laser sensors). 

The low level $Z$ skills correspond to recognition results (recognized type information and location information) at a given location. High-level $Z$ concepts correspond to the actions taken in different driving settings (e.g., turn slightly right when an obstacle is detected on the left-hand side). 

Using this framework, DN-2 learns multiple $Z$ concepts using different levels of reasoning: ``what'' action to take at a specific location (``where'') and ``which'' recognized object is most relevant to the current situation.

A DN-2 under such framework is similar to a UTM that simulates the navigation TMs in several ways:
\begin{enumerate}
	\item DN-2 reads natural input from the environment using its sensors from different modalities. This is equivalent to a Turing Machine reading input from an input tape.
	\item DN-2 writes to the environment by its effectors (action motor). An explicit way of writing is to put down markers of distance during navigation thus changing the input tape explicitly (as is implemented in simulation). A more subtle way is to associate navigation landmarks with recognition results in DN-2, thus changing the internal computation of the input using the hierarchy of concepts (as is implemented in the real-world environment). This is equivalent to a Turing machine writing symbols onto the input tape. 
	\item DN-2 is controlled by the emergent Finite Automaton learned through its sensorimotor experience. As we argued in previous sections, this emergent FA is computationally equivalent to the FA hand-crafted to control the transitions inside a TM. 
\end{enumerate}

Unlike a TM where the internal states must be human-defined and their transition rules must be hand-crafted, DN-2 uses emergent representation internally to learn clear logic with optimality. 

\subsection{Why planning?}
\label{sec: planning module vs. concept}
Planning is extremely important in the context of vision-guided navigation. Planning helps the agent to look ahead and avoid sudden turns and stops. It usually requires the agent to deduct what's going to happen in the current navigation context and make corresponding changes to the current action. Single image classification cannot replace planning as a single frame lacks context. For example, a bicycle moving toward the agent should be treated differently compared to a bicycle standing still, even though these two bicycles may have the same appearance in the current frame.

Planning in traditional reinforcement learning is usually treated as a separate module apart from the action execution module of the agent.  The transition from planning mode to execution mode is handcrafted and hard-coded. The execution module often uses a different form of representation than the planning module of the agent \cite{jimenez2012review}.

In the context of DN-2, planning and acting are emergent. The behavior `think' (planning), `speak' (finish planning and speak out the planned route), and `none' (overt action execution) is treated as a set of high-level concepts taught by the teacher. The behavioral definition of these states is defined by the programmer (equivalent to being defined by the agent's DNA), but the transition among those states are learned through the agent's interaction with the environment. There is no separate planning module or planning network. All the concepts use the same neuron update and firing mechanism. In our experiment, we teach the agent to compare costs, speak out its planned result, and then execute its plan. Fig.\ref{fig:planningcomparison}A and Fig.\ref{fig:planningcomparison}B illustrate this major difference. 

Fig.~\ref{fig:planningcomparison}A: In traditional automatic planning agent, planning and execution modules are separate and often use different representations. Figure from \cite{jimenez2012review}.  Fig.~\ref{fig:planningcomparison}B: In DN enabled agent, action execution and planning are treated as different concepts zones but using similar firing and learning mechanisms. Transition from planning to execution is taught by the teacher instead of hard-coded rules. 

\begin{figure}
	\centering
	\includegraphics[width=1.0\linewidth]{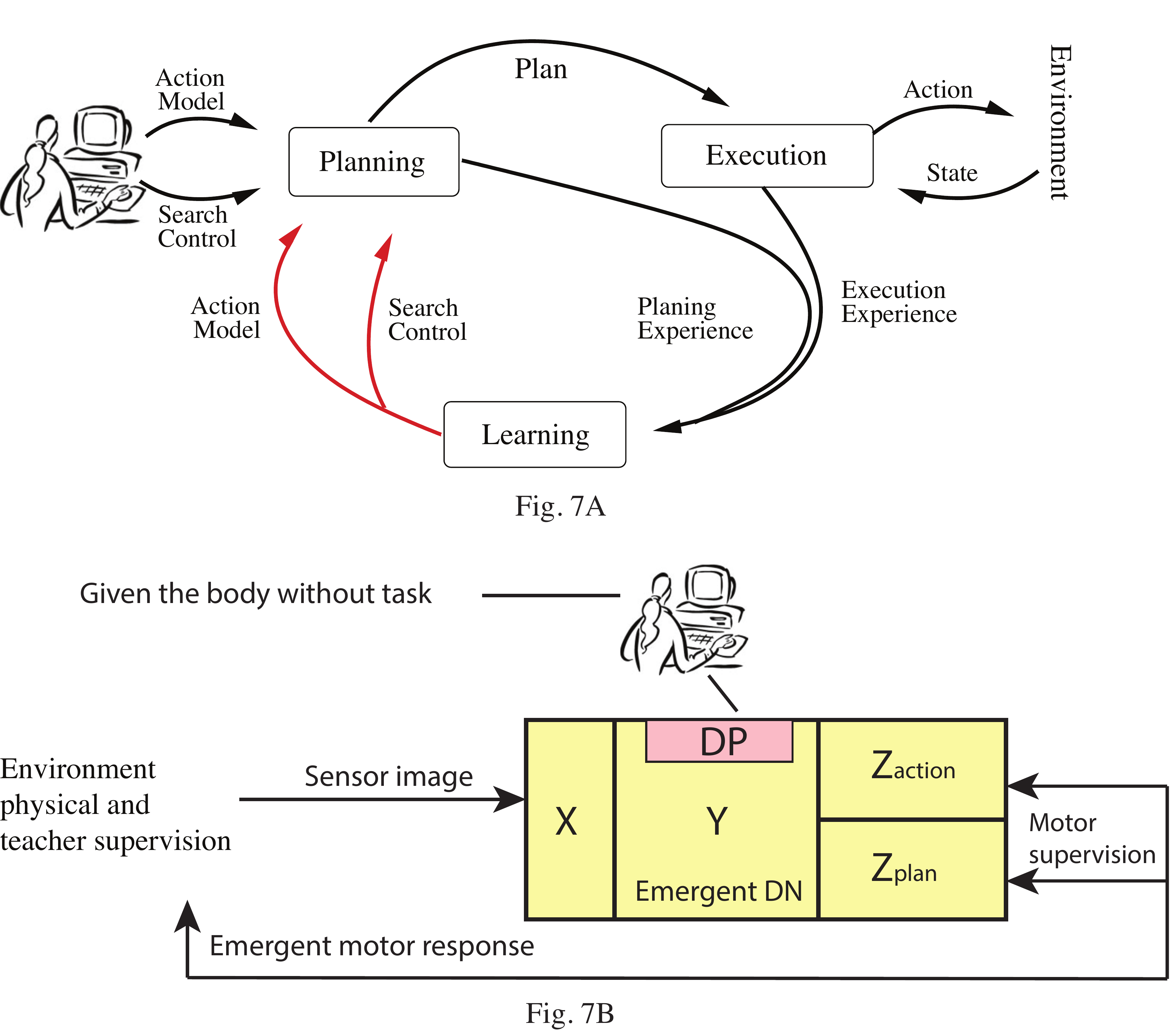}
	\caption{}
	\label{fig:planningcomparison}
\end{figure}

\subsection{Definitions}
In DN-2, planning is based on the current context, defined as $C(t) = \{\textbf{x}(t), \textbf{y}(t), \textbf{z}(t)\}$, which denotes the response vector in the $X$, $Y$ and $Z$ zone at time t. Then different skills are specific segments of the agent's learning experience: $C(t_{\textnormal{skill}_i}) \rightarrow C(t_{\textnormal{skill}_i} + 1) \rightarrow ... \rightarrow C(t_{\textnormal{skill}_i} + |\textnormal{skill}_i|)$, where $t_{\textnormal{skill}_i}$ means the starting time of the i th skill, and $|\textnormal{skill}_i|$ is the length of the i th skill.

\subsection{Hierarchical concepts in DN-2 and attention to specific concept}

Planning requires the agent to pay attention to specific parts in the current context, as not all firing concepts in the current context are relevant to the navigation tasks at hand. E.g., when planning a right turn the attention should be focused on higher-level concepts like the traffic light detection, obstacle avoidance, etc. Low-level recognition like shadow edge recognition should be ignored as these low-level features are distractors.

\begin{figure}
	\centering
	\includegraphics[width=0.9\linewidth]{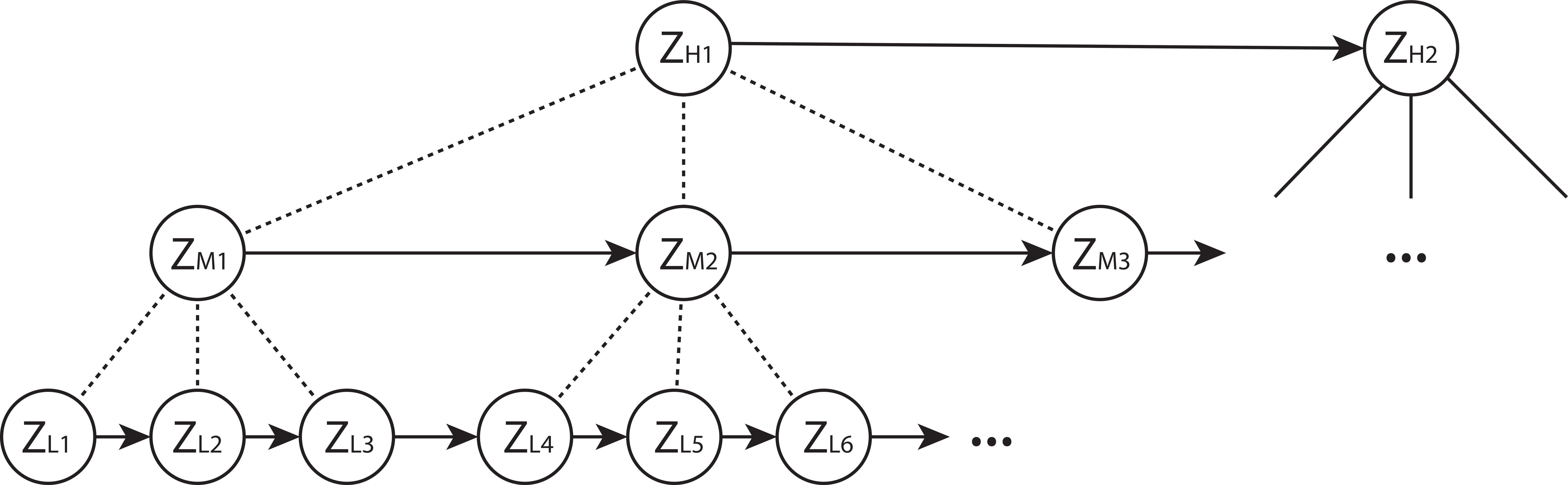}
	\caption{}
	\label{fig:sketchconcepthierarchy}
\end{figure}

In DN-2, an agent can have multiple levels of $Z$ motors, represented as different $Z$ concept zones. For example, an agent can have low-level concepts as $Z_L$, mid-level concepts as $Z_M$, and high-level concept as $Z_H$. The current context of the agent $C(t)$ is thus defined as $C(t) = \{Z_L(t), Z_M (t), Z_H(t), Y(t), X(t)\}$. A sketch of such concept hierarchy is presented in Fig.\ref{fig:sketchconcepthierarchy}. Each circle represents a high, mid, or low level concept. Arrows represent a transition among different concepts. In DN2 the arrows can be viewed as lateral connections between $Z$ neurons.

Under the DN-2 framework, each $Y$ neuron learns the following transition: $C(t) \xrightarrow[]{} C(t+1)$. However, if a $Y$ neuron has global connections to all the components in the context the number of neurons required to learn a complex task would explode. Thus attention mechanism is needed. The attended context is denoted as $C(t)^* \subseteq C(t)$. For example, a $Y$ neuron $y_i$ can pay attention to only mid-level concept transitions, which means that its attended context $C_{y_i}^* = \{Z_M, Y\}$.

Each $Z$ concept zone has a \textbf{none} neuron, which is used when training attention on specific $Z$ motors. There is also a \textbf{none} $X$ input pattern which is specifically designed when trimming attention on the $X$ zone. Attention can be trained using the following scheme:

\begin{lemma}
	To train $Y$ that learns transition on attended context: $C_{t}^* \rightarrow C_{t+1}^*$, where $C_{t}^* \subset C_{t}$, we can set the other unattended $Z$ concepts to \textbf{none} neuron firing (if $X$ is not attended, set it to \textbf{none} as well). This is under the assumption that we have enough $Y$ neuron resources.
\end{lemma}
\textbf{Proof}: If a $Y$ neuron is attending to an area, e.g. $Z_i$ where $i \in \{L, M, H\}$, outside of $C_t^*$, its connection to $Z_i$ will not be the `none' neuron. Thus its firing response will not be `almost perfect value'. If it wins the top-$k$ competition, it will not update as we will initialize a new neuron to learn the attended context perfectly due to its low response. If it does not win the top-$k$ competition, then it will also not update according to our learning rules. In any case, there will be a $Y$ neuron learning the attended concept perfectly.

If there are not sufficient number of neurons available during the above discussion, then the $Y$ neurons with the correct attention region would have a leverage over the other $Y$ neurons as their response from the unattended regions would be perfect (as they are connected to 'none' neurons) and thus more likely to win during top-$K$ competition.

\subsection{Context prediction and skill chaining }
\subsubsection{Context prediction}
When supervision for certain parts of the context (in $X$ or $Z$ zone) is not available (e.g. GPS failing in autonomous navigation) in a certain time frame, the network can generate prediction of that specific part of the context with the available parts of the concept. The problem can be formulated as: given $C(t)' = C(t) - C(t)^-$, where $C(t)^-$ is the current missing context, the network needs to generate $C(t)' \rightarrow C(t+1)$ and fill in the missing components. During network updates, the input vector from the missing regions is set to zero vectors.

Admittedly if crucial information is missing from the current context, e.g. if GPS fails at a crossroad and we do not know when to turn, the prediction would fail. But if the available context has enough information for us to make confident statistical inference, e.g., we know we are heading back home ( the high-level concept is available) and we have driven that route enough times, turning direction can be recalled.
\begin{lemma}
	\label{lemma: context prediction}
	For a context transition $C(t) \rightarrow C(t+1)$, remembered by neuron $y_i$. If $P(C(t) | C(t)') = 1$, then DN will correctly predict $C(t+1)$ given the incomplete current context $C(t)'$.
\end{lemma}
\textbf{Proof} We show that the firing neuron in $Y$ zone would still be neuron $y_i$ by contradiction. If there is another $y_j$ firing, with its remembered transition to be $C(t_j) \rightarrow C(t_j + 1)$, it would mean that $C(t_j + 1) - C(t)^-  = C(t)'$, as the weights connected to $C(t)^-$ does not participate in the network update. This would mean that $C(t')$ corresponds to $C(t)$ and $C(t_j)$ as well, violating the condition $P(C(t) | C(t)') = 1$.

If the condition $P(C(t) | C(t)') = 1$ is no longer provided, meaning that there are several contexts that can correspond to an incomplete representation $C(t)'$. Then DN-2 would choose the context that it has experienced for the most times, due to the top-$k$ competition rules in the $Y$ zone.
\subsubsection{Skill chaining to learn higher level concepts}
Higher level concepts can be learned by chaining the lower level concepts together following Lemma \ref{lemma: context prediction}.

Here we expand the context $C(t) $ to  $(Z_L(t), Z_M(t), C'(t))$., and the algorithm below serves as an example to chain a sequence of low level $Z_L$ patterns together to form a $Z_M$ pattern.

Stage 1: learning low level transitions ($(Z_L(t), \textbf{0}, C'(t)) \rightarrow (Z_L(t+1), \textbf{0}, C'(t+1))$).

Stage 2: learning high-level transitions ($\textbf{0}, Z_M(t), C'(t) \rightarrow (*, Z_M(t+1), C'(t+1)) $), where $*$ means emergent and not supervised.

According to Lemma \ref{lemma: context prediction}'s assumption, as long as the $C'(t+1)$ is informative enough, correct $Z_L(t+1)$ would emerge, and the corresponding mid-level concept would be remembered during the update.

\subsection{Planning with covert and overt motor neurons}
To enable planning, we train covert and overt actions with the transitions between covert and overt motors explicitly. Thus, each motor neuron has two states: covert and overt. If the firing motor neuron is in overt state, then the agent would perform that specific action. Otherwise, the $Z$ neuron would still fire but the action is not carried out.

Now we can add a \textbf{plan} neuron into each $Z$ concept zone. This neuron is never associated with any specific context or $Y$ neurons thus would form a unique firing pattern.

Transition into and out of covert firing is taught explicitly with specific contexts: If the \textbf{plan} neuron is firing inside a specific concept zone $Z_i(t)$, with the current context $C(t)^\textnormal{plan} = \{Z_i(t)^\textnormal{plan}, C'(t)\}$, then the teacher would teach $C(t)^\textnormal{covert} \rightarrow C(t+1)^\textnormal{covert}$, with $C(t+1)^\textnormal{covert}_{p_1} = \{Z_i(t)^\textnormal{covert} = p_1, C'(t+1)\}$. Then the teacher teaches the sequence $C(t_i)^\textnormal{covert}\rightarrow C(t_{i+1})^\textnormal{covert}$, with the planned motor always firing in covert mode.

Thus, the entire planning procedure can be written in the following training process:
$C(t)^\textnormal{plan} \rightarrow C(t+1)^\textnormal{covert}_{p_1} \rightarrow ... \rightarrow C(t+|p_1|)^\textnormal{covert}_{p_1}$.

Each plan ends with a cost neuron firing. Then the network compares the cost of different plans, the result of which is represented as a specific neuron firing in the comparison motor zone.  Transition out of covert firing is also trained, using the specific pattern in the comparison motor zone, which we present in the next subsection.

\subsection{Choosing plans based on cost and comparison}
Here we discuss the cost and comparison mechanism in DN-2 in more detail and more concept zones.

To compare the cost of two different routes to the same destination, we need to train the network to learn the cost associated with the specific plan. For every possible plan, there will be a cost concept zone associated with that plan. If we design $k$ cost concept zones, then we can compare $k$ possible plans at the same time.  In our simulation, $k = 2$.

Each cost concept zone has $c + 2$ neurons representing cost 0 to cost $c$, with one specific neuron `none'.  During skill training the cost concepts are all firing with the neuron `none'. The cost is learned during the learning procedure of planning, where $\{C(t)^\textnormal{plan},  \textnormal{cost}_\textnormal{0}\} \rightarrow \{C(t+1)^\textnormal{covert}_{p_1}, \textnormal{cost}_{p_1}\} \rightarrow ... $.

To switch from plan $p_1$ to plan $p_2$, the teacher needs to teach the comparison concept between the two cost concept zones.

\subsection{Autonomous navigation: an example}

Here we apply our DN-2 theory to a real-world scenario: autonomous navigation. In a navigation setting, we have the $X$ inputs as the sensor inputs (e.g., GPS input, vision input from cameras, or LIDAR input from the laser sensors).

The low level $Z$ skills correspond to the action at a given location. Mid-level $Z$ concepts correspond to the individual skills at different driving settings (e.g., turn right when GPS indicates to turn right). high-level $Z$ concepts correspond to the destination and route taken by the agent to reach the destination. And the cost of each trial is the cost of that specific route.

Using this framework, DN-2 learns multiple $Z$ concepts using different levels of reasoning: ``what'' action to take at specific location (``where''), ``which'' driving settings is most relevant to the current situation, ``when'' to switch route and ``why'' the network favors one specific route than another. The correspondence among concepts is presented in Fig. \ref{fig:levelz}.  The Where-What Network concept architecture serves as a general framework for AI problems that concerns with multiple hierarchies of concepts. In this paper we show a solid application of such framework in the simulated maze environment.

\begin{figure}
	\centering
	\includegraphics[width=0.4\linewidth]{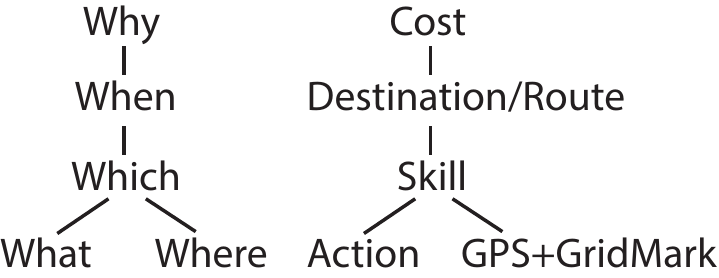}
	\caption{}
	\label{fig:levelz}
\end{figure}


In the following section, we present an implementation of DN-2 navigating in a simulated environment with the ability to plan and choose the optimal path based on its incremental learning experience.

\subsection{Maze environment}
The navigation agent and a sample of the simulated environment are presented in Fig. \ref{fig:mazegui}A,
Fig. \ref{fig:mazegui}B and Fig. \ref{fig:mazegui}C.   Fig.~\ref{fig:mazegui}A and  Fig.~\ref{fig:mazegui}B: simulated environment and corresponding GUI. The environment is 450 pixels by 450 pixels, with at most nine by nine block. The agent is 20 pixels by 20 pixels, moving continuously in the maze environment. (1) wall blocks are red blocks in the GUI. (2) obstacle blocks are blue blocks in the GUI.  (3) destination blocks are green blocks in the GUI. (4) reward/punishment blocks are yellow blocks in the GUI. (5) The agent's route is presented in the GUI. Black routes are skills already learned or going back routes. Red routes are novel skills to be learned. (6) Agent in the environment. Discussed in detail in Fig.~\ref{fig:mazegui}C.  During planning the panel displays additional information from the agent's $Z$ motors. Fig.~\ref{fig:mazegui}C: agent representation in GUI. (9) GPS signals from the environment. GPS indicates where the destination is in the simulated environment. But GPS is not aware of the obstacles in the environment thus the agent needs to learn skill ``avoid obstacle''.

\begin{figure*}
	\centering
	\includegraphics[width=0.48\linewidth]{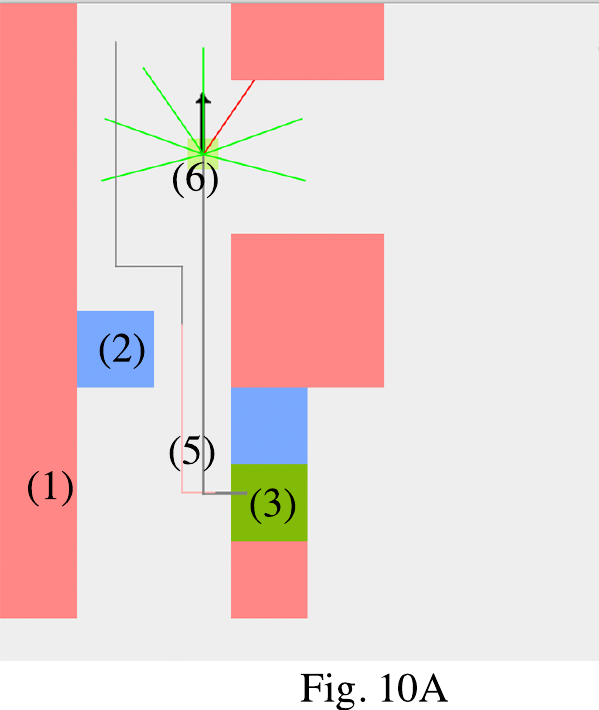}\\
	\vspace{0.3cm}
	\includegraphics[width=0.48\linewidth]{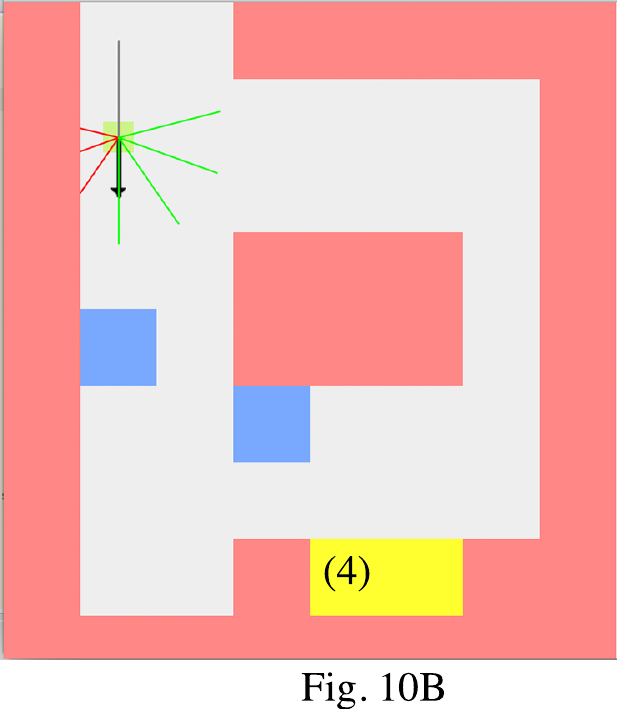}
	\includegraphics[width=0.45\linewidth]{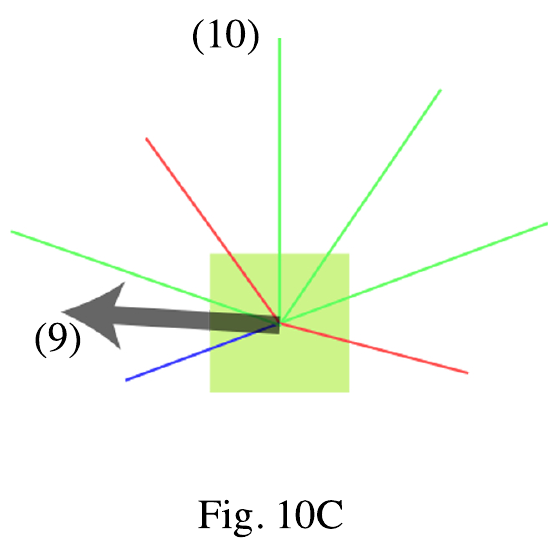}
	\caption{}
	\label{fig:mazegui}
\end{figure*}

\subsection{Environment design}
The environment is a block-based maze with different types of blocks: open, wall, obstacle, destination, and reward. Each block can be only of one type, with the size of 50 pixels in height and 50 pixels in width. Open blocks are transparent. Wall blocks are red. Obstacles are blue blocks. Destinations are green blocks and rewards are yellow blocks.

The teaching environment of the simulation was generated by the following rules: 
\begin{enumerate}
	\item Follow GPS direction when there is no obstacle or traffic light within the range of vision of the agent.
	\item If there is an obstacle seen by the agent and the obstacle is less than 20 pixels away (or, wider than 8 pixels in the vision image of the agent), the agent should turn and move toward the open block to avoid hitting the obstacle.
	\item If there is a purple traffic light seen by the agent, the agent should stop and wait.
	\item If there is a light green traffic light seen by the agent, the agent follows the other rules. 
	\item When stepping from a block to its adjacent block, the agent should put down a marker to keep track of distance.
	\item After the agent reaches the destination, it should go back to the starting point.
\end{enumerate}

\subsection{Agent design}

The agent is of size 20 by 20 pixels and moves continuously inside the simulated environment.

\begin{figure}
	\centering
	\includegraphics[width=0.8\linewidth]{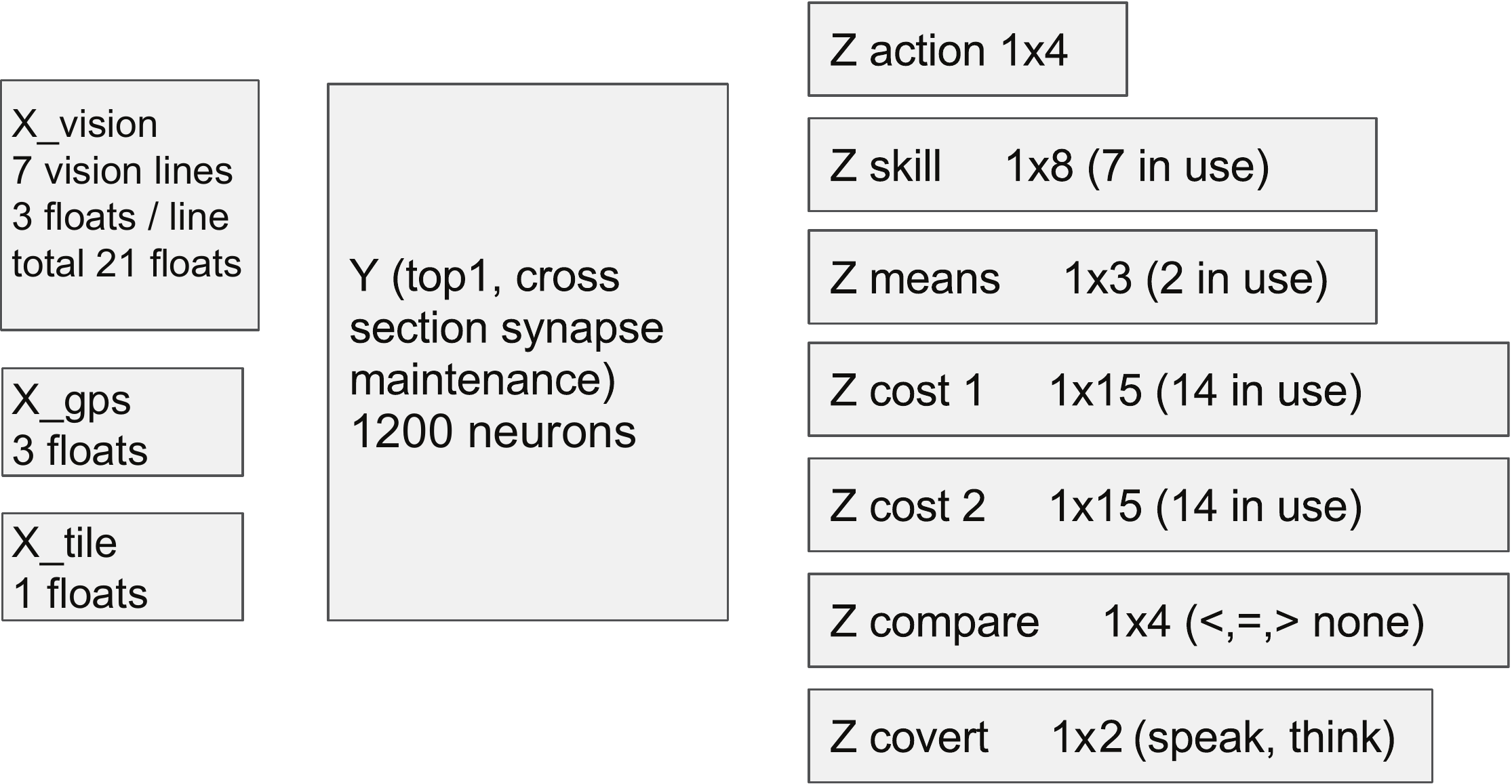}
	\caption{}
	\label{fig:DN2Network}
\end{figure}

The network inside the agent is presented in Fig. \ref{fig:DN2Network}.  All the $Y$ neurons start with global bottom-up connections from $X$, global top down connections from $Z$, and global lateral connection from $Y$. Synapse maintenance and cross-section synapse maintenance shapes the global receptive field to attend to specific areas. All $Z$ neurons have bottom-up connection from $Y$ and lateral connections from $Z$ zone.

The DN-2 nerwork has three $X$ areas:

$X_\textnormal{vision}$ : The agent is equipped with vision sensors in 7 different directions. Each vision sensor is represented as a line in the GUI. The vision sensor returns the distance to the nearest obstacle/wall and also the type of that block. The agent's vision is limited by its vision range, set to 75 pixels in our current experiment. Each vision line has three bits indicating whether the current type is open, obstacle, or wall. Thus there are 21 neurons in this area.

$X_\textnormal{gps}$: The agent can use a simulated GPS to find the right turn at each crossroad. GPS is shown as a black arrow in the GUI. GPS cannot sense the presence of obstacles thus the agent cannot follow the GPS signals blindly. When an obstacle is present the agent need to walk around the obstacle or navigation would fail. There are three neurons in this area, corresponding to left, forward, and right.

$X_\textnormal{tile}$: The agent is equipped with a tile sensor (a single neuron) which would be firing with value one if the agent is crossing the grid line.

The agent also has 7 $Z$ areas:

$Z_\textnormal{action}$: Lowest level of concept. There are four neurons in this area: forward, left, right and stop. Each forward movement advances the agent's position toward its heading position by 1 pixel. Each left/right turn increases/decreases the agent's heading by 10 degrees.

$Z_\textnormal{skill}$: Mid-level concept. There are eight neurons in this area corresponding to 8 different scenarios that are essential for navigation in the simulated environment.

$Z_\textnormal{means}$: High-level concept. There are two ways to reach the same destination in the final test. After learning each way of navigation, the agent needs to choose the one with lower cost. Thus this area has three neurons: means 1, means 2, and go back.

$Z_\textnormal{cost1}$: The first cost concept. There are 15 neurons in this area, corresponding to cost 0 to 13 and none. During teaching this concept is associated with the incrementing cost in means 1.

$Z_\textnormal{cost2}$: The second cost concept. There are 15 neurons in this area, corresponding to cost 0 to 13 and none. During teaching this concept is associated with the incrementing cost in means 2.

$Z_\textnormal{comparison}$: Comparison result concept. There are four neurons in this area: less, equal, more and none. During teaching this concept is associated with the comparison result by comparing the value in $Z_\textnormal{cost1}$ and $Z_\textnormal{cost2}$.

$Z_\textnormal{plan}$: Planning concept. There are two neurons in this area: speak and think. We discuss how thinking is taught in the next subsection.


\begin{algorithm}[h]
\caption{Teaching agent individual skills}
	\label{algo: teaching skills}
	\SetAlgoLined
	\KwResult{agent with the ability to plan and compare cost }
	initialize agent DN-2; initialize environment\;
	\For{$\textbf{z}_\textnormal{skill}$ $\gets$ 1 to skill\_num}{
		\For{$i \gets 1$ to epoch\_num}{
			Initialize random maze with crucial blocks\;
			// Agent moves to the destination block\;
			\While{Agent is not at destination}{
				Get supervised action $\textbf{z}_\textnormal{action}$ from teacher\;
				Get current $X$ input $\textbf{x}$ (for all $X$ areas)\;
				Supervise agent with $\textbf{x}$ and $\textbf{z}_\textnormal{action}$, $\textbf{z}_\textnormal{skill}$ (other concepts are zeros vectors)\;
				Update agent's DN-2\;
				agent moves according to supervision\;
			}
			// Agent goes back to initial position \;
			\While{Agent is not at initial position}{
				Get supervised action $\textbf{z}_\textnormal{action}$ from teacher\;
				Get current $X$ input $\textbf{x}$ (for all $X$ areas)\;
				$\textbf{z}_\textnormal{skill} \gets$ go back
				Supervise agent with $\textbf{x}$ and $\textbf{z}_\textnormal{action}$, $\textbf{z}_\textnormal{skill}$  (other concepts are zeros vectors)\;
				Update agent's DN-2\;
				agent moves according to supervision\;
			}
		}
	}
\end{algorithm}

\begin{algorithm}[h]
	\SetAlgoLined
	\For{$\textbf{z}_\textnormal{means}$ $\gets$ 1 to means\_num}{
		Initialize maze with specific means\;
		// Agent chains previously learned skills together\;
		// $\textbf{z}_\textnormal{cost}$ is cost1 zone if means is 1\;
		// $\textbf{z}_\textnormal{cost}$ is cost2 zone if means is 2\;
		\While{Agent is not at destination}{
			Get current $X$ input $\textbf{x}$ (for all $X$ areas)\;
			Get current cost $\textbf{z}_\textnormal{cost}$ from environment\;
			Supervise agent with $\textbf{x}$, $\textbf{z}_\textnormal{means}$ and $\textbf{z}_\textnormal{cost}$  (other concepts are emergent from previous update)\;
			Update agent's DN-2, record emergent skill and action\;
			agent moves according to emergent action\;
		}
		// Teach planning of this specific means \;
		\For{$\textbf{z}_\textnormal{skill}$ in \textnormal{skills experienced in current means}}{
			$\textbf{x} \gets X_\textnormal{background}$\;
			$\textbf{z}_\textnormal{cost} \gets$  final cost of current means\;
			$\textbf{z}_\textnormal{plan} \gets$ think\;
			$\textbf{z}_\textnormal{action} \gets$ stop\;
			Supervised agent with $\textbf{x}$, $\textbf{z}_\textnormal{action}$, $\textbf{z}_\textnormal{skill}$, $\textbf{z}_\textnormal{cost}$, $\textbf{z}_\textnormal{means}$,  $\textbf{z}_\textnormal{plan}$ \;
			Update agent's DN-2\;
		}
		// Agent goes back to initial position\;
		\While{Agent is not at initial position}{
			Get supervised action $\textbf{z}_\textnormal{action}$ from teacher\;
			Get current $X$ input $\textbf{x}$ (for all $X$ areas)\;
			$\textbf{z}_\textnormal{skill} \gets$ go back\;
			$\textbf{z}_\textnormal{means} \gets$ go back\;
			Supervise agent with $\textbf{x}$ and $\textbf{z}_\textnormal{action}$, $\textbf{z}_\textnormal{skill}$  (other concepts are zeros vectors)\;
			Update agent's DN-2\;
			agent moves according to supervision\;
		}
	}
	\caption{Teaching agent means, corresponding costs and planning}
	\label{algo: teaching means, planning and cost}
\end{algorithm}

\begin{algorithm}[h]
	\SetAlgoLined
	\For{$\textbf{z}_\textnormal{cost1} \gets$ 1 to 13}{
		\For{$\textbf{z}_\textnormal{cost2} \gets$ 1 to 13}{
			Get $\textbf{x}_\textnormal{background}$\;
			Get $\textbf{z}_\textnormal{compare}$ from teacher \;
			$\textbf{z}_\textnormal{plan} \gets$ think\;
			Supervise agent with $\textbf{x}_\textnormal{background}$, $\textbf{z}_\textnormal{cost1}$, $\textbf{z}_\textnormal{cost2}$ and $\textbf{z}_\textnormal{plan}$ (other concepts are zero vectors)\;
			Update agent's DN-2\;
			$\textbf{z}_\textnormal{plan}\gets$ speak\;
			$\textbf{z}_\textnormal{means}\gets$ (cost1 $\geq$ cost2)? means1 : means2\;
			Supervise agent with $\textbf{x}_\textnormal{background}$, $\textbf{z}_\textnormal{compare}$, $\textbf{z}_\textnormal{plan}$, $\textbf{z}_\textnormal{means}$ (other concepts are zero vectors)\;
			Update agent's DN-2\;
		}
	}
	\caption{Teaching agent compare cost and execute planned result}
	\label{algo: teaching comparison}
\end{algorithm}

\begin{algorithm}[h]
	\SetAlgoLined
	$X \gets \textbf{x}_\textnormal{background}$\;
	$\textbf{z}_\textnormal{plan} \gets$ think\;
	$\textbf{z}_\textnormal{means} \gets$ \{means1, means2\} \;
	Supervise agent with $\textbf{x}_\textnormal{background}$, $\textbf{z}_\textnormal{means}$, and $\textbf{z}_\textnormal{plan}$ (other concepts are zero vectors)\;
	Update agent's DN-2\;
	$Z \gets$ emergent response \;
	// Planning: choose between two means. \;
	\While{$Z_\textnormal{plan}$: think neuron not winning}
	{Update agent's DN-2 using emergent $Z$ \;
		$Z \gets$ emergent response \; }
	// Execute the selected means\;
	$\textbf{z}_\textnormal{means} \gets $ winning means in $Z_\textnormal{means}$\;
	\While{Agent is not at destination}{
		Get current $X$ input $\textbf{x}$ (for all $X$ areas)\;
		Supervise agent with $\textbf{x}$, and $textbf{z}_\textnormal{means}$ (all other $Z$ concepts are emergent)\;
		Update agent's DN-2, record emergent skill and action\;
		agent moves according to emergent action\;
	}
	\caption{Emergent planning behavior in DN-2}
	\label{algo: actual planning}
\end{algorithm}

\begin{table}
	\centering
	\caption{Simulation experiment result}
	\label{table: simulation result}
		\begin{tabular}{|l|l|l|l|l|}
			\hline
			Noise level      & 0\%   & 5\%   & 10\%    & 20\%    \\ \hline
			Means 1 chaining & 15/15 & 15/15 & 10/15   & 2/15    \\ \hline
			Means 1 gps  blurring   & 15/15 & 15/15 & 6/15    & 0/15    \\ \hline
			Means 2 chaining & 15/15 & 15/15 & 15/15   & 15/15   \\ \hline
			Means 2 gps  blurring    & 15/15 & 15/15 & 14/15   & 13/15   \\ \hline
			Planning         & 15/15 & 15/15 & 10/15   & 0/15    \\ \hline
			Total            & 75/75 & 75/75 & 55/75   & 32/75   \\ \hline
			Success rate     & 100\% & 100\% & 74.60\% & 41.30\% \\ \hline
		\end{tabular}
\end{table}

\begin{table}
	\centering
	\caption{Real-time training and testing detail}
	\label{table: real-time training testing detail}
		\begin{tabular}{|l|l|l|l|l|}
			\hline
			\multirow{5}{*}{Training} & Date     & Time                                                     & Detail                                                                                       & \begin{tabular}[c]{@{}l@{}}\# of \\ samples\end{tabular} \\ \cline{2-5}
			& 8/5/2017 & \begin{tabular}[c]{@{}l@{}}14:30 - \\ 15:30\end{tabular} & Training type 100                                                                            & 3051                                                     \\ \cline{2-5}
			& 8/6/2017 & \begin{tabular}[c]{@{}l@{}}14:15-\\ 15:30\end{tabular}   & \multirow{2}{*}{\begin{tabular}[c]{@{}l@{}}Training type 111\\ Neuron num: 500\end{tabular}} & 1492                                                     \\ \cline{2-3} \cline{5-5}
			& 8/7/2017 & \begin{tabular}[c]{@{}l@{}}14:00-\\ 15:45\end{tabular}   &                                                                                              & 940                                                      \\ \cline{2-5}
			& \multicolumn{4}{l|}{\begin{tabular}[c]{@{}l@{}}Right: 214, SLRight: 338, Forward: 1182, \\ SLLeft: 352, Left: 210, Stop: 136\end{tabular}}                                                                                    \\ \hline
			\multirow{3}{*}{Testing}  & \multicolumn{4}{l|}{Total Segment: 18, Total Steps: 1155}                                                                                                                                                                     \\ \cline{2-5}
			& \multicolumn{4}{l|}{Diff count: 62 (5.36\%)     Error Count: 9 (0.78\%)}                                                                                                                                                      \\ \cline{2-5}
			& \multicolumn{4}{l|}{\begin{tabular}[c]{@{}l@{}}Right: 92, SLRight: 163, SLLeft: 156, Left: 97,\\ Forward: 560, Stop: 87\end{tabular}}                                                                                         \\ \hline
		\end{tabular}
\end{table}

 \subsection{Skills, means, and costs}

\begin{figure*}
	\centering
	\includegraphics[width=1.0\linewidth]{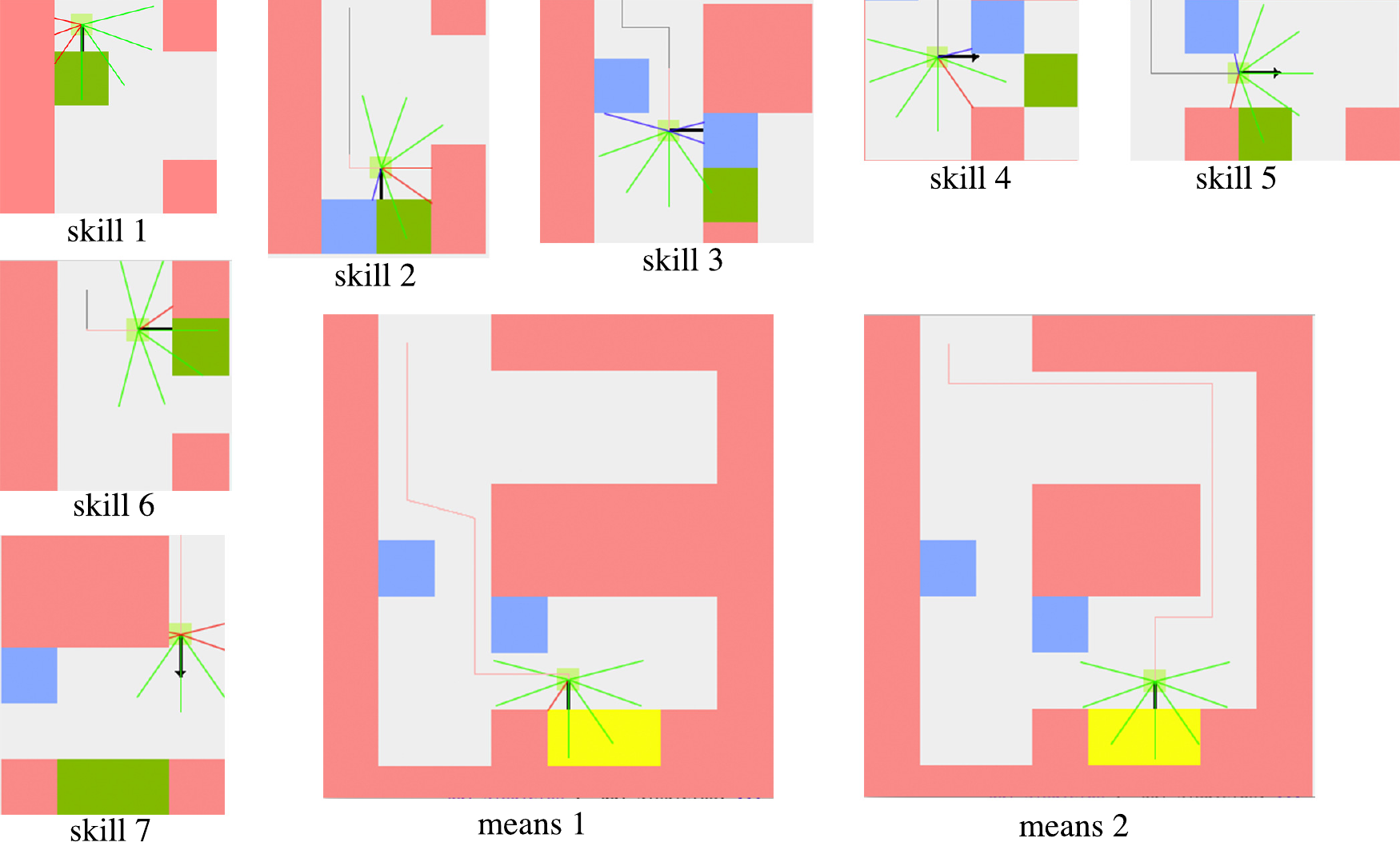}
	\caption{}
	\label{fig:skillsroutes}
\end{figure*}

\subsubsection{Teaching skills}
In this experiment, we train the agent 7 basic skills to navigate in the maze environment. These skills are presented in Fig. \ref{fig:skillsroutes}.   Skill 1: move forward when the next block is open. Skill 2: avoid obstacles and correct facing direction. Skill 3: turn left on the corner with GPS indicates left and avoid the obstacle. Skill 4: move through the narrow path. Skill 5: turn right when GPS indicates right. Skill 6: turn left when GPS indicates left. Skill 7: turn left and then turn right to reach the destination.

Note that the presented mazes are only one instance among numerous randomly generated maze environment with key blocks unchanged. Thus the learned skills are environment invariant when learning takes enough iterations.

When teaching skills, we are supervising the $Z_\textnormal{action}$ motor and the $Z_\textnormal{skill}$ motor in each network update. The other concepts are 0 vectors, and no connection is learned in the irrelevant concept zones. Teacher's behavior follows Algorithm \ref{algo: teaching skills}.

After the agent reaches the destination, the teacher leads the agent to move back to the starting position. As the agent always updates and lives continuously (no jumping back to starting position), the teacher supervises its $Z_\textnormal{skill}$ to skill\_back and the $Z_\textnormal{action}$ to the going back actions.

\subsubsection{Teaching means and costs}

After each skill is learned for enough iterations, we teach the agent two ways to navigate to the entrance at the lower right corner of a specific environment. The teacher's behavior is presented in detail in Algorithm \ref{algo: teaching means, planning and cost}.

At this stage the $Z_\textnormal{action}$ and $Z_\textnormal{skill}$ are emergent (provided by the computation of the agent). The teacher only supervises $Z_\textnormal{means}$ and the corresponding cost area ($Z_\textnormal{cost1}$ if current means is means 1; $Z_\textnormal{cost2}$ if current means is means 2). The cost starts with cost 0 and increments by one every time the agent steps into an adjacent open block, at which time the $X_\textnormal{tile}$ neuron would fire with response 1.

After the agent reaches destination, the teacher would teach the agent to plan with the experienced skills and the final cost by supervising $Z_\textnormal{action} = \textnormal{stop}$, $Z_\textnormal{skill} = \textnormal{emerged skills}$, $Z_\textnormal{means} = \textnormal{current means}$, and $Z_\textnormal{cost} = \textnormal{final cost}$. The teacher also supervises the agent with $Z_\textnormal{plan} = \textnormal{think}$. This trains the agent to recall experienced skills and the cost of the specific means. This also trans the agent to continue thinking during planning. Transition out of planning is taught in the next subsection.

After planning is taught, the teacher leads the agent back to starting position as in the previous subsection.

\subsubsection{Teaching cost comparison}
Comparison behavior is taught explicitly by the teacher. Teacher's behavior is presented in detail in Algorithm \ref{algo: teaching comparison}.

At this stage, teacher supervises the two cost zones to have two cost values. Then the teacher teaches the transition (\textbf{0}, \textnormal{cost1}, \textnormal{cost2}) $\rightarrow$ (\textbf{0}, \textnormal{compare\_result}, \textnormal{corresponding\_means}, \textnormal{speak}). Where $\textbf{0}$ means that irrelevant motors are not supervised and thus no neuron is firing.

This allows the agent to speak out his planned means based on comparing the cost values in the two different areas, and also the formed lateral connection allows the agent to switch out of thinking mode and execute the chosen means.

\subsubsection{Putting things together}
At the final stage, the teacher tests the agent's learning result by leading the agent back to the starting position and leave all motors unsupervised except for two motors: $Z_\textnormal{means}$ and $Z_\textnormal{plan}$. The teacher tells the agent to choose between the two learned means to reach the next section, thus the two neurons in $Z_\textnormal{means}$ are all firing, and in the $Z_\textnormal{plan}$ motor the teacher supervises neuron `think' to fire.

The agent continues to update until the `speak' neuron fires with a firing rate greater than the `think' neuron.

Once the agent `speaks' with the winning means, the agent continues then executes the spoken means with its eyes open, meaning that it is no longer using $X_\textnormal{background}$ but the actual input from the environment instead. Then agent then navigates according to the planned means without any supervision from the teacher.  This process is shown in Algorithm \ref{algo: actual planning}.

\subsection{Simulation experiment results}

\subsubsection{Skilling learning and chaining}
We initialized 15 DN-2 agents in parallel to learn the lower-level skills first according to Algorithm \ref{algo: teaching skills}. Then after these basic skills are mastered by the agent, the teacher leaves the lower-level skills emergent while supervised the higher level concept zones according to the first part in Algorithm \ref{algo: teaching means, planning and cost}. At this stage, the agent needs to chain different lower-level skills together, with the navigation context different from the context during the previous stage.   As shown in Table \ref{table: simulation result}, all of the 15 DN-2 agents successfully chained these lower level skills together when training higher-level skills.

\subsubsection{GPS blurring and noisy inputs}
The next experiment we did is to blur the GPS input and the sensor inputs at the same time by different degrees, shown in Table \ref{table: simulation result}. As learning means-1 (5 subtasks) is a more difficult task compared to learning means-2 (3 subtasks), more agents failed at learning means-1 when the GPS is blurred. Nevertheless, all agents succeeded in learning both means when the noise level is relatively low (less than five percent).

\subsubsection{Planning and cost comparison}
The agents that successfully learned the two means went into the next stage of tests: planning and cost comparison, as described in Algorithm \ref{algo: teaching comparison}. Again as expected, agents always succeed with moderate levels of noises in the GPS input and the vision sensor inputs, shown in Table \ref{table: simulation result}. 

\subsection{Discussion}
The video recording the entire training and testing scenario can be found at \url{https://youtu.be/CbhS1qvWZn0}. The network successfully learned the navigation rules listed in this simulated maze environment with 100\% accuracy when noise in input or GPS signals is small.  The result verifies our claims: 1) with enough resources, DN-2 learns the FA error-free using emergent representation, and 2) multiple types of neurons helped the agent to develop hierarchical representation and associate higher-level concept with lower-level concepts.

	\begin{table*}[tb]
	\scriptsize
		\centering
		\caption{Real-time testing results}
		\label{table: testing details}
		\begin{tabular}{|l|l|l|l|l|l|l|l|l|l|l|l|}
			\hline
			Section & Total& Diff & Error& L  & SL  & F   & SR  & R  & Stop & Descriptions                                                                                                                                                                                                                                                                                   & Detail                                                                                                                                      \\ \hline
			1       & 50          & 4          & 0           & 7  & 6   & 27  & 10  &    &      & \begin{tabular}[c]{@{}l@{}}F-\textgreater SL-\textgreater F-\textgreater  \\ L -\textgreater  SR-\textgreater L-\textgreater F\end{tabular}                                                                                                                                                          & \begin{tabular}[c]{@{}l@{}}rain stain, rocks, \\ facing direction correction\end{tabular}                                                   \\ \hline
			2       & 50          & 3          & 0           &    & 8   & 21  & 5   & 11 & 5    & \begin{tabular}[c]{@{}l@{}}F-\textgreater R-\textgreater SL-\textgreater R\\ -\textgreater F-\textgreater SR-\textgreater F-\textgreater Stop\end{tabular}                                                                                                                                            & \begin{tabular}[c]{@{}l@{}}rain stain, overturn, \\ facing direction correction\end{tabular}                                                \\ \hline
			3       & 59          & 1          & 0           &    & 8   & 41  & 3   & 7  &      & \begin{tabular}[c]{@{}l@{}}F-\textgreater SR-\textgreater F-\textgreater R\\ -\textgreater SL-\textgreater R-\textgreater F\end{tabular}                                                                                                                                                             & \begin{tabular}[c]{@{}l@{}}rocks, overturn, \\ facing direction correction\end{tabular}                                                     \\ \hline
			4       & 60          & 3          & 0           & 10 & 11  & 18  & 5   & 8  & 8    & \begin{tabular}[c]{@{}l@{}}F-\textgreater L-\textgreater SR-\textgreater L\\ -\textgreater F-\textgreater R-\textgreater SL\\ -\textgreater R-\textgreater F-\textgreater Stop\end{tabular}                                                                                                             & \begin{tabular}[c]{@{}l@{}}rain stain, overturn, \\ facing direction correction\end{tabular}                                                \\ \hline
			5       & 107         & 4          & 0           &    & 17  & 62  & 13  & 10 & 5    & \begin{tabular}[c]{@{}l@{}}F-\textgreater SR-\textgreater SL-\textgreater SR\\ -\textgreater R-\textgreater SL-\textgreater R-\textgreater F\\ -\textgreater Stop\end{tabular}                                                                                                                         & \begin{tabular}[c]{@{}l@{}}shadows, dirt road, protruding trees, \\ overturn, facing direction correction\end{tabular}                      \\ \hline
			6       & 104         & 6          & 3           & 27 & 13  & 35  & 22  &    & 7    & \begin{tabular}[c]{@{}l@{}}F-\textgreater SL-\textgreater SR-\textgreater SL\\ -\textgreater L-\textgreater SR-\textgreater L-\textgreater SL\\ -\textgreater SR-\textgreater F-\textgreater Stop\end{tabular}                                                                                           & \begin{tabular}[c]{@{}l@{}}shadows, protruding trees, overturn, \\ facing direction correction, error at overturn\end{tabular}              \\ \hline
			7       & 73          & 5          & 2           &    & 7   & 40  & 9   & 13 & 4    & \begin{tabular}[c]{@{}l@{}}F-\textgreater R-\textgreater SR-\textgreater F-\textgreater \\ SL(Error)-\textgreater F-\textgreater Stop\end{tabular}                                                                                                                                                   & \begin{tabular}[c]{@{}l@{}}novel obstacle, shadows, bushes, \\ overturn, error at untrained obstacle\end{tabular}                           \\ \hline
			8       & 75          & 4          & 2           & 12 & 14  & 29  & 8   & 7  & 5    & \begin{tabular}[c]{@{}l@{}}F-\textgreater SL(Error)-\textgreater F-\textgreater \\ L-\textgreater F-\textgreater SL-\textgreater R-\textgreater F-\textgreater Stop\end{tabular}                                                                                                                       & \begin{tabular}[c]{@{}l@{}}shadows, facing direction correction, \\ overturn\end{tabular}                                                   \\ \hline
			9       & 56          & 3          & 2           & 11 &     & 26  & 7   & 6  & 6    & F-\textgreater SR-\textgreater L-\textgreater F-\textgreater Stop                                                                                                                                                                                                                                  & facing direction correction, shadows                                                                                                        \\ \hline
			10      & 55          & 2          & 0           &    & 7   & 33  & 12  &    & 3    & \begin{tabular}[c]{@{}l@{}}F-\textgreater SR-\textgreater SL-\textgreater F-\textgreater \\ SL-\textgreater Stop\end{tabular}                                                                                                                                                                       & untrained obstacle, shadows, bushes                                                                                                         \\ \hline
			11      & 55          & 0          & 0           &    & 9   & 27  &     & 13 & 6    & \begin{tabular}[c]{@{}l@{}}F-\textgreater SL-\textgreater F-\textgreater SL-\textgreater \\ F-\textgreater R-\textgreater F-\textgreater Stop\end{tabular}                                                                                                                                            & \begin{tabular}[c]{@{}l@{}}facing direction correction, untrained \\ obstacle\end{tabular}                                                  \\ \hline
			12      & 66          & 7          & 0           & 13 & 7   & 25  & 13  &    & 8    & \begin{tabular}[c]{@{}l@{}}F-\textgreater SR-\textgreater L-\textgreater F-\textgreater \\ SL-\textgreater F-\textgreater Stop\end{tabular}                                                                                                                                                          & facing direction correction, bushes, shadows                                                                                                \\ \hline
			13      & 71          & 5          & 0           & 11 & 12  & 23  & 8   & 10 & 7    & \begin{tabular}[c]{@{}l@{}}F-\textgreater SR-\textgreater F-\textgreater SL-\textgreater \\ L-\textgreater SR-\textgreater L-\textgreater F-\textgreater \\ R-\textgreater SL-\textgreater R-\textgreater F-\textgreater Stop\end{tabular}                                                                 & \begin{tabular}[c]{@{}l@{}}facing direction correction, overturn, \\ rocks on the side of rode, untrained obstacles, \\ bushes\end{tabular} \\ \hline
			14      & 53          & 2          & 0           &    & 10  & 26  & 14  &    & 3    & \begin{tabular}[c]{@{}l@{}}F-\textgreater SR-\textgreater F-\textgreater SL-\textgreater F\\ -\textgreater SR-\textgreater F-\textgreater SL-\textgreater SR\\ -\textgreater F-\textgreater SL-\textgreater F-\textgreater SR-\textgreater \\ SL-\textgreater SR-\textgreater F-\textgreater Stop\end{tabular} & winding road, constant facing direction correction                                                                                          \\ \hline
			15      & 41          & 3          & 0           & 6  & 5   & 20  & 6   &    & 4    & \begin{tabular}[c]{@{}l@{}}F-\textgreater SR-\textgreater SL-\textgreater F-\textgreater SR\\ -\textgreater L-\textgreater F-\textgreater Stop\end{tabular}                                                                                                                                           & \begin{tabular}[c]{@{}l@{}}winding road, bushes, \\ facing direction correction, overturn\end{tabular}                                      \\ \hline
			16      & 59          & 8          & 0           &    & 7   & 34  & 11  &    & 7    & \begin{tabular}[c]{@{}l@{}}F-\textgreater SR-\textgreater F-\textgreater SL-\textgreater SR\\ -\textgreater R-\textgreater SL-\textgreater F-\textgreater Stop\end{tabular}                                                                                                                            & \begin{tabular}[c]{@{}l@{}}shadows, uphill, facing direction correction, \\ overturn\end{tabular}                                           \\ \hline
			17      & 70          & 0          & 0           &    & 11  & 41  & 6   & 7  & 5    & \begin{tabular}[c]{@{}l@{}}F-\textgreater SR-\textgreater F-\textgreater SL-\textgreater F\\ -\textgreater SR-\textgreater F-\textgreater R-\textgreater F-\textgreater Stop\end{tabular}                                                                                                               & untrained obstacles, bushes, winding road                                                                                                   \\ \hline
			18      & 51          & 2          & 0           &    & 4   & 32  & 11  &    & 4    & \begin{tabular}[c]{@{}l@{}}F-\textgreater SR-\textgreater F-\textgreater SR-\textgreater F\\ -\textgreater SL-\textgreater F\end{tabular}                                                                                                                                                            & untrained obstacles, bushes, winding road                                                                                                   \\ \hline
			Total   & 1155        & 62         & 9           & 97 & 156 & 560 & 163 & 92 & 87   &                                                                                                                                                                                                                                                                                                &                                                                                                                                             \\ \hline
		\end{tabular}
	\end{table*}

	\clearpage

	\section{Audition}
\label{SE:Audition}
We also use DN-2 for phoneme recognition through time series as one example since DN-2 is modality independent. During the learning procedure, 2 different types of $Y$ neurons (type 100 and type 101) mainly grow to learn feature patterns. In early stage, random sound frames with different volume levels are fed to DN-2, type 100 $Y$ neurons primarily grow to learn the sensory inputs and extract the volume information due to the synaptic maintenance. Later, type 101 $Y$ neurons grow fast to extract feature patterns and these 2 types of neurons fire together to learn the pair of sensory and motor inputs.

\begin{figure}[htb]
\centering
\includegraphics[width=0.8\linewidth]{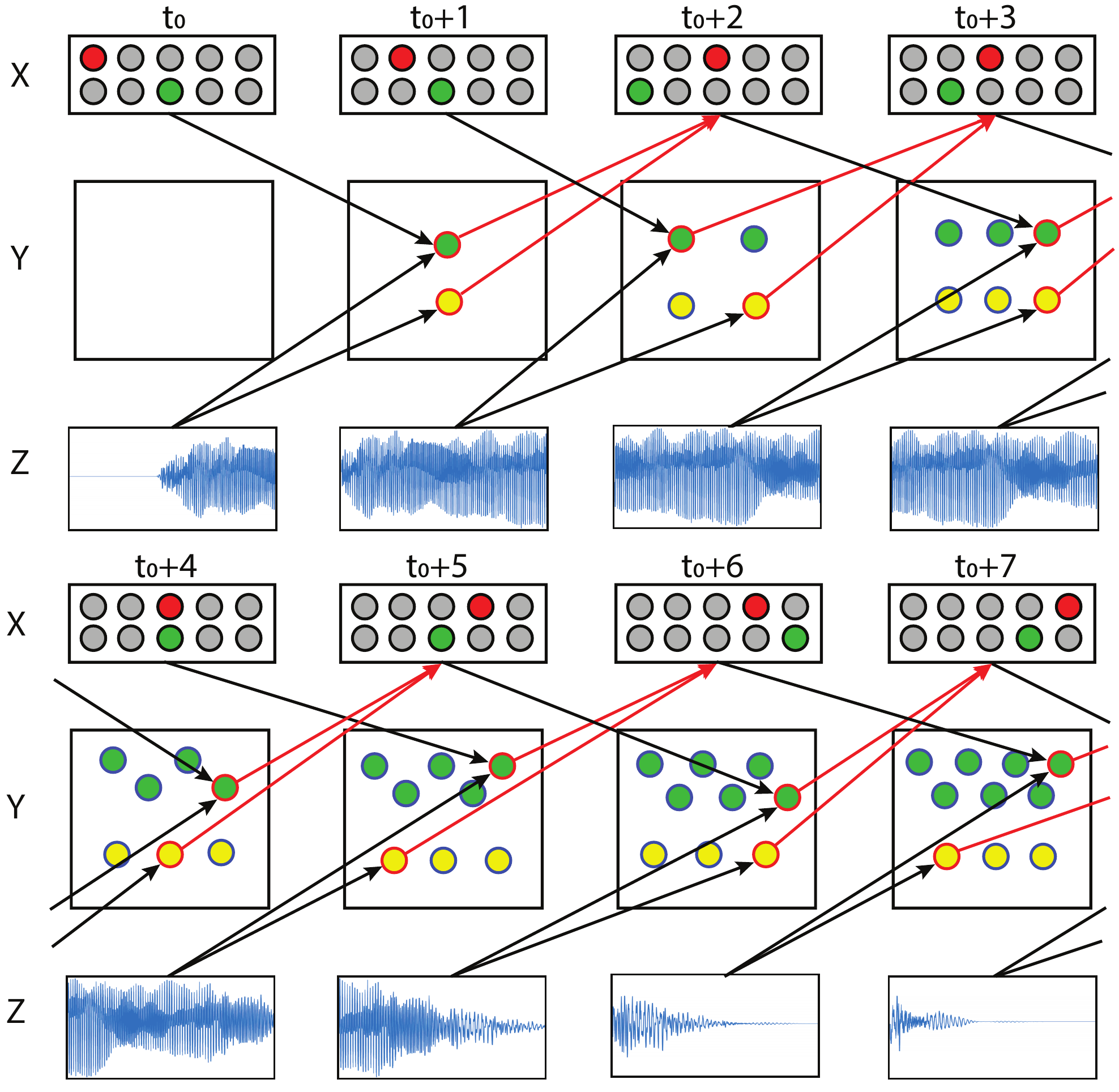}
\caption{}
\label{fig_diagram}
\end{figure}

This procedure is shown in Fig.~\ref{fig_diagram}.  Each vector $\x\in X$ is from the sensory input frame and each vector $\z\in Z$ is from the supervised motor zone. In this diagram, we only show the 2 types of $Y$ neurons (type 100 and type 101) which mainly grow for learning in the phoneme recognition experiments. The yellow nodes represent type 100 neurons and the green nodes represent type 101 neurons. The nodes with red outline mean firing neurons.

DN dynamically decides whether a new neuron is initialized by the inputs based on the degree of best matched $Y$ neuron of each type. After all neurons have been initialized, it begins to update the top-k winner neurons. DN computes in an asynchronous mode, no neurons need to wait for another neuron to complete the computation, and updates at least twice for each new pair of input pattern $\x$ and motor pattern $\z$.

DN-1 has been used for temporal processing and reached very good performance \cite{wu2017actions}. There are 2 powerful properties that help DN-1 to process sequential data: Concepts sharing and free of labeling.

\subsection{Properties}
Both DN-1 and DN-2 have three zones: a sensory zone denoted as the $X$ zone; a hidden zone as the $Y$ zone; and a motor zone as the $Z$ zone. The DN uses $\z=(\z_1, \z_2)$ as states in action zone $Z$, where $\z_1$ and $\z_2$ are the patterns.

\begin{theorem}[Concepts sharing]
Taught with $\z=(\z_1, \z_2)$, where $\z_1$ and $\z_2$ are two concepts with two different properties, respectively. Then, each of the concepts in DN can automatically utilize the other concept.
\end{theorem}
\begin{IEEEproof} Sketch only.  Shown in Fig.~\ref{fig_diagram}, the links are from $\z=(\z_1, \z_2)$ to each $Y$ neuron and then to the next $\z'=(\z'_1, \z'_2) $.  Hebbian learning determines the link values.
\end{IEEEproof}

Another property is the freedom of ``labeling''.  When dealing with some complex temporal sequences, DN will meet a large number of states/actions. It is intractable to label all such states and actions for
each task, especially the task emerges in lifelong learning.   Without knowing the current task, how can a programmer label?

\begin{theorem}[Free of labeling]

If the states/actions have emergent representations (i.e., patterns), the inner-product space
of neurons automatically sorts out the distance between patterns without a need for otherwise
intractable handcrafted discrete labeling for each pattern.
\end{theorem}
\begin{IEEEproof}  To be more intuitive, let us prove by examples.  Suppose $\v_1, \v_2, \v_3$ are three patterns (vectors) that represent three states/actions.   If they were labeled, let them have the labels $l_1, l_2, l_3$, respectively, where
some of these labels can be equal.  For example, $l_1 = l_2$, but $l_1 \ne l_3$.  Hand-labeling of $\v_1$ and $\v_2$ by the same label $l_1 = l_2$ is static and impractical in task-nonspecific settings,
because such labels may change according to task.  For example, in speaker identification, male and female voices must be distinguished, but in speech recognition, male and female voices should not be
distinguished as such.   In Hebbian learning, a normalized version of the inner production $\v\cdot \w$ was used as the primary value for pre-response, where $\v$ is the state/action, and $\w$ is the weight vector of the neuron that
takes $\v$ as input.  If $\v_1$ is similar to $\v_2$, but $\v_1$ is very different from $\v_3$, such representation differences in $\v$ vectors is naturally reflected in the value of inner product although
the normalization will disregard the length of $\v$ and $\w$.  This corresponds to a kind of ``soft'' labeling using vector representations without requiring the programmer or the teacher to discretely link $\v_1$ and
$\v_2$ to be the same but $\v_1$ and $\v3$ to be different.
\end{IEEEproof}

Of course, discrete labeling is different from such soft labeling, but the former is static and intractable for complex tasks in real time and the latter is dynamic and automatic.

In our experiments, we did not label every symbol for the stage concept zone. We designed a large space with enough $Z$ neurons for the stage concept zone and let $Z$ neurons fire automatically to emerge a huge number of different patterns during learning.

Besides these properties inherited from DN-1, DN-2 has additional mechanisms which can help it process temporal sequences. We'd like mention one important mechanism: multiple types of neurons, which we believe mainly helps DN-2 outperform DN-1 in the phoneme recognition experiments.

\subsection{Multiple types of neurons}
According to  \cite{chugani1998critical}, most regions (e.g., sensorimotor cortex, thalamus, and brain stem) have been formed in the newborn brain, and the early wiring between neurons has also developed. To simulate these different early connections, we define several types of early Y neurons with different basal connections, instead of using synaptic maintenance shaping wirings during learning from the same initial connections. With this condition, the distribution of neurons becomes widespread, and DN-2 will speed up the learning during early stage.

We consider all possible connection categories for the $Y$ neurons. One common type neuron is that with connections with $X$ zone and $Z$ zone. This type is modeled on the neurons in the pathway of sensorimotor cortex (e.g. on the cerebellar cortex) which transport sensory information to motor cortex \cite{glickstein2000visual}.These neurons are used in previous DN and very useful for learning specific patterns in the early stage.

We also define the $Y $neuron only having connections with $Z$ zone. This kind of $Y$ neuron simulates neurons located on the primary motor cortex and premotor cortex of human brain \cite{campbell1905histological}. They are triggered by a specific concept or action from motor zone and fire to produce impulses conducted to motor zone. Similarly, the type of  Y neuron with connection only to $X$ zone is defined. The role of these neurons is like the cone and rod cells in the retina or the hair cells in the cochlea.

The neuron with connections only to other $Y$ neurons is also included as one category. The neurons on lateral Intraparietal (LIP) cortex, which contribute to the working memory associated with visual attention \cite{bisley2003neuronal}, are almost identical to this category. We also define the category of neuron having connections with $Y$ and $Z$ zones. For DN-2, this kind neuron mainly supports the thinking or imagination mode. The category of neurons having connections with $X$ and $Y$ zones and the category of neurons with connections to $X$, $Y$ and $Z$ zones are defined. We believe these neurons will focus on the later stage learning and help DN-2 to generate the hierarchical structure.

When different types of $Y$ neurons fire simultaneously, the firing pattern is composed of these neurons. In this case, more situations can be represented when the total number of $Y$ neurons is certain. When there are $2m$ single type neurons in $Y$ zone, $P_{2m}^{k}$ situations can be represented ($k$ is the top-k number,  $P$ represents the permutation). If the $2m$ neurons are equally distributed in 2 types, $(P_{m}^{k})^2$ situations can be represented. For a specific neuron, the probability of firing increases from $\frac{k}{2m}$ to $\frac{k}{m}$.

Let us consider a simple case, suppose there are 10 single type of neurons in $Y$ zone (with top-1 competition), then only 10 different features or situations can be learned. If there are 2 types of neurons, and each type has 5 neurons. With the combination, there are 25 different features or situations can be learned. For more complex cases, the multiple types of neurons will have more powerful representation.

\subsection{Locations of the neurons}
Unlike many other artificial neural networks, in which the hidden neurons are location-free, the hidden neurons have locations inside DN-2's ``skull''. This new mechanism encourages the smoothness of hidden representations--nearby neurons detect similar features. It allows the recruitment of neurons during the lifelong learning implemented by Hebbian learning. This
process of recruitment gradually adapts a liquid representation to better fit the changing
distribution.

Based on their functions, the brain is separated into many regions \cite{andreasen1996automatic}. In this sense, the neurons in the brain are distributed according to their responsibilities. Early experience plays a key role in the development of the neuron distributions and connections in the brain since the brain has to adapt continuously to the outside environment. The loss of a particular sense often leads to the rewiring of the deprived cortical zone and other modality inputs during the development. We believe the neurons grow and locate unevenly in the brain based on the learning experience.

We design a mechanism in DN-2 to automatically arrange and update the location for each neuron. This mechanism mainly simulates the pulling effects between neurons instead of including all the complex biological functions. We design a fixed size skull and arrange a fixed number of glial cells evenly distributed in it. The first neuron is located in the center of the skull, and later each newborn neuron will settle in a random location closest to an existing neuron with the most similar learning experience (highest match of input patterns) until all resources are depleted. During the learning procedure, each glial cell will pull the nearest specific number of neurons towards it periodically. In an early stage, the neurons' locations are near the center of the skull. During the development, the neurons will be dispersed by glial cells throughout whole skull space. In the later stage, the neurons with similar learning experiences are located closely, the neurons with different learning experiences are located far away. The neurons' distribution is uneven in the skull. In the later stage, there are more neurons learning the common features during learning procedure which are densely located in specific regions. There are fewer neurons learning rare features which are sparsely distributed in the skull.

\subsection{Experiments}
\label{SE:exp}
We train DN-2 with the audition sequential data, and use one re-substitution test and three partial disjoint test sequences to measure DN-2's performance of phoneme recognition.  We also compared the DN-2's experimental results with DN-1's.

In the experiments, the audition data consists of recordings of 44 phonemes (according to the Received Pronunciation of Standard English) and silence. All these audio data are recorded in the same natural environment (not noise-free). Our sampling rate of the recordings is $f_s=44.1$KHz.

There are 2 steps in the training stage. We first train DN-2 with pre-training sequence. During this procedure, type 100 $Y$ neurons mainly grow and DN-2 can learn volume information. Then the training sequence is used to train DN-2 and type 101 $Y$ neurons mainly grow to learn feature patterns. During the test stage, one re-substitution test and 3 partial disjoint tests are used to test DN-2.

The pre-training sequence contains random sound frames with different volume, including silence. The training sequence is composed of 44 subsequences. Each subsequence contains one phoneme's frames. The phoneme's frames are between 500ms starting silence frames and 500ms ending silence frames.  The re-substitution test sequential data is the same as training sequential data. The partial disjoint test sequences have the same structure, but half subsequences are changed to other records recorded in the same environment.

\begin{figure}[tb]
\centering
\includegraphics[width=0.9\linewidth]{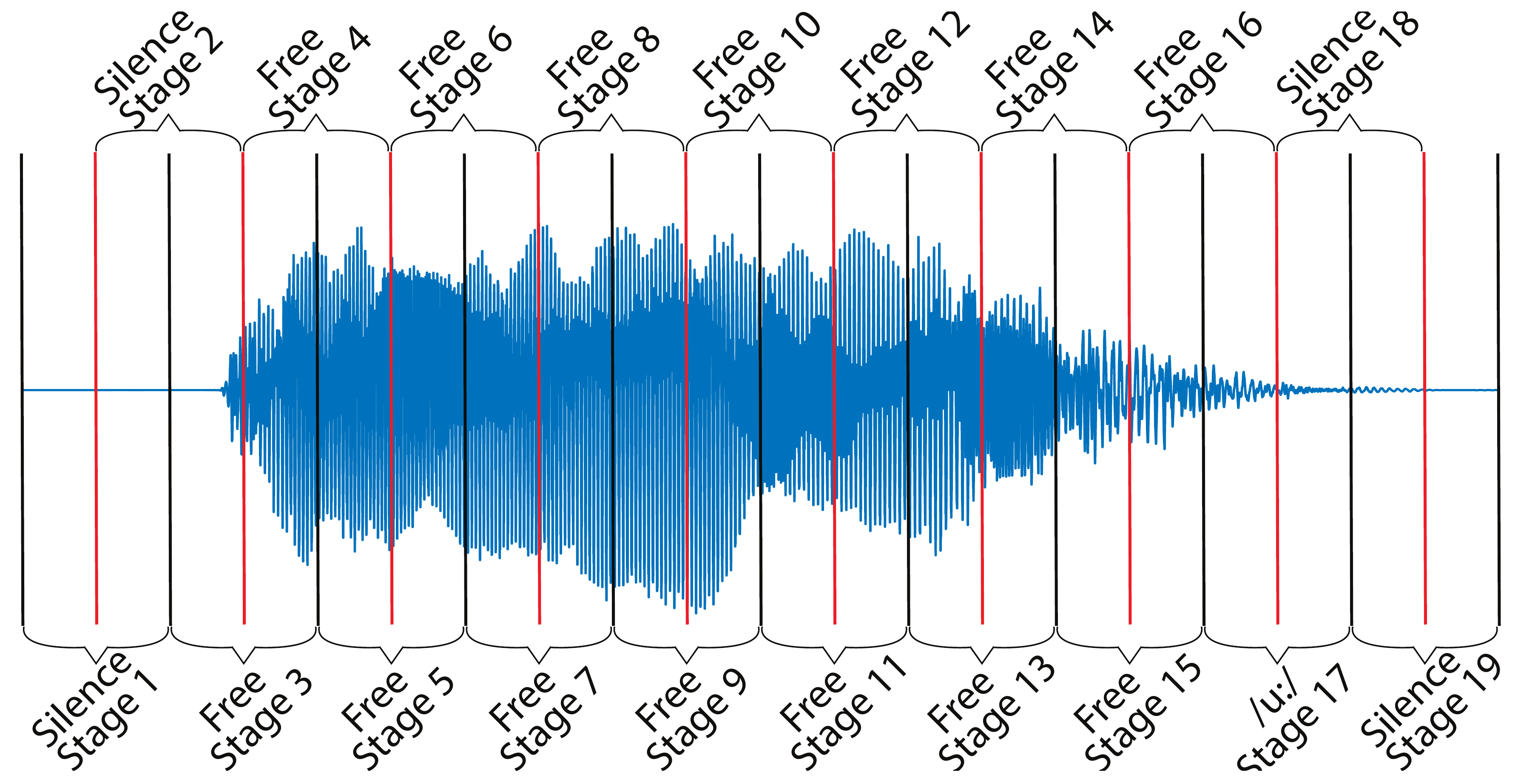}
\caption{}
\label{fig_cutting}
\end{figure}

As a tradeoff between rich of features and stationary, we cut the phoneme and silence audio data into 40 ms frames with 20ms overlap ($T=40$ms) now. Frames with 20ms to 40ms length are used widely in audition processing systems. The overlap between neighbor frames makes inputs smoother. An example of this setting is schematically illustrated in Fig.~\ref{fig_cutting}, using the case of /u:/.   Each segment is 40ms long with 20ms overlap.

Supervision is always provided during training, and is also available in the first frame during the re-substitution test and partial disjoint tests. In other words, DN-2 will use the states generated by itself as contexts during tests.

\subsubsection{Simulated cochlea for waveform pre-processing}

\begin{figure}[tb]
\centering
\includegraphics[width=0.8\linewidth]{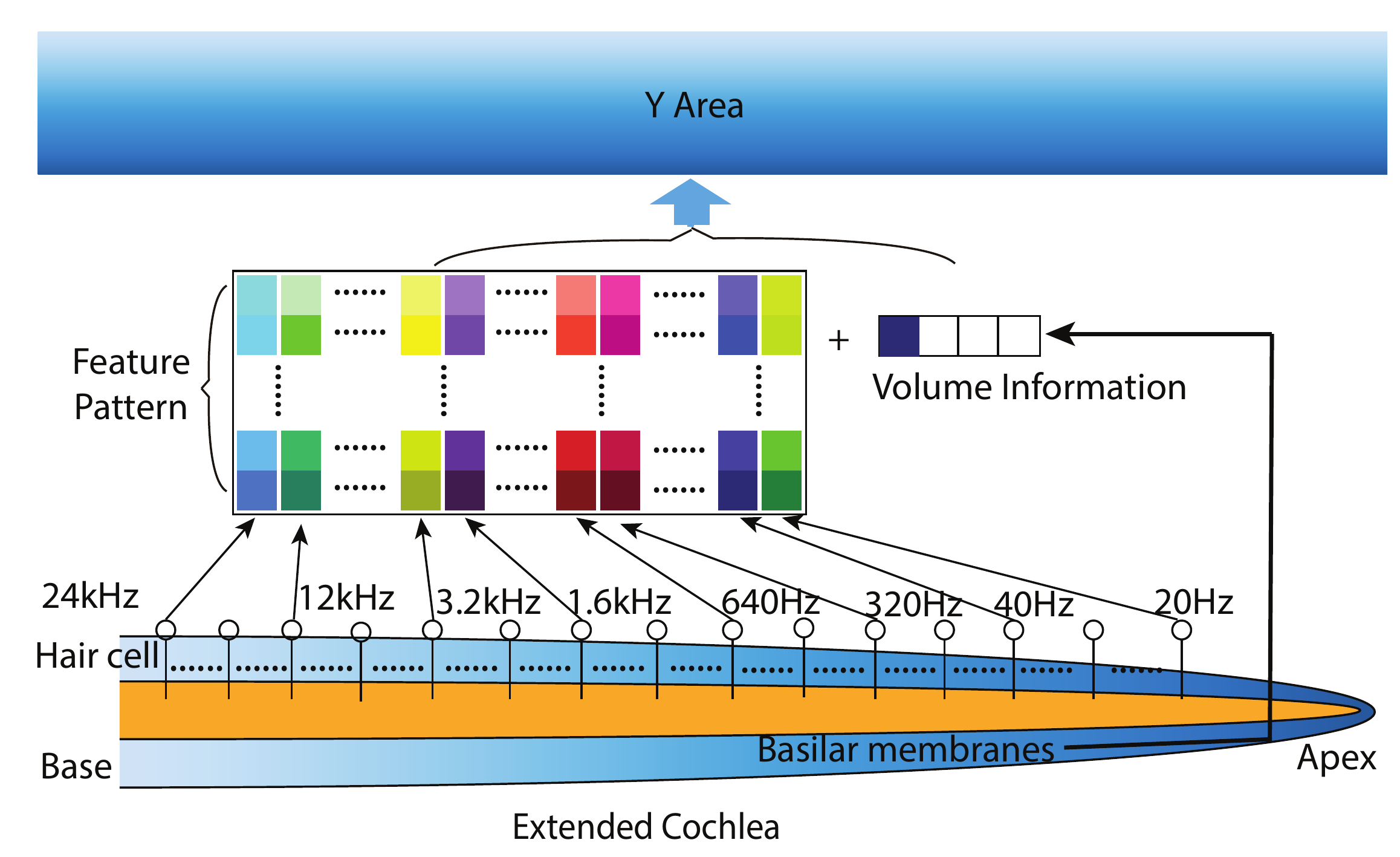}
\caption{}
\label{fig_labelcochlea}
\end{figure}

For each input data frame, we model the cochlea to generate a feature pattern from the raw waveform. The details of the simulated cochlea are shown in Fig.~\ref{fig_labelcochlea}. 
The feature pattern is extracted through a series of filters with different passband (different columns) and different initial phase (different rows) as long as the simple volume information pattern from the basilar membrane.  
With this kind of patterns as representations, we can retain many prime features of the audio data.

The idea of time shifting is included in our simulated cochlea. When acoustic signals arrive at the cochlea,  the basilar membrane and organ of Corti start to vibrate. Thousands of hair cells sense the motions and convert the vibrations into electrical signals.  These nerve signals will be transferred to our brain for further process. With different features in different positions \cite{tilney1992actin}, these hair cells do not actually detect the motions at the same time. In other words, there is a tiny time delay between the hair cells in different positions. According to \cite{liu1997effects}, there exists a feedback mechanism through the outer hair cells in the cochlea. This feedback loop also implies a time delay. We need to implement time shifting for our audio frame data to simulate time delay.

Linear Prediction Coefficients (LPCs) and Mel-frequency cepstrum coefficients (MFCCs) computation techniques \cite{deng2003speech} are common feature mapping methods used in the acoustic signals processing. By Fourier Transformation, these methods analyze signals based on power spectra in the frequency domain.  In this way, the phase information in signal has been ignored. In fact, phase information contains many useful features. As summarized in \cite{liu1997effects}, the perception of consonants depends mainly on phase information.

Our method, in a different way, analyzes acoustic signals in time domain. We utilize a series of sine functions with different frequencies and different initial phases, as filters, to multiply the raw data and obtain a matrix with components corresponding to different frequencies and initial phases. Extraction with filters of different frequencies means extracting features in different frequencies without missing phase information. The filters with low pass-band correspond to the hair cells in the apex of the cochlea, and the filters with high pass-band correspond to the hair cells in the base of the cochlea. The generation function of each component $a_{i,j}$ in each frame's feature matrix $A$ is as follows:

\begin{equation}
\label{eq_generation function}
a_{i,j} = \sum_{t=1}^T u(t)  \sin(\omega_i t + \theta_j)
\end{equation}
where $u(t)$ is the raw frame data, $\omega_i$ is proportional to the frequency of pass band $f_i$ ($\omega_i= 2 \pi f_i$). $\theta_j$ is the initial phase, different initial phase means different time shifting for the frame. $1 \leq i \leq N_f$, $1 \leq j \leq N_s$. The assignment of $\theta_j$ is as follows:

\begin{equation}
\label{eq_theta function}
\theta_j = 2 j\pi /N_s, \ 1 \leq j \leq N_s.
\end{equation}

In our experiments, we define $N_s=10$ , and $\theta_1=1/5 \pi$. The frequency of each pass band is set to increase $\alpha$ times each time.

\begin{equation}
\label{eq_frequency}
\omega_{i+1} = \alpha \omega_i, \ 1 \leq i \leq N_f-1
\end{equation}

Since the range of human hearing is between $20Hz$ to $20KHz$, we set $f_1=20$ and $f_{N_f}=20000$. With logarithm ($N_f-1=\log_\alpha f_{N_f}/f_1$),  $N_f$ can be obtained. We set $\alpha = 2$, and $N_f=11$. The frequencies can be calculated as follows:
\begin{equation}
\label{eq_frequency2}
f_i = 2^{i-1} \cdot 20, \ 1 \leq i \leq 11.
\end{equation}

After above processing, we get a feature matrix $A$ $(A \in \mathbb{R}^{11 \times 10})$ to represent each frame data. To increase the contrast of each element in the feature matrix, we use Gamma correction for each element \cite{poynton2012digital}. The formula is as follows:
\begin{equation}
\label{eq_gamma}
a_{i,j} = a_{i,j}^{\gamma}, \ \gamma =0.5.
\end{equation}

To avoid silence frame data becomes noise interference after normalization, we add volume information in our model of the cochlea. As demonstrated in \cite{fuchs2010oxford, hudspeth1989ear} and \cite{sarma2014phoneme}, The amplitude of a sound determines how many nerves associated with the appropriate location fire, and a loud sound excites nerves along with a fairly wide region of the basilar membrane, whereas a soft one excites only a few nerves at each locus. To simplify this procedure, we only use one $1 \times 4$ feature vector to represent different volume levels. The representation details are in the Table \ref{table: representation of volume}. We set three thresholds and use the $l_2$-norm of each frame to compare with the thresholds to decide the volume of the frame. In this way, the volume information is fed to DN together with the corresponding frame in real-time.

\begin{table}[tb]
	\centering
	\caption{representation of volume information}
	\label{table: representation of volume}
	\resizebox{0.5\columnwidth}{!}{%
	\begin{tabular}[c]{|c||c|c|c|c|}
		\hline
		Vector  & 1000 & 0100 & 0010 & 0001   \\ \hline
        Representation & Silence & Low &Medium &High \\ \hline
	\end{tabular}
	}
\end{table}

\subsubsection{Labeling and emergence}
We designed three motor concept zones to make DN-2 automatically generate dense states in every frame, which can provide necessary temporal contexts during the training and tests. Concept 1 consists of phoneme recognition states, concept 2 is composed of stage states. Concept 1 has silence, free, and 44 different phoneme states (totally 46), and concept 3 consists of 4 different volume levels. The silence and 44 phoneme states correspond to silence and current phoneme category. Free state means DN has not yet decided, and needs more input frames to determine. The states in concept 2 zone are designed following the FA rule. These states represent every new situation DN meets, and can also serve as contexts to help DN to make decisions. The states in  concept 3 zone represent the 4 different volume levels which mainly help DN-2 learn volume information.

For the /u:/ case in Fig.~\ref{fig_cutting} , the state of concept 1 keeps as silence when inputs are silence frames. It becomes the free state when phoneme frames are coming in, and changes to /u:/ state when the first silence frame shows up after phoneme frames. In the meanwhile, the state of concept 2 continues counting stages.

We labeled the concept 1 and concept 3 zones since they do not contain many states. With specific labeling of these concept zones' states, we can measure recognition performance accurately.

We did not label the concept 2 zone manually since there are much more stage states needed. We just design this concept zone with enough $Z$ neurons. (In our current experiment, the number of $Z$ neurons is 800.) According to the free of labeling theorem mentioned above, different $Z$ neurons fire according to $Y$ zone's firing pattern in last frame and different firing patterns will emerge in the $Z$ zone to form different stage actions.

\subsubsection{Results and analysis}
We chose performance of concept 1, which indicates phoneme recognition rates, to compare with results of DN-2 and DN-1. In these experiments, 1400 $Y$ neurons in DN-1 grow to learn while 1300 $Y$ neurons in DN-2 grow to learn. The comparison between these two experimental results is shown in Table \ref{table: comparison1}.  The DN-1's performances are listed in the first row and the DN-2's performances are listed in the second row. The second to fifth columns demonstrate performances of re-substitution and three partial disjoint tests respectively. The average performance is listed in the last column. With multiple types of $Y$ neurons firing to form a more stable firing pattern, the average phoneme recognition error rate reduces $76.9\%$.

\begin{table}[tb]
	\centering
	\caption{Comparison of error rates (\%): DN-1/DN-2.  Column 2: the error rate of the re-substitution test; column 3-5: error rates of three partial disjoint tests respectively.}
	\label{table: comparison1}
	\begin{tabular}[c]{|c||c|c|c|c||c|}
		\hline
		Methods / Tests & 1 & 2 & 3 & 4 & Average  \\ \hline
DN-1  & 1.96 & 10.62 &6.56 &4.53 &5.92\\ \hline
DN-2 & 0.0 & 1.53 &2.25 &1.68 &1.37\\ \hline
	\end{tabular}
\end{table}

\begin{figure}[tb]
\centering
\includegraphics[width=0.7\linewidth]{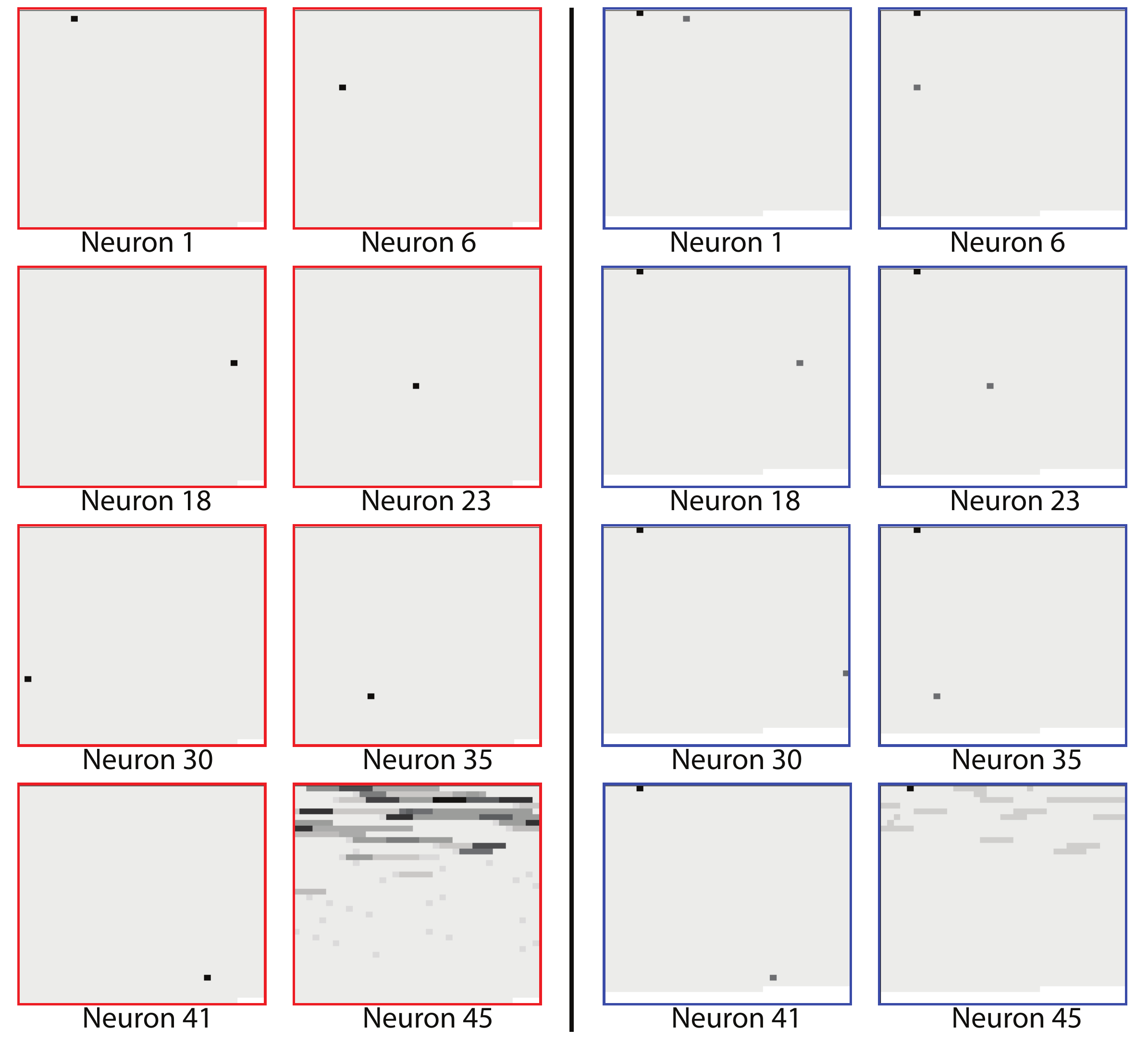}
\caption{} 
\label{fig_comparsion}
\end{figure}
 
We selected 8 neurons from motor concept zone 1 of DN-1 and DN-2 respectively, and visualized their bottom-up weights which can show the $Y$ zone firing patterns in Fig.~\ref{fig_comparsion}.  Both in DN-1 and DN-2, neuron 1, 12, 18, 23, 30, 35, 41 learned a specific phoneme class and neuron 45 learned the silence class. The darker color means higher value. The lighter color means lower value. The white region in the last row is used for completeness.

The motor neurons with same serial number in DN-1 and DN-2 learned the same content of concept. Both in DN-1 and DN-2, the first seven selected motor neurons (neuron 1, 12, 18, 23, 30, 35, 41) learned a specific phoneme class and the last selected motor neuron (neuron 45) learned the silence class. We reshaped each motor neuron's bottom-up weight vector to a $38 \times 37$ matrix, and use the white area in the final row to complete the graphs.

 From the right side of the image, we can see clearly that 2 types of $Y$ neurons (type 100 and type 101) fired together in DN-2 and these $Y$ neurons composed more static firing pattern in $Y$ zone. These combined firing patterns also reduce the number of $Y$ neurons needed to learn all pairs of sensory and motor inputs. From the bottom-up weights of the motor neuron (neuron 45) learned silence class, we can find that less type 101 $Y$ neurons are grown to learn the silence feature since the type 100 $Y$ neurons learned volume information help to distinguish silence from other phoneme frames.

\begin{figure}[tb]
\centering
\includegraphics[width=0.7\linewidth]{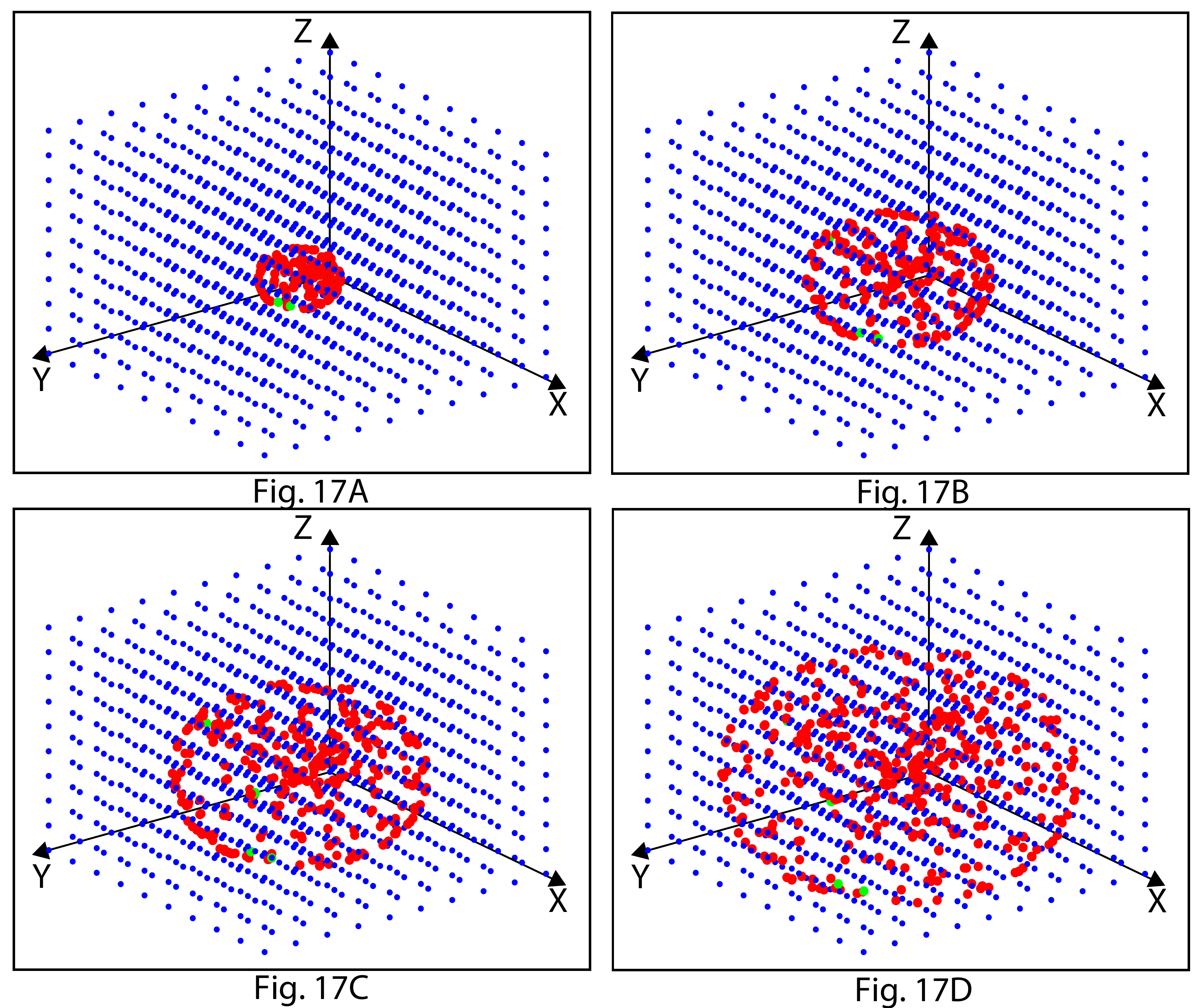}
\caption{}
\label{fig_location}
\end{figure}

The $Y$ neurons' locations during the procedure of growth are visualized in Fig.~\ref{fig_location}A to Fig.~\ref{fig_location}D.  Fig.~\ref{fig_location}A: early stage; Fig.~\ref{fig_location}B and Fig.~\ref{fig_location}C: intermediate stage's situations with Fig.~\ref{fig_location}C later than Fig.~\ref{fig_location}B; Fig.~\ref{fig_location}D: final stage. The little blue points, evenly distributed in the skull space, are the somas of glial cells, and the red points are the somas of type 101 $Y$ neurons. The green points are somas of type 100 $Y$ neurons. We can see clearly that the $Y$ neurons explore from the center to the whole space. The $Y$ neurons are distributed unevenly because each $Y$ neuron births at a different time and the number of neurons with a similar learning experience is different.

	\section{Conclusions}
	\label{SE:conclusions}
As far as we know, presented here is the first developmental method that is backed by the theory of emergent Universal Turing Machines.   Theoretically, Strong AI is based on a solid 
mathematical foundation of logic --- Universal Turing Machines --- and a solid mathematical foundation of
optimality --- maximum likelihood.   

Strong AI seems to be practical too:  Because of the high dimensionality of emergent internal features and the long lifetime, 
 the seemingly computationally intractable problem of batch optimization has been efficiently addressed through maximum likelihood estimation of the new DN-2.  The optimality is under the conditions of incremental learning, limited learning experiences, and limited resources.  There are no iterations needed.   DN-2 enables a fluid hierarchy of international representation.  This fluid hierarchy hopefully is
 useful for us to reconsider the human 
 handcrafted hierarchies of brain regions (e.g., those of Van Essen \cite{FellemanVanEssen91} and others but do not explain the cross-modal plasticity of Sur and coworkers \cite{VonMelchner00}) which are still popular in neuroscience.  The time complexity of a network like DN-2 is linear 
in time if the number of neurons is a large number but constant, as proved in Weng 2012 \cite{weng2012symbolic}.   Namely, only a constant amount of
computation is needed at each time instance --- suitable for an imbedded computer hardware running
in real time and we do not need to worry that it will slow down its per-frame response regardless of 
tasks.   The slowdown will take place when the task is complex and the skills of the learner are
still weak.   This is caused by the lack of confidence on the part of the learner.  

The experimental result seems to support that in practice this mechanism improves the generalization powers of alternative methods that use input and action patterns only.  The visualization of the internal representations supports the theory of fluid hierarchy.

\newpage

		\bibliographystyle{plain}
	\bibliography{ZhengZejiaCite,WuXiangCite}

\newcommand{\noop}[1]{}
\begin{thebibliography}{10}

\bibitem{andreasen1996automatic}
N.~C. Andreasen, R.~Rajarethinam, T.~Cizadlo, S.~Arndt, L.~A. Flashman, D.~S.
  O???leary, J.~C. Ehrhardt, and W.~T. Yuh.
\newblock Automatic atlas-based volume estimation of human brain regions from
  mr images.
\newblock {\em Journal of computer assisted tomography}, 20(1):98--106, 1996.

\bibitem{bisley2003neuronal}
J.~W. Bisley and M.~E. Goldberg.
\newblock Neuronal activity in the lateral intraparietal area and spatial
  attention.
\newblock {\em Science}, 299(5603):81--86, 2003.

\bibitem{JOURNALtutorial}
P.~Boer, D.~Kroese, S.~Mannor, and R.~Rubinstein.
\newblock A tutorial on the cross-entropy method.
\newblock {\em Annals of operations research}, 134(1):19--67, 2005.

\bibitem{campbell1905histological}
A.~W. Campbell.
\newblock {\em Histological studies on the localisation of cerebral function}.
\newblock University Press, 1905.

\bibitem{chugani1998critical}
H.~T. Chugani.
\newblock A critical period of brain development: studies of cerebral glucose
  utilization with pet.
\newblock {\em Preventive medicine}, 27(2):184--188, 1998.

\bibitem{deng2003speech}
L.~Deng and D.~O'Shaughnessy.
\newblock {\em Speech processing: a dynamic and optimization-oriented
  approach}.
\newblock CRC Press, 2003.

\bibitem{FellemanVanEssen91}
D.~J. Felleman and D.~C. {Van Essen}.
\newblock Distributed hierarchical processing in the primate cerebral cortex.
\newblock {\em Cerebral Cortex}, 1:1--47, 1991.

\bibitem{fish2013novelty}
J.~Fish, L.~Ossian, and J.~Weng.
\newblock Novelty estimation in developmental networks: Acetylcholine and
  norepinephrine.
\newblock In {\em Neural Networks (IJCNN), The 2013 International Joint
  Conference on}, pages 1--8. IEEE, 2013.

\bibitem{fuchs2010oxford}
P.~Fuchs.
\newblock {\em Oxford Handbook of Auditory Science: The Ear}.
\newblock OUP Oxford, 2010.

\bibitem{glickstein2000visual}
M.~Glickstein.
\newblock How are visual areas of the brain connected to motor areas for the
  sensory guidance of movement?
\newblock {\em Trends in neurosciences}, 23(12):613--617, 2000.

\bibitem{hopcroft2006automata}
J.~Hopcroft, R.~Motwani, and J.~Ullman.
\newblock Automata theory, languages, and computation.
\newblock {\em International Edition}, 24, 2006.

\bibitem{hudspeth1989ear}
A.~J. Hudspeth.
\newblock How the ear's works work.
\newblock {\em Nature}, 341(6241):397--404, 1989.

\bibitem{WWN1}
Z.~Ji, J.~Weng, and D.~Prokhorov.
\newblock Where-what network 1: Where and what assist each other through
  top-down connections.
\newblock In {\em Development and Learning, 2008. ICDL 2008. 7th IEEE
  International Conference on}, pages 61--66. IEEE, 2008.

\bibitem{jimenez2012review}
S.~Jim{\'e}nez, T.~Rosa, S.~Fern{\'a}ndez, F.~Fern{\'a}ndez, and D.~Borrajo.
\newblock A review of machine learning for automated planning.
\newblock {\em The Knowledge Engineering Review}, 27(04):433--467, 2012.

\bibitem{lehmann2006theory}
E.~Lehmann.
\newblock {\em Theory of point estimation}.
\newblock New York: Wiley, 1983.

\bibitem{liu1997effects}
L.~Liu, J.~He, and G.~Palm.
\newblock Effects of phase on the perception of intervocalic stop consonants.
\newblock {\em Speech Communication}, 22(4):403--417, 1997.

\bibitem{WWN3}
M.~Luciw and J.~Weng.
\newblock {Where What Network 3}: Developmental top-down attention with
  multiple meaningful foregrounds.
\newblock In {\em Proc. IEEE Int'l Joint Conference on Neural Networks}, pages
  4233--4240, Barcelona, Spain, July 18-23, 2010.

\bibitem{WWN4}
M.~Luciw and J.~Weng.
\newblock Where-what network-4: The effect of multiple internal areas.
\newblock In {\em Development and Learning (ICDL), 2010 IEEE 9th International
  Conference on}, pages 311--316. IEEE, 2010.

\bibitem{martin1991introduction}
J.~Martin.
\newblock {\em Introduction to Languages and the Theory of Computation},
  volume~4.
\newblock McGraw-Hill NY, 1991.

\bibitem{piaget1959language}
J.~Piaget.
\newblock {\em The language and thought of the child}, volume~5.
\newblock Psychology Press, 1959.

\bibitem{piaget1973child}
J.~Piaget.
\newblock {\em The child and reality: Problems of genetic psychology.(Trans.
  Arnold Rosin).}
\newblock Grossman, 1973.

\bibitem{piaget2013success}
J.~Piaget.
\newblock {\em Success and understanding}.
\newblock Routledge, 2013.

\bibitem{piaget2015grasp}
J.~Piaget.
\newblock {\em The Grasp of Consciousness (Psychology Revivals): Action and
  Concept in the Young Child}.
\newblock Psychology Press, 2015.

\bibitem{piaget1952origins}
J.~Piaget and M.~Cook.
\newblock {\em The origins of intelligence in children}.
\newblock International Universities Press, Madison, New York, 1952.

\bibitem{poynton2012digital}
C.~Poynton.
\newblock {\em Digital video and HD: Algorithms and Interfaces}.
\newblock Elsevier, 2012.

\bibitem{sarma2014phoneme}
M.~Sarma and K.~K. Sarma.
\newblock {\em Phoneme-based speech segmentation using hybrid soft computing
  framework}, volume 550.
\newblock Springer, 2014.

\bibitem{schmidhuber2015deep}
J.~Schmidhuber.
\newblock Deep learning in neural networks: An overview.
\newblock {\em Neural networks}, 61:85--117, 2015.

\bibitem{WWN5}
X.~Song, W.~Zhang, and J.~Weng.
\newblock Where-what network 5: Dealing with scales for objects in complex
  backgrounds.
\newblock In {\em Proc. Int'l Joint Conference on Neural Networks}, pages
  2795--2802, San Jose, CA, July 31 - August 5, 2011.

\bibitem{Song15}
X.~Song, W.~Zhang, and J.~Weng.
\newblock Types, locations, and scales from cluttered natural video and
  actions.
\newblock {\em IEEE Transactions on Autonomous Mental Development},
  7(4):273--286, 2015.

\bibitem{tilney1992actin}
L.~G. Tilney, M.~S. T, , and D.~J. Derosier.
\newblock Actin filaments, stereocilia, and hair cells: how cells count and
  measure.
\newblock {\em Annual Review of Cell Biology}, 8(8):257--274, 1992.

\bibitem{turing1937computable}
A.~Turing.
\newblock On computable numbers, with an application to the
  entscheidungsproblem.
\newblock {\em Proceedings of the London mathematical society}, 2(1):230--265,
  1937.

\bibitem{VonMelchner00}
L.~VonMelchner, S.~L. Pallas, and M.~Sur.
\newblock Visual behaviour mediated by retinal projections directed to the
  auditory pathway.
\newblock {\em Nature}, 404:871--876, 2000.

\bibitem{Voss13}
P.~Voss.
\newblock Sensitive and critical periods in visual sensory deprivation.
\newblock {\em Frontiers in Psychology}, 4:664, 2013.
\newblock doi: 10.3389/fpsyg.2013.00664.

\bibitem{vygotsky1980mind}
L.~Vygotsky.
\newblock {\em Mind in society: The development of higher psychological
  processes}.
\newblock Harvard university press, 1980.

\bibitem{WWN6}
Y.~Wang, X.~Wu, and J.~Weng.
\newblock Brain-like learning directly from dynamic cluttered natural video.
\newblock In {\em Proc. International Conference on Brain-Mind}, pages 51--58,
  East Lansing, MI, July 14-15, 2012.
\newblock Also technical report MSU-CSE-12-5 at
  http://www.cse.msu.edu/\~{}weng/research/LM.html.

\bibitem{cobb1994}
S.~Wanska.
\newblock The relationship of spatial concept development to the acquisition of
  locative understanding.
\newblock {\em The Journal of genetic psychology}, 145(1):11--21, 1994.

\bibitem{WengNAI12}
J.~Weng.
\newblock {\em Natural and Artificial Intelligence: Introduction to
  Computational Brain-Mind}.
\newblock BMI Press, Okemos, Michigan, 2012.

\bibitem{weng2012symbolic}
J.~Weng.
\newblock Symbolic models and emergent models: A review.
\newblock {\em {IEEE Trans. Autonomous Mental Development}}, 4(1):29--53, 2012.

\bibitem{weng2015brain}
J.~Weng.
\newblock Brain as an emergent finite automaton: A theory and three theorems.
\newblock {\em International Journal of Intelligent Science}, 5(2):112--131,
  2015.
\newblock received Nov. 3, 2014 and accepted by Dec. 5, 2014.

\bibitem{weng1993motion}
J.~Weng, T.~S. Huang, and N.~Ahuja.
\newblock {\em Motion and Structure from Image Sequences}.
\newblock Springer-Verlag, New York, 1993.

\bibitem{weng2009dually}
J.~Weng and M.~Luciw.
\newblock Dually optimal neuronal layers: Lobe component analysis.
\newblock {\em {IEEE Trans. Autonomous Mental Development}}, 1(1):68--85, 2009.

\bibitem{WWN7}
X.~Wu, Q.~Guo, and J.~Weng.
\newblock Skull-closed autonomous development: Wwn-7 dealing with scales.
\newblock In {\em Proc. International Conference on Brain-Mind. East Lansing,
  Michigan: BMI Press}, pages 1--8. Citeseer, 2013.

\bibitem{wu2017actions}
X.~Wu and J.~Weng.
\newblock Actions as contexts.
\newblock In {\em 2017 Int. Joint Conf. Neural Networks}, pages 214--221,
  Anchorage, AK, 2017.

\bibitem{zhengj2016mobile}
Z.~Zheng and J.~Weng.
\newblock Mobile device based outdoor navigation with on-line learning neural
  network: a comparison with convolutional neural network.
\newblock In {\em Proc. 7th Workshop on Computer Vision in Vehicle Technology
  (CVVT 2016) at CVPR 2016}, pages 11--18, Las Vega, June 269 2016.

\end{thebibliography}

\clearpage

	\section*{Appendix}

	\subsection{Proof of Lemma \ref{th: learns MLE}}
	\textit{Proof.} We break down the proof into two steps. Step 1 shows that the firing pattern in each zone is actually a binary ML estimator of the current input conditioned on $N(t)$. Step 2 shows that after learning the weights are then incrementally updated according to ML estimation.

	\textit{Step 1, case 1} (new neuron enters learning stage at time $t + 1$): When a new neuron $g$ enters learning stage for the first time, it memorizes the input vector $\p(t + 1)$ perfectly. This neuron also suppresses other neurons inside its $\bar{g}$ from firing (eliminating noises), thus becoming the ML estimator of the current input $\p(t + 1)$.

	\textit{Step 1, case 2} (no new neurons are added at time $t + 1$): For each neuron $g \in Y, Z$, consider its inhibition zone at time $t$: $\bar{g}(t)$. $c(g, t)$ is the number of neurons in its inhibition zone $\bar{g}(t)$. The normalized weights of these $c$ neurons can be denoted as $(\w_1, \w_2, ..., \w_{c(g, t)})$.

	Then we can define $c(g, t)$ Voronoi regions $R_j, j = 1,2,...,c(g, t)$ in $L(g, t)$ (receptive field of $g$ at time $t$, which is a linear subspace of $X \times Y \times Z$), where each $R_j$ contains all $\p \in L(g, t)$ that are closer to $\w_j$ than to other $\w_i$:

	\begin{equation}\nonumber
		R_j = \{\p|j = \arg\max_{1 \leq i \leq c} \w_i \cdot \p \},\quad j = 1,2, ..., c
	\end{equation}

	Given observation $\p(t + 1)$, the conditional probability density $h(\p(t + 1)|L(g, t), W(g, t))$ is zero if $\p(t + 1)$ falls out of the Voronoi region of neuron $g$:

	\begin{equation}
		\label{eq: voronoi prob}
		f\{\p(t + 1)|L(g, t), W(g, t)\} =
		\begin{cases}
			f_i \{\p(t + 1)|L(g, t), W(g, t)\}, \\
			\quad \quad \quad \textnormal{if}\quad \p(t + 1) \in R_i; \\
			0, \quad \textnormal{otherwise}
		\end{cases}
	\end{equation}
	where $f_i \{\p(t + 1)|L(g), W(g)\}$ is the probability density within $R_i$. Note that the distribution of $f_i \{\p(t + 1)|L(g), W(g)\}$ within $R_i$ is irrelevant as long as it integrates to 1.

	Given $\p(t + 1)$, the ML estimator for the binary vector $\y(t)$ (also $\z(t)$) needs to maximize $f\{\p(t + 1)|N(t)\}$, which is equivalent to finding the set of firing neuron $n(t + 1)$:
	\begin{eqnarray}
		\label{eq: learning response MLE}
		n(t + 1) &=& \{ \arg \max_{g\in Y, Z}f\{\p(t + 1) | N(t)\} \} \nonumber\\
		&=& \{g | g = \arg \max_{g\in Y, Z}f\{\p(t + 1) |L(g, t), W(g, t)\} \nonumber \\
		&=& \{g | g = \arg \max_{j \in \bar{g}(t)} \w(j) \cdot \p(t + 1) \}
	\end{eqnarray}
	since finding the ML estimator in Eq. \eqref{eq: voronoi prob} is equivalent to finding the Voronoi region to which
	\(\p(t)\) belongs to.  This is exactly what the $Y$ zone does, supposing $k = 1$ for top-$k$ competition for each neuron's competition zone.

	\textit{Step 2}: Here we are going to show the statistical efficiency of the LCA learning rule of $W(g)$, where $g\in n(t)$ (meaning that $g$ is among the firing neurons at time $t$).

	Eq. (11) in \cite{weng2009dually} shows that the \textit{candid} version of LCA is actually an incremental estimation of the average of the input that triggers firing in that specific neuron:
	\begin{eqnarray}
		\w(g, t + 1) & = & \frac{a(g, t) - 1}{a(g, t)}\w(g, t) + \frac{1}{a(g, t)} \p(t + 1) \nonumber\\
		& = & \frac{1}{a(g, t)} \Sigma_{i = 1}^{a(g, t)} \p(t_i)
	\end{eqnarray}
	where $\w(g,t)$ is the weight vector $\w(g)$ at time $t$. $a(g, t)$ is the age of neuron $g$ at time $t$. $t_i, i = 1,2,...,a(g, t)$ are the times where neuron $g$ wins dynamic top-$k$ competition and fires.

	Same as the proof of optimality in LCA \cite{weng2009dually}, statistical estimation theory reveals that for many distributions (e.g., Gaussian and exponential distributions), the sample mean is the most efficient estimator of the population mean. This follows directly from Th. 4.1, p.429-430 in \cite{lehmann2006theory}, which states
	that under some regularity conditions satisfied by many distributions (such as Gaussian and exponential distributions), the maximum likelihood estimator (MLE) $\hat{\boldsymbol{\theta}}$ of the parameter vector $\boldsymbol{\theta}$ is
	asymptotically efficient, in the sense that its asymptotic covariance matrix is the Cramer-Rao information bound (the lower bound) for all unbiased estimators via convergence in probability to a normal distribution:
	\begin{equation}
		\sqrt{n}(\hat{\boldsymbol{\theta}} - \boldsymbol{\theta}) \xrightarrow{p} N \{0, I(\boldsymbol{\theta})^{-1}\}
	\end{equation}
	in which the Fisher information matrix $I(\boldsymbol{\theta})$ is the covariance matrix of the score vector 
	\[
	\{(\partial f(\mathbf{x}, \boldsymbol{\theta}))/(\partial \theta_1), ...,  (\partial f(\mathbf{x}, \boldsymbol{\theta}))/(\partial \theta_k) \}.
	\]
	 And $ f(\mathbf{x}, \boldsymbol{\theta})$ is the probability density of random vector $\mathbf{x}$ if the true parameter value is $\boldsymbol{\theta}$. The matrix $I(\boldsymbol{\theta})^{-1}$ is called information bound since under some regularity constraints, any unbiased estimator $\tilde{\boldsymbol{\theta}}$ of the parameter vector $\boldsymbol{\theta}$ satisfies $\text{cov}(\tilde{\boldsymbol{\theta}} - \boldsymbol{\theta}) \geq I(\boldsymbol{\theta})^{-1} / n$ (see, e.g., \cite{lehmann2006theory}, p. 428 or \cite{weng1993motion}, p. 203-204]).

	Thus, as Weng et al. showed in \cite{weng2009dually}, the LCA learning algorithms learns the optimal weights in the sense of maximum likelihood estimation.

	Combining step 1 and step 2 we would have:

	\begin{equation}
		\nonumber
		\theta(t + 1) = \max_\theta(t) f\{\p(t)|N(t-1) \}
	\end{equation}

	\subsection{Proof of Theorem \ref{th:optimal_learning}}
	\textit{Proof.} Intuitively, although  $f'\{\p(t+1) | \cX_0^t, \cZ_0^t, \Gamma\}$ is in a different format compared to the $f\{\p(t + 1) | N(t)\}$ in Lemma \ref{th: learns MLE}, the two probability functions are acting the same as $N(t)$ is determined by $\cX_t, \cZ_t$ and $\Gamma$. As there is no random weight initialization in DN-2, two DN-2 equipped learning agents would be exactly identical given the same hyper-parameter $\Gamma$ and learning experience $\cX_t, \cZ_t$. To formally prove this we are going to recursively use the conclusion of Lemma \ref{th: learns MLE}.



	At $t+1$, we can reuse Eq. \eqref{eq: learning response MLE} from Lemma \ref{th: learns MLE} together with the condition of incremental learning:
	\begin{eqnarray}
		\theta(t + 1) & = & \max_{\theta(t)}f\{\p(t + 1) | N(t)\}                                          \nonumber\\
		& = & \max_{\theta(t)}f\{\p(t+1) | \cL(N(t-1), \x(t), \z(t), \Gamma)\}  \nonumber\\
		& = & \max_{\theta(t)}f^1\{\p(t+1) | N(t-1), \cX_{t}^t, \cZ_{t}^t, \Gamma\}      \nonumber\\
		& = & \max_{\theta(t)}f^2\{\p(t+1) | N(t-2), \cX_{t-1}^t, \cZ_{t-1}^t, \Gamma\}      \nonumber\\
		& = & ... \nonumber \\
		& = & \max_{\theta(t)}f^t\{\p(t+1) | N(0), \cX_{0}^t, \cZ_{0}^t, \Gamma\}  \nonumber \\
		& = & \max_{\theta(t)}f'\{\p(t+1) | \cX_{0}^t, \cZ_{0}^t, \Gamma\}
	\end{eqnarray}
	where $\cL$ is the learning function of the network, and $f^i$ is the probability density function of the current input $\p$, conditioned on the network $i$ time steps ago and the learning experience from $t-i$ to $t$.

\clearpage
\section*{CLAIMS}
What is claimed is
\begin{enumerate}
\item Claims 21: (currently amended) An improvement to a developmental network implemented in computer hardware, the developmental network comprising at least one neural network having a plurality of neurons organized into a hierarchy of levels comprising an $X$ area associated with sensory information, a $Z$ area associated with motor information, and a hidden $Y$ area between the $X$ area and the $Z$ area, the improvement comprising:\\
\hspace*{0.3in}
each neuron has an associated location in a computer simulated physical space; and\\
\hspace*{0.3in}
the internal representation of the hierarchy of the $X$ area, $Y$ area, and $Z$ area is a fluid hierarchy.
\item Claim 22 (currently amended): The improvement of Claim 21, wherein each neuron has an inhibition zone that is a zone for neuronal competition that determines whether the neuron fires. 
\item Claim 23 (currently amended): The improvement of Claim 22, wherein each neuron associated with at least one negative neuron.  
\item Claim 24 (currently amended): The improvement of Claim 23, wherein each neuron includes one or more of excitatory and inhibitory connections, the excitatory connections correspond to a receptive field of the neuron and the inhibitory connections correspond to a receptive field of one or more negative neurons that inhibit the neuron, such that the inhibition zone for any particular positive neuron changes according to a training or exploration experience of the particular neuron. 
\item Claim 25 (currently amended):  The improvement of claim 24, wherein each neuron is classified as presynaptic, post-synaptic, or both; and wherein each post-synaptic neuron uses the strengths of connections from all the input neurons to its negative neuron to identify the boundary of its competition zone.
\item Claim 26 (currently amended):  The improvement of claim 25, wherein each neuron has a pre-response and each post-synaptic neuron determines whether to fire by comparing its pre-response with all the pre-responses of all neurons in its competition zone. 
\item Claim 27 (currently amended): The improvement of claim 26, wherein each post-synaptic neuron has a pre-response and wherein each post-synaptic neuron makes the determination of whether to fire by determining whether the pre-response of the neuron is valued among the top-$k$ values in the competition zone, where $k$ is a percentage number.
\item Claim 28 (currently amended): The improvement of claim 22, wherein each neuron in the Y area has a neuronal weight associated with a pre-synaptic neuron and a post-synaptic neuron and wherein the post-synaptic neuron is updated by a Hebbian-like mechanism. 

\item Claim 29 (previously amended):  The improvement of claim 21, wherein each associated location is a 3D location and the computer-simulated physical space is a 3D space. 

\item Claim 30 (previously amended):  The improvement of claim 29, further comprising a process for computer visualization of the neurons on a computer screen that shows one or more neuronal parameters for each neuron according to its associated location.

\item Claim 31 (currently amended):  The improvement of claim 21, further comprising a process for updating the neural network recursively from each indexed time $t$ to next indexed time $t+1$\\
\hspace*{0.2in} wherein the said process recursively decomposes the computation of an unbounded number of consecutive life times optimality in maximum-likelihood from an initial time to time $t + 1$ into:\\
\hspace*{0.2in} using a previous neural network machine that has already optimally computed recursively   from the initial time to time $t$, and determining a computation of an optimal neural network machine for time $t+1$; and\\
\hspace*{0.2in} wherein the said recursive decomposition is conditioned on an incremental learning restriction, a sensory experience, and a limited resource.
\item Claim 32 (currently amended): The improvement of Claim 21 further comprising a process for incrementally generating neurons through an experience of training or exploration.
\item Claim 33 (currently amended)  The improvement of claim 21, further comprising:\\
\hspace*{0.2in} an initialization process for constructing the neural network that specifies whether to connect with one or more of the X, Y, and Z areas; and \\
\hspace*{0.2in} a synaptic maintenance process for changing the fluid hierarchy;\\
\hspace*{0.2in} wherein each neuron is classified as pre-synaptic, post-synaptic, or both and during the synaptic maintenance, each post-synaptic neuron determines whether to cut an input connection where a deviation of weights is high and to re-connect an input connection when the deviation of weights is low. 
\item Claim 34 (previously amended):  The improvement of claim 21, further comprising a learning process, wherein the neural network learns an emergent Turing machine or an emergent universal Turing machine as representations of experience of teaching or exploration.

\item Claim 35 (previously amended):   The improvement of claim 21, wherein the fluid hierarchy internally spans a lifetime.  

\item Claim 36 (previously amended):   The improvement of claim 21, wherein the neural network selectively uses or selectively disregards combinations of features at different levels of representations within the fluid hierarchy.

\item Claim 37 (currently amended):   The improvement of claim 21, \\
\hspace*{0.2in}  wherein the Z area of the neural network associated with motor information has at least one overt and at least one covert neuron for each of a plurality of effectors;\\
\hspace*{0.2in}  wherein the neural network is trained with one or more plans with or without task costs using the explicitly taught overt and covert neurons; and\\
\hspace*{0.2in}  wherein a recalled experienced skill that is not exactly the same as taught or explored is called an emergent skill. 

\item Claim 38 (currently amended):   The improvement of claim 37, where the neural network chooses to use one plan from the one or more plans after the network has been trained with one or more plans.

\item Claim 39 (previously amended):   The improvement of claim 21, further comprising a use of the neural network for at least two modalities of vision, audition, natural language, in a real or a simulated environment.

\item Claim 40 (previously amended):   The improvement of claim 21, further comprising a use of the neural network for strong Artificial Intelligence (AI) wherein the strong AI is not task-specific but is conditioned on incremental learning framework restrictions, a learning or exploration experience, and a limited amount of computational resources. 
\end{enumerate}

\clearpage

\section*{ABSTRACT OF THE DISCLOSURE}
This invention includes a new type of neural network that is able to automatically and incrementally  
generate an internal hierarchy without a need to handcraft a static hierarchy of network areas
and a static number of levels and the static number of neurons in each network area or level.
This capability is achieved by enabling each neuron to have its own dynamic  
inhibitory zone using neuron-specific inhibitory connections.

\end{document}